\documentclass[10pt,journal,compsoc]{IEEEtran}
						
						\usepackage[pagebackref=true,breaklinks=true,letterpaper=true,colorlinks, bookmarks=false]{hyperref}
						\usepackage[cmex10]{amsmath}
						\usepackage{epsfig}
\usepackage{times}
											\usepackage{graphicx}
											\usepackage{amsmath}
											\usepackage{amssymb}
											\usepackage{subfigure}
											\usepackage{bm}
											\usepackage{cite}
											\usepackage{array}
											\usepackage{xcolor,colortbl}
										        \usepackage[english]{babel}
\begin{document}
						
						\title{Understanding Trajectory Behavior: A Motion Pattern Approach}

						\author{Mahdi~M. Kalayeh,
						        Stephen~Mussmann$^{*}$, Alla~Petrakova$^{*}$, Niels da Vitoria Lobo
						        and~Mubarak~Shah

						\thanks{
 Mahdi M. Kalayeh, Alla Petrakova, Niels da Vitoria Lobo and Mubarak Shah are with Department of Electrical Engineering and Computer Science, University of Central Florida, Orlando, Fl, 32816. Stephen Mussmann is with Department of Mathematics and Department of Computer Science, Purdue University,  West Lafayette, IN 47907. Authors* have contributed equally and are listed alphabetically. E-mail: Mahdi@eecs.ucf.edu.\protect
						}
						}

						\IEEEcompsoctitleabstractindextext{%
\begin{linespread}{1.0}
						\begin{abstract}
			Mining the underlying patterns in gigantic and complex data is of great importance to data analysts. In this paper, we propose a motion pattern approach to mine frequent behaviors in trajectory data. Motion patterns, defined by a set of highly similar flow vector groups in a spatial locality, have been shown to be very effective in extracting dominant motion behaviors in video sequences. Inspired by applications and properties of motion patterns, we have designed a framework that successfully solves the general task of trajectory clustering. Our proposed algorithm consists of four phases: flow vector computation, motion component extraction, motion component's reachability set creation, and motion pattern formation. For the first phase, we break down trajectories into flow vectors that indicate instantaneous movements. In the second phase, via a Kmeans clustering approach, we create motion components by clustering the flow vectors with respect to their location and velocity. Next, we create motion components' reachability set in terms of spatial proximity and motion similarity. Finally, for the fourth phase, we cluster motion components using agglomerative clustering with the weighted Jaccard distance between the motion components' signatures, a set created using path reachability. We have evaluated the effectiveness of our proposed method in an extensive set of experiments on diverse datasets. Further, we have shown how our proposed method handles difficulties in the general task of trajectory clustering that challenge the existing state-of-the-art methods.
						
						\end{abstract}
\end{linespread}

						}

						\maketitle

						\IEEEdisplaynotcompsoctitleabstractindextext

						\IEEEpeerreviewmaketitle

						\section{Introduction}

						\IEEEPARstart{R}{ecently}, there has been a great amount of research in mining frequent behaviors from trajectory data. A trajectory is an N dimensional path that does not necessarily correspond to a physical object, but may correspond to the time evolution of an N dimensional feature vector. In this paper, we propose a novel approach, inspired by the concept of motion patterns, for extracting dominant trajectory behaviors; while our method will be described for the case of 2 dimensional trajectories, it can be extended  to higher dimensions for further applications. Seeking to find common behaviors in trajectory data, some methods group whole trajectories into clusters while others attempt to mine regional patterns that trajectories follow during part of their evolutions. Our proposed method aligns more closely to the latter category. In this section, we begin with introducing a subset of very rich literature in trajectory clustering and try to keep a balance between the recent methods and popular ones. Due to the existence of a large number of methods in the literature and the space limitation, selecting a subset of the literature is inevitable. Next, we explain how methods performing scene activity understanding in video sequences can be seen as trajectory clustering tools. And finally, we intuitively introduce our motion pattern approach as a general tool for understanding trajectory behaviors. 

Many of the existing trajectory clustering methods approach the problem by first defining a similarity function for trajectories and then using one of the well-established clustering procedures. For instance, \cite{Atev} compared clustering results obtained by using the trajectory similarity measures Dynamic Time Warping (DTW), Longest Common Subsequence (LCSS), and modified Hausdorff distance in combination with agglomerative and spectral clustering. Fu $\it{et. al.}$\cite{Fu} used the average distance between corresponding trajectory points as the similarity measure, which required pre-processing and resampling of trajectories. The clustering was later applied in two steps, with the first step producing clusters corresponding to the larger dominant paths, and the second one subsequently refining results obtained by the first one. Morris and Trivedi \cite{Morris} surveyed performance of a wide variety of distance measures and clustering algorithms on several datasets with varying characteristics. Specifically, they compared trajectories using \cite{HU}, PCA \cite{PCA}, DTW \cite{DTW}, LCSS \cite{LCSS}, \cite{PF}, and modified Hausdorff \cite{atev2006} distance and then obtained clusters using direct \cite{morris2008}, divisive \cite{biliotti2005}, agglomerative \cite{LCSS}, hybrid \cite{Hybrid}, graph \cite{li2006} and spectral clustering techniques \cite{HU}. Ferreira $\it{et. al.}$\cite{vector_field} have proposed Vector field Kmeans that treats the trajectories as a whole and attempts to follow an iterative model similar in form to Kmeans clustering. In this approach, clusters are subsets of the trajectories. First, the trajectories are partitioned randomly into K clusters. Then, for each cluster, the trajectories in that cluster are used to form one of K vector fields. Further, clusters are updated by assigning each trajectory to the vector field that it fits best. These two steps are repeated in an iterative fashion until a convergence criteria is met.
DivCluST proposed by Wu $\it{et. al.}$\cite{Divclust} is a partition based approach that clusters trajectories according to their speed in addition to their shape and position. First, it partitions the trajectories into representative segments by combining adjacent trajectory segments that have similar direction and speed. Then, a method similar to Kmeans is used to find cluster centers. The clusters are initialized as random subsets of the representative segments. Next, the representative segments of each cluster are averaged to form a cluster center. Finally, as in Kmeans, each of the representative segments is associated with a cluster center, then each cluster center is updated to be the average of the cluster's representative segments. This process will be repeated until a convergence criteria is met. Ulm $\it{et. al.}$\cite{ulm} have proposed a trajectory clustering approach that is distinctive in its online clustering ability. It does not explicitly treat the trajectories as a whole, but in practice, it behaves as if it does. They defined a cluster as a vector field defined on a subset of the spatial range. When a trajectory is received, it is converted into a new vector field. If at any time two vector fields are similar, they are merged together into one. This method uses the entire trajectory to make a vector field and merging process largely conserves the shape of both fields that are being merged. Thus, the algorithm in practice behaves similarly to those approaches that treat the trajectories as a whole.

Giannotti $\it{et. al.}$\cite{giannotti} proposed a very different way of treating trajectories where their density is used to form frequently visited rectangular regions. Then, each of the trajectories is represented as a sequence of visited regions with the transition times. They mined common subsequences with similar transition times that represent commonly taken paths similar in interpretation to trajectory clusters. It is worth mentioning that because of the exact phrasing of the problem, this approach gives very redundant clusters in practice. Lee $\it{et. al.}$\cite{Traclus} observed that clustering trajectories as a whole fails to reveal portions of trajectories that exhibit a common behavior. To address this, they proposed TRACLUS, an algorithm that first partitions trajectories into a set of line segments and then groups similar segments into clusters using a density-based clustering algorithm for line segments (largely analogous to DBSCAN). This work has been further extended in \cite{Traclass}, which proposed a feature generation framework, TraClass, that uses a combination of region-based and trajectory-based clusterings. Specifically, TraClass finds homogeneous regions where trajectories of a certain class are predominant, then uses a class-conscious adaptation of TRACLUS to group trajectory partitions into clusters that have high discriminative power. After generating features in this manner, TraClass maps trajectories into a feature vector, where each entry corresponds to a region-based or a trajectory-based cluster.

										\subsection{Scene Activity Understanding Methods}		
			Given the decreasing cost of collecting data and the great importance of surveillance videos for providing security and monitoring ongoing activities, many researchers, mostly in the computer vision community, dedicated major efforts to develop automatic scene modeling and intelligent activity understanding systems. These research works aim to learn frequent movement behaviors and activity profiles in videos. Using that, they can also detect abnormal behaviors and in some cases improve object tracking performance. Given a video sequence, some of these methods obtain trajectories associated with the moving objects in the scene, while others, due to the challenges such as highly dense scenes, frequent moving object occlusions or poor object tracking performance, either compute short-term trajectories (tracklets) or compute sparse \cite{lucas1981} or dense optical flows \cite{gurka1999} in the feature extraction phase. Then, different kinds of algorithms are employed to infer frequent motion behaviors taking place in the video sequences from these extracted features. In this paper we will refer to these frequent motion behaviors as motion patterns. Semantically, motion patterns are very similar to the regional patterns mined by those groups of trajectory clustering methods that seek common behaviors in subsegments of trajectories. Therefore, we claim that a motion pattern approach can be employed to solve the general task of trajectory clustering.		
			
Junejo $\it{et. al.}$\cite{junejo2004} cluster similar trajectories by performing a min-cut graph clustering algorithm recursively. Each node in the graph represents a trajectory while the weights of the edges between nodes are determined by the Hausdorff distance between corresponding trajectories. Modeled paths are later used to detect unusual trajectories based on spatial velocity and curvature features. \cite{hu2008detecting} computes instantaneous flow vectors that include location and velocity information from either optical flows or long-term/short-term trajectories. Given the location of a point and it's velocity (based on flow vectors), a kernel based estimation similar to mean shift approach \cite{comaniciu2002} is employed to generate the corresponding velocity in the next step. The next location of each point can be estimated using its current location and computed velocity. Performing this procedure recursively will generate the sink path that estimates the movement path of the initial point. Next, obtained sinks are clustered based on the similarity of their location and directions in addition to the Hausdorff distance between their sink paths. These clusters represent the frequent motion behaviors. Basharat $\it{et. al.}$\cite{basharat2008learning} have proposed a framework that does not explicitly generate motion patterns; however, the per-pixel learned multivariate probability density function can be used to understand the frequent motion behaviors in the scene. Given a set of trajectories obtained from moving objects, transition vectors that represent size of the moving object, transition time and location of the object after the transition are computed. Then, for any given location in the scene, transition vectors that pass through that location contribute to a multivariate Gaussian Mixture Model (GMM) that models the pixel-level pdf. Given such a model, most likely paths can be generated by initializing the starting points in the same fashion as \cite{lasdas2012} or by taking a Markov Chain Monte Carlo (MCMC) sampling based scheme such as \cite{saleemi2009}. 
Hu $\it{et. al.}$\cite{hu2008learning} compute flow vectors by extracting sparse or dense optical flow from video sequences. Then, they create a neighborhood directional graph that indicates the distance between flow vector pairs. Finally, a hierarchical agglomerative clustering algorithm generates motion patterns as the graph clusters of the neighborhood graph. Saleemi $\it{et. al.}$\cite{saleemi2009} have proposed a probabilistic framework to model scene dynamics. Given the long-term/short-term trajectories of moving objects in a video sequence obtained from an object tracking algorithm, a 5 dimensional feature space is created where each feature point represents the location information of initial and final stages of a moving object and the duration of the time interval for such a transition. Then, motion patterns , in the form of a multivariate pdf of spatiotemporal parameters, will be learned using kernel density estimation. Finally, a Markov Chain Monte Carlo (MCMC) sampling based scheme is utilized to sample the pdf. For any starting location as the initial state, MCMC sampling based scheme generates the probable random walks that are most likely to happen in the scene in a progressive procedure. These likely paths express the most probable paths that trajectories have taken in the training stage. Therefore, those are semantically similar to motion patterns. Lasdas $\it{et. al.}$\cite{lasdas2012} present a method to extract dominant dynamic properties in crowded scenes. Their proposed pipeline begins with extracting fixed length tracklets using KLT tracker \cite{baker2004lucas}. Then, a set of validation tests and pre-processing steps refine the tracklet set. In order to have a more robust representation, a grid of equally spaced points is overlaid on the scene. The pipeline continues with clustering tracklets in the neighborhood of each of these points using a mean shift clustering algorithm with respect to the direction of the tracklets. Next, mean flow vectors that are assigned to each grid point are modeled as a Gaussian Process (GP) using \cite{rasmussen2006gaussian}. Finally, given a starting point and the mixture of GP regression models, a sequential sampling scheme generates the trajectories that follow the most frequently observed patterns in the training phase. These generated paths are supposed to be the most probable paths that might originate from the starting point and are semantically similar to the motion patterns.

Zhao and Medioni \cite{zhao2011} proposed a framework to infer motion patterns in videos in an unsupervised learning fashion. Given a video with moving objects, they extract foreground motion blobs (connected pixels that belong to a moving object) using Robust Alignment by Sparse and Low Rank Decomposition (RASL) \cite{peng2010rasl}. Performing local associations using \cite{prokaj2011inferring} and \cite{prokaj2011using}, they obtain the initial tracklets. Here the tracklets are sequences of ordered spatial coordinates of motion blobs. Then, using a 2 dimensional Tensor Voting algorithm \cite{mordohai2010dimensionality}, they compute the tangent direction of every tracklet point. The direction information provided after 2 dimensional Tensor Voting is less noisy and more consistent than transition vectors from one observation to the next one. Further, a similarity graph of tracklets based on multiple kernel types is generated and, employing a graph spectral clustering approach, the nonlinear manifold of tracklets is grouped into segments. These segments indicate local motion patterns; while \cite{zhao2011} focuses on global motion patterns. Therefore, a kernel density estimation algorithm is utilized to propagate the local motion pattern information to all the pixels in the scene and form global motion patterns.
 As the last piece of work to cover in this section, we briefly explain the approach that \cite{saleemi2010} took to provide a scene understanding framework, a framework that we believe is capable of solving the trajectory clustering task. The pipeline begins with dividing a video into disjoint video clips and extracting dense optical flow from them. Then, each video clip is represented as a 4 dimensional feature space that contains location, magnitude and orientations of the instantaneous optical flow vectors belonging to that particular video clip. Next, each of the 4 dimensional feature spaces are modeled as a mixture of Gaussians where Kmeans clustering is employed to initialize the model parameters. The pipeline continues with generating a graph that treats each Gaussian component as a node where the edge between two nodes is determined by their reachability from each other and associated temporal proximity. To clarify, the reachability is defined as the probability of observing a component in the neighborhood of the other one in a specific time interval. Finally, through a connected component analysis on the aforementioned graph, motion patterns are obtained. Since similar motion patterns might occur at different time stamps, Kullback-Leibler (KL) divergence is employed to merge similar motion patterns happening at different times.

											\subsection{Proposed Approach}		

	We explained the trajectory clustering problem and gave a brief overview of a group of methods that seek to find regional patterns in trajectory data rather than clustering whole trajectories. Then we showed that the motion pattern inference methods trying to perform scene understanding in video sequences are semantically generating the same output as the aforementioned trajectory clustering methods. Therefore, we claim that a motion pattern approach can be employed to solve the general task of trajectory clustering. In this paper, we propose a novel framework that is designed to overcome the challenges provided by the trajectory clustering task. Given a set of trajectories, we break them down into sets of flow vectors and generate a 4 dimensional feature space. Each point in this feature space represents location and velocity of a flow vector. Then, using Kmeans clustering, we represent the feature space in a less noisy and compact form. Clusters obtained from Kmeans clustering are called motion components. Next, we define the reachability set in which a pair of two motion components exist if they are reachable from each other. The reachability set maintains local similarities between motion components and cannot evaluate motion component's contribution in generating the global trajectory behaviors. Hence, we define the signature which represents global behavior of a motion component. Finally, a weighted Jaccard distance between motion components' signatures is used as the distance metric in an agglomerative clustering scheme to find motion patterns in trajectory data. These motion patterns illustrate regional behaviors of trajectories and are semantically similar to the trajectory clusters of those methods which break trajectories into segments and seek for regional similarities among them. The main contributions of our proposed method are as follows:

	\begin{itemize}
			
										\item {\textit{We propose a motion pattern approach to mine trajectory clusters of arbitrary shapes, simultaneously in dense and sparse regions of data. Generally, trajectory clustering methods either mine clusters in dense regions and lose clusters in regions where trajectories are spread sparsely or extract clusters in sparse regions while generating redundant clusters in dense regions}}.

										\item {\textit{The proposed approach is capable of recovering the whole trajectory cluster as a single motion pattern even in cases of highly curved underlying behaviors. Most of the trajectory clustering methods break down the whole cluster into segments as they fail to establish the connection between the segments when trajectories happen to have highly curved shapes}}.

	\item {\textit{By defining the signatures of motion components, we combine the local and global contributions of motion components such that it helps us distinguish between trajectory clusters that are identical in some localities but are different in the global scale. This strategy is useful for mining trajectory clusters that merge together or split into multiple ones}}.

	\item {\textit{Generally, trajectory clustering methods cluster similar trajectories that are spread over a wide region into multiple clusters as model parameters are set globally. However, our proposed approach generates only a single cluster in these cases as the capability of allowing semi-lateral movement is embedded in its framework}}.

	\item {\textit{The proposed method is able to correctly cluster two spatially overlapping groups of trajectories that differ only in the movement direction}}.
			\end{itemize}
								
								The remainder of this paper is organized as follows. Section 2 presents the problem definition and formulation details of our proposed method. Then, the detailed description of the experiments is given in Section 3, followed by the implementation details in Section 4. In Section 5, we discuss how our motion pattern approach can be used to explore temporal changes in frequent trends in the data with applications in various domains. Finally, we conclude the paper in Section 6.

						\section{Methodology}
											In this section, we will introduce our algorithm for mining frequent trajectory behaviors. Each block in our pipeline will be explained in detail while associated parameters will be discussed separately in the implementation details (Section 4).

											\subsection{Flow Vector Computation}
											
											Given a dataset of N trajectories, we represent its \math j^{th}$  trajectory with length \math L_{j}$ as an ordered sequence of points $(t^j_i, x^j_i, y^j_i)$ where $(x^j_i, y^j_i)$ denotes the location of \math j^{th}$ trajectory at time $t^j_i$ and  $i=1,2,\ldots,L_j $. Then its corresponding flow vector is a set of 4 dimensional vectors $\bold X^j_i=(x^j_i, y^j_i, x^j_{i+1}-x^j_i, y^j_{i+1}-y^j_i)=(x^j_i, y^j_i, u^j_i, v^j_i)$ where $i=1,2,\ldots,L_j-1 $. The first and second dimensions of the flow vector indicate its location while the third and forth dimensions indicate its velocity. After obtaining flow vectors from all trajectories in the dataset, we will have a vast set of 4 dimensional flow vectors ($\bold X=(x,y,u,v)$) that indicate magnitude ($\sqrt {{u}^2+{v}^2}$) and direction ($\mathit tan^{-1} (\frac {\mathit u}{\mathit v}) $) for the movement of each trajectory at all locations in the 2 dimensional surface (x and y). This data is noisy and we cannot directly use it to understand the behavioral movement of trajectories. Therefore, a Kmeans clustering algorithm is employed to group it into segments of spatially proximal flow vectors that have similar velocity properties. Such segmentation of the flow vector set provides us homogeneous regional segments that are representative and less noisy. We should note that, since in many cases trajectories are defined as sequences of points and the actual time stamp for each point is not available, only the order of the sequence is used in computing the velocity.
						\begin{figure*}[!ht]

							  \begin{center}
							  \subfigure[]{\label{clustering1}
							\includegraphics[width=0.18\textwidth]{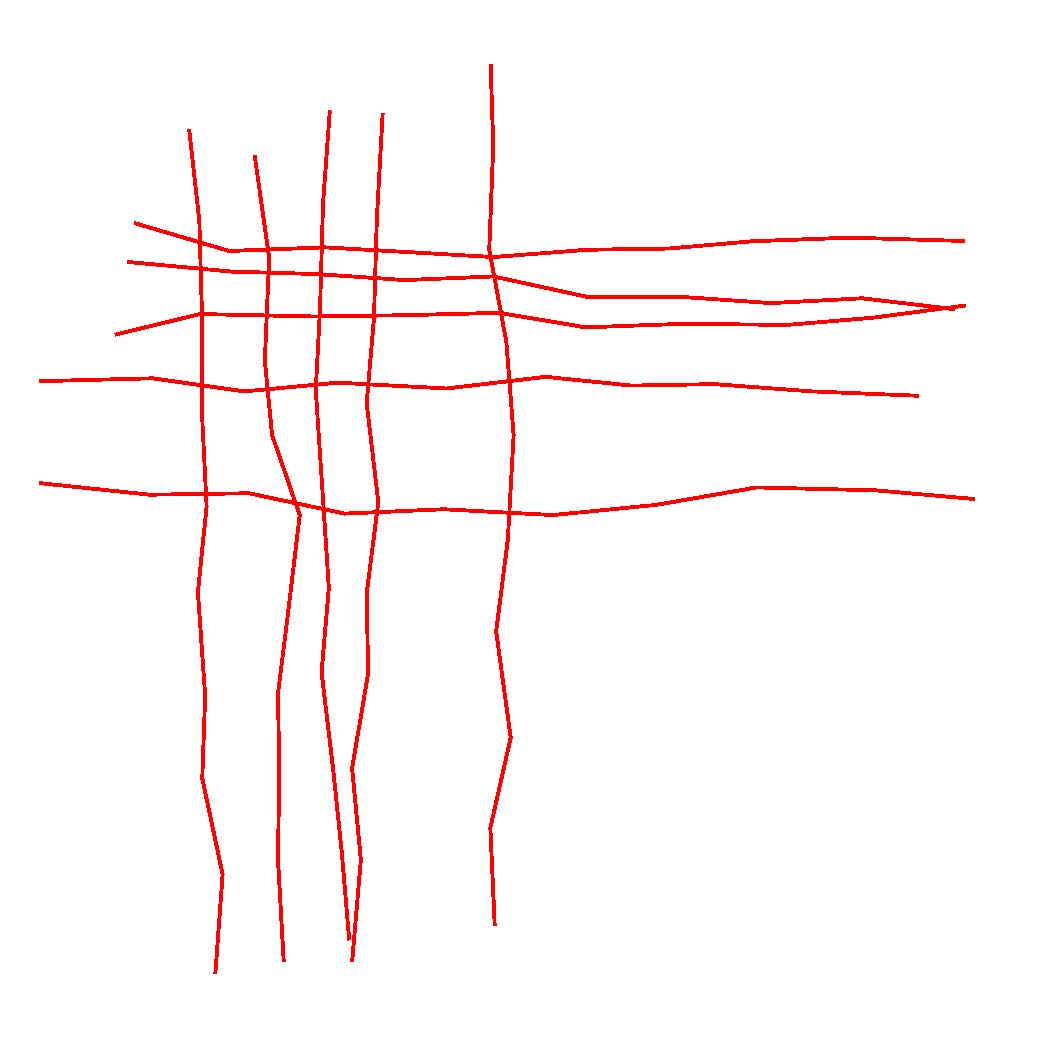}}
							  \subfigure[]{\label{clustering2}
							\includegraphics[width=0.18\textwidth]{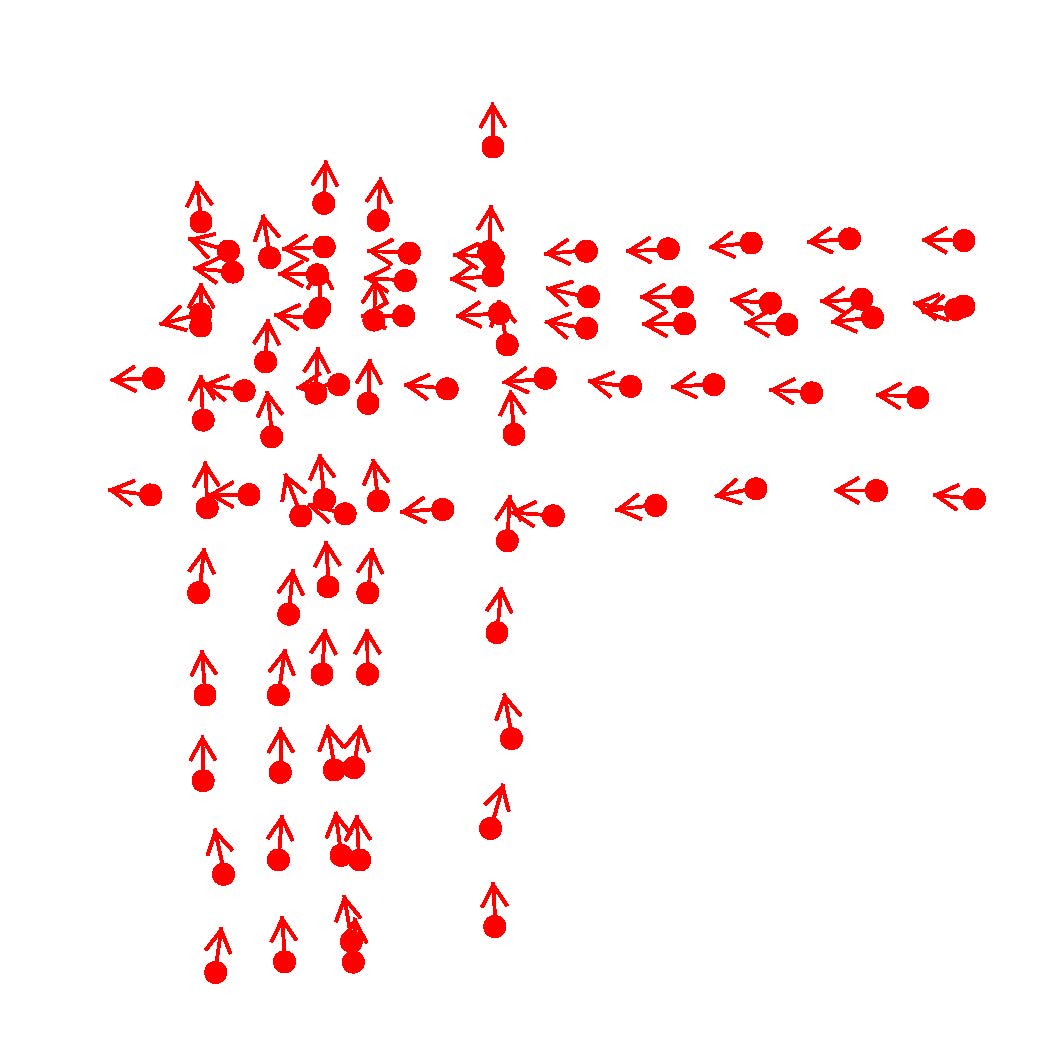}}
							  \subfigure[]{\label{clustering3}
							\includegraphics[width=0.18\textwidth]{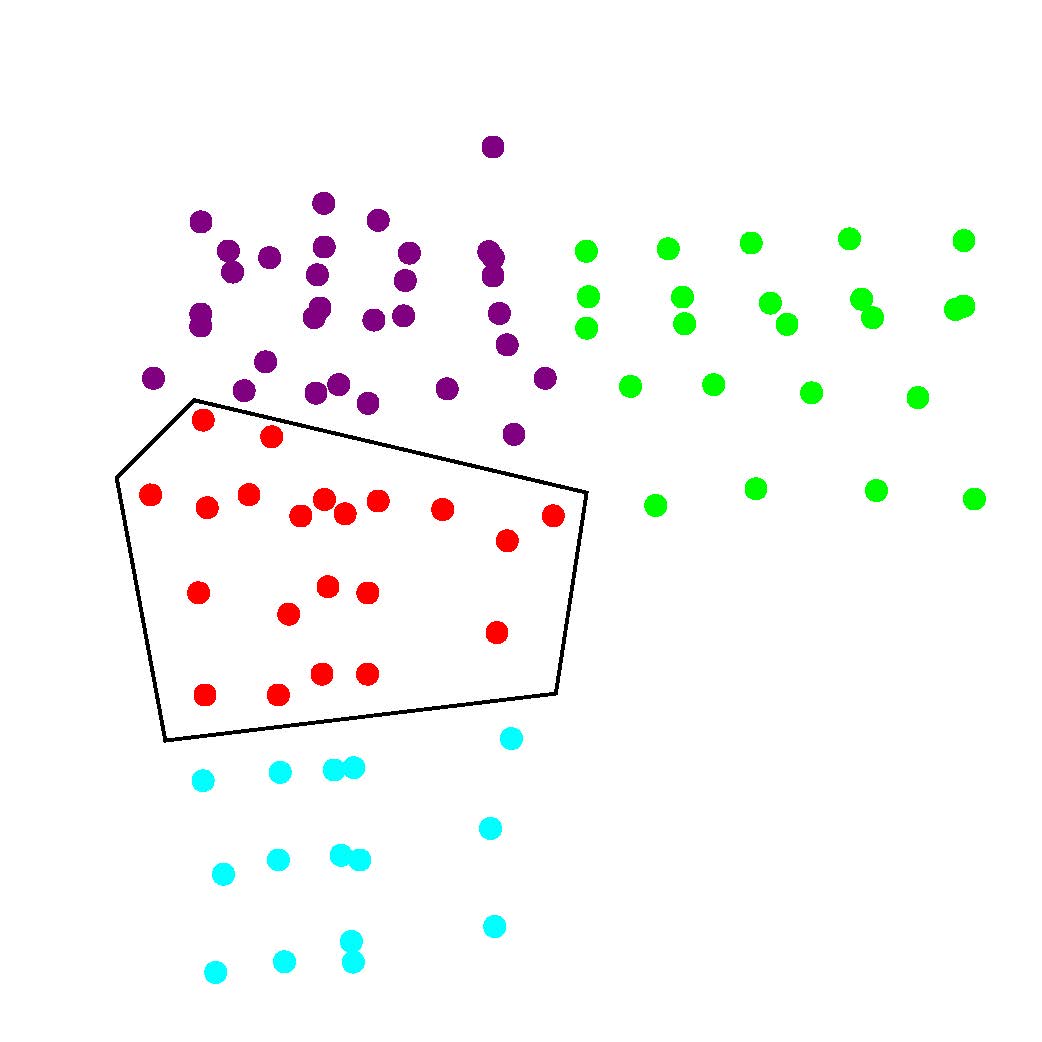}}
							  \subfigure[]{\label{clustering4}
							\includegraphics[width=0.18\textwidth]{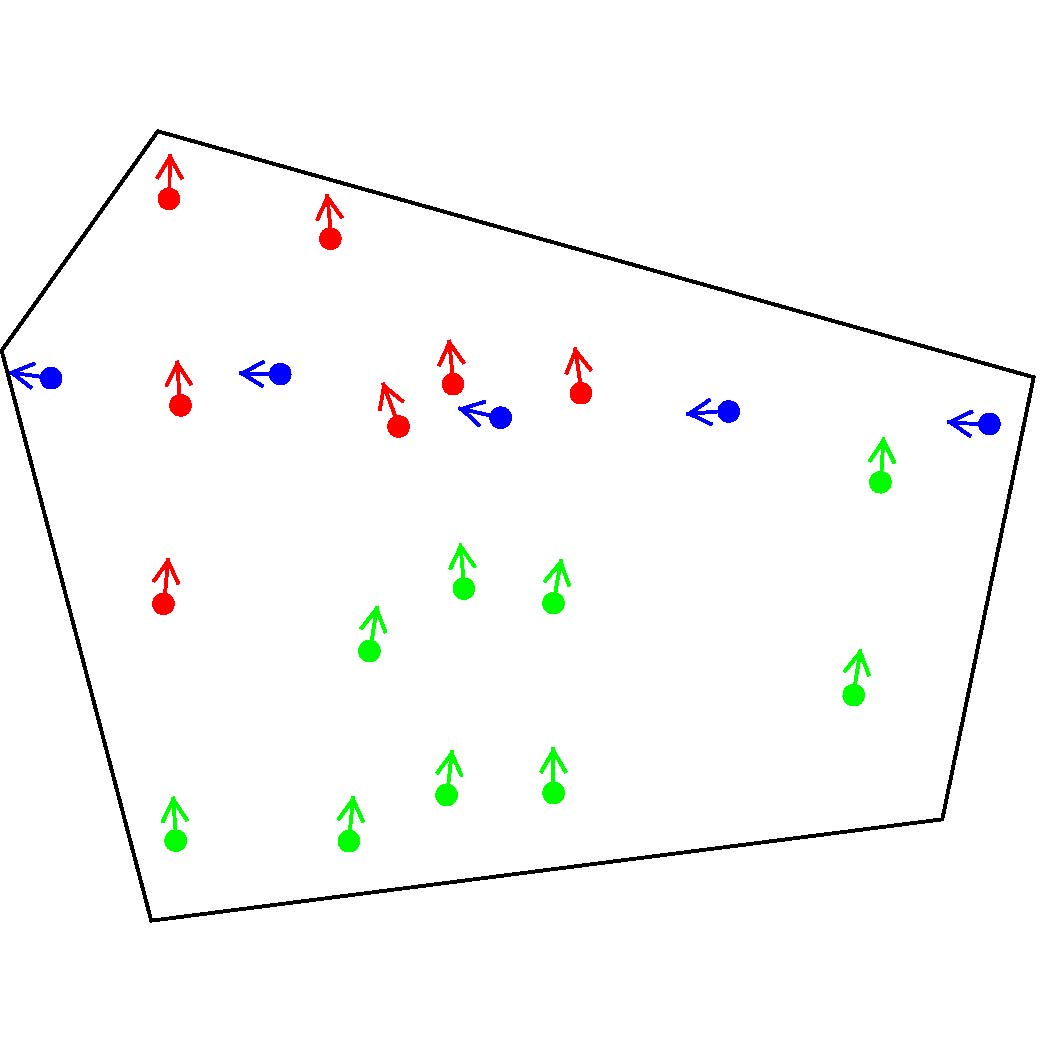}}
				 \subfigure[]{\label{clustering5}
							\includegraphics[width=0.18\textwidth]{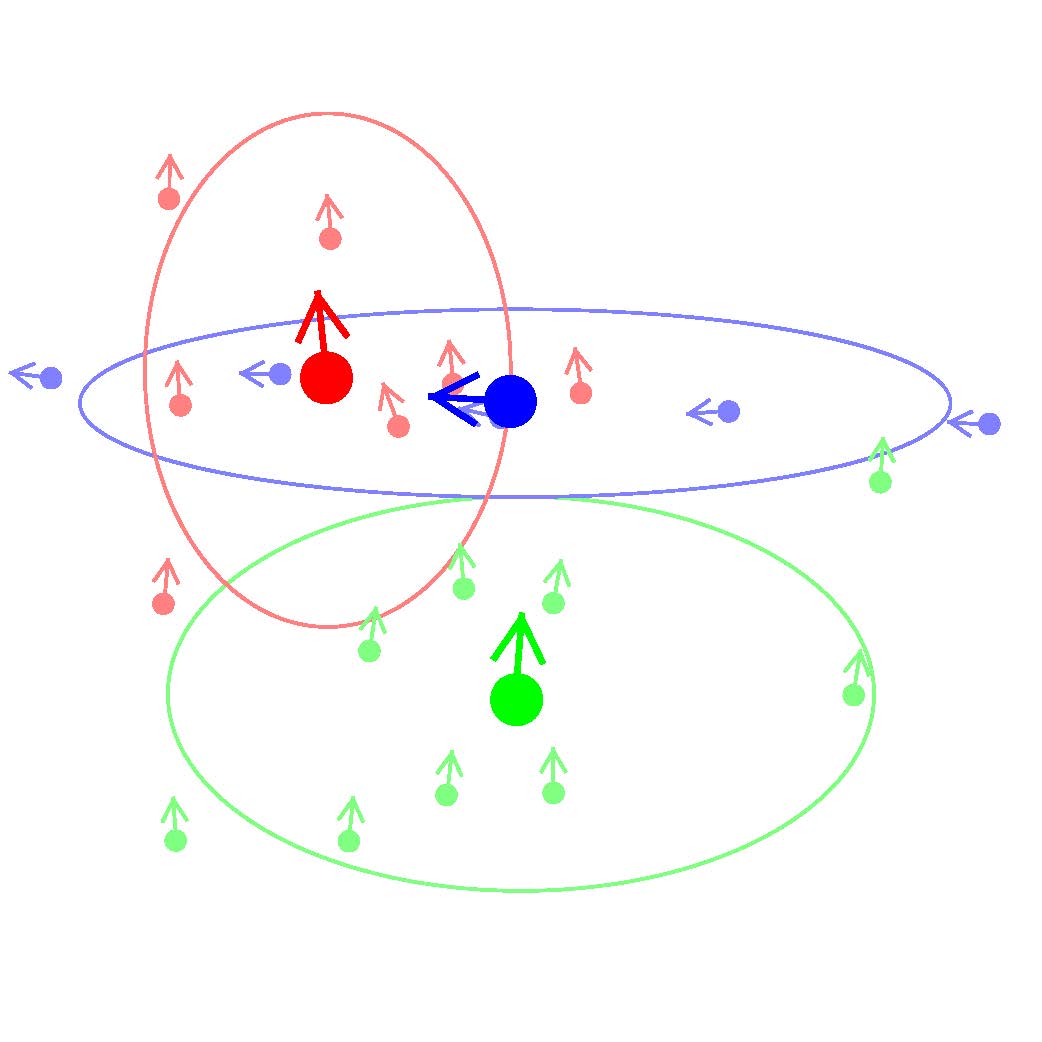}}
				 
							  \caption{\footnotesize{Given a trajectory set like \ref{clustering1}, we begin with computing flow vectors and obtain \ref{clustering2}. Then, a Kmeans clustering algorithm is employed to cluster the 4 dimensional flow vector set into homogeneous groups. To have a more clear visualization, we first show those flow vectors that are spatially proximal in different colors in \ref{clustering3}. Since the Kmeans clustering is in 4 dimensions, only those flow vectors that are similar in velocity in addition to being spatially proximal will get grouped together. Figure \ref{clustering4} shows flow vectors that are spatially proximal but differ in velocity properties (magnitude and orientation). The 4 dimensional Kmeans clustering provides clusters like those illustrated in \ref{clustering5} where flow vectors that are spatially proximal (shown in \ref{clustering3}) and have similar velocity (shown in \ref{clustering4}) are grouped together.}}
							\label{clustering_MC}
							 \end{center}
\vspace{-6mm}
							\end{figure*}

						\subsection{Motion Component Extraction}
											
											In order to generate the intermediate representation of the data that is less noisy and more suitable for the process of mining motion patterns, we cluster the flow vector set using Kmeans clustering where the number of clusters is set to K. Obtained clusters are the $\textit{motion components}$. The distance measure between \math p^{th}$ flow vector of  \math m^{th}$ trajectory and \math q^{th}$ flow vector of \math n^{th}$ trajectory used to create motion components is shown in Equation \ref{distPsOF}. Note that \math \beta$ defines the weight of velocity similarity versus spatial proximity, of flow vectors in formation of motion components and will be discussed in implementation details section.
											\begin{eqnarray} \label{distPsOF}
											 D(\bold X^m_p,\bold X^n_q)^2=({x^m_p-x^n_q})^2+({y^m_p-y^n_q})^2+\nonumber \\
					\beta({u^m_p-u^n_q})^2+\beta({v^m_p-v^n_q})^2
											\end{eqnarray}
											As the output of this step, we will have K motion components that are represented by their 4 dimensional means as $M_i \sim (\mu^{x}_i,\mu^{y}_i,\mu^{u}_i,\mu^{v}_i)$ where $i=1,2,\ldots,K$. We also denote the vector that spatially connects $m^{th}$ motion component to the $n^{th}$ one by $ \rho_{mn}= [(\mu^{x}_n-\mu^{x}_m),(\mu^{y}_n-\mu^{y}_m)]$. This definition will be used in the following section to determine the reachability of two motion components. Given a sample trajectory set, Figure \ref{clustering_MC} illustrates the steps which our method takes to extract motion components. We should note that unlike \cite{basharat2008learning} and \cite{saleemi2009} that aim at learning a mixture of probability density functions or a unified pdf, respectively, to model the underlying dynamics of the entire scene, we cluster the flow vectors only to generate intermediate local representations.

						\subsection{Motion Component Reachability Set Creation}
\begin{figure}[!ht]
\vspace{0mm}
							  \begin{center}

							  \subfigure[]{\label{wedge-shape}
							\includegraphics[width=0.2\textwidth]{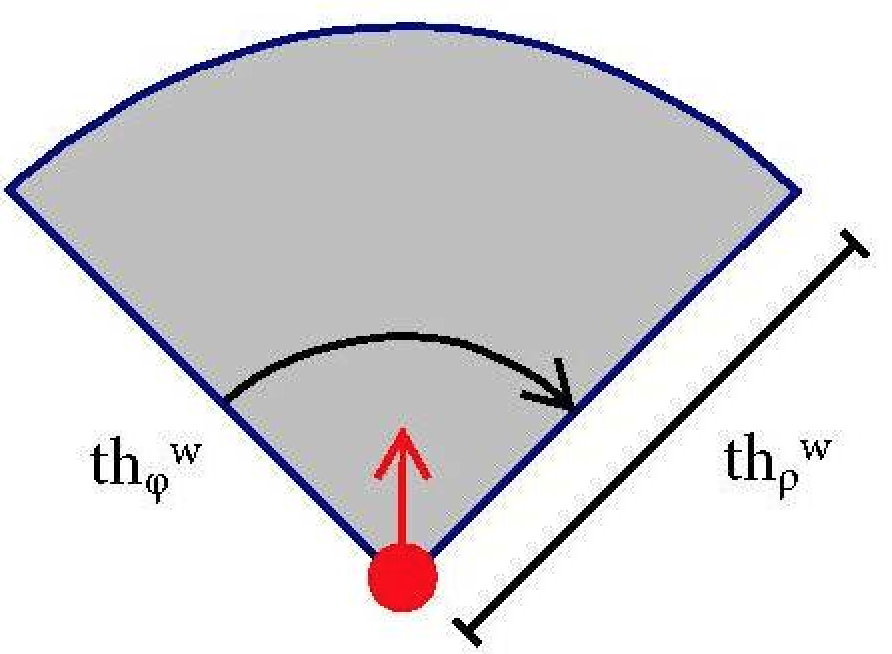}}
							  \subfigure[]{\label{doubleellipse-shape}
							\includegraphics[width=0.2\textwidth]{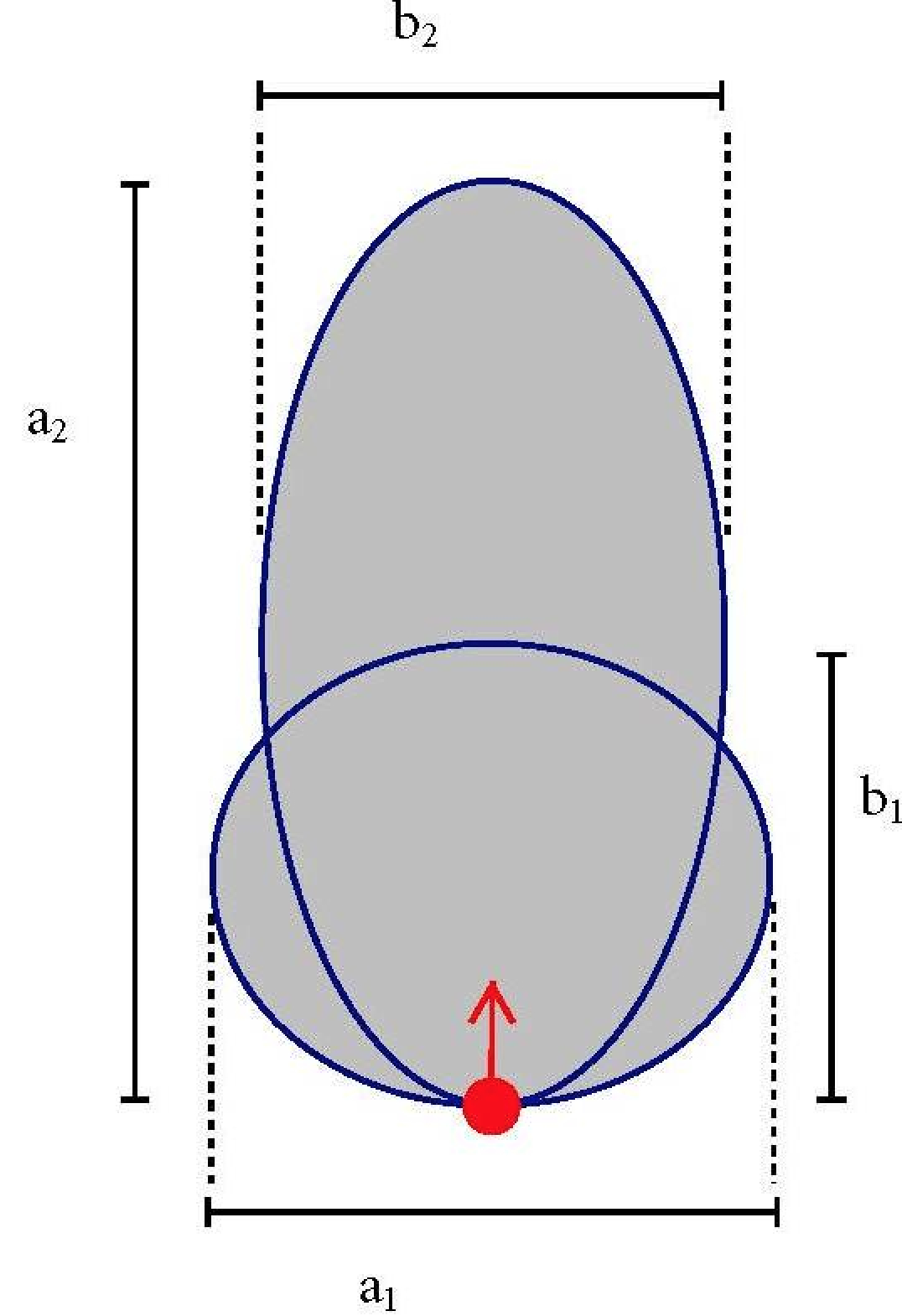}}
							  \subfigure[]{\label{twoComp}
							\includegraphics[width=0.3\textwidth]{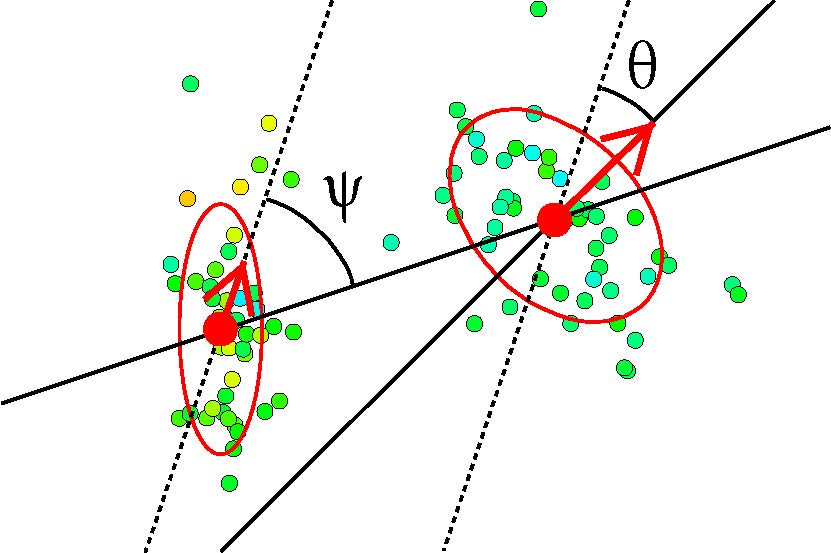}}
							  \caption{\footnotesize{Spatial realization of wedge and double-ellipse are illustrated in \ref{wedge-shape} and \ref{doubleellipse-shape}. \ref{twoComp} shows how we define $\theta$ and $\psi$.}}
							 \end{center}
\vspace{-8mm}
							\end{figure}

											We now look for $\it reachability$ between the motion components. Mathematically, reachability is an asymmetric relation between motion components. Intuitively, motion component A is $\it reachable$ from motion component B if a particle traveling with the initial motion and position prescribed by A could reasonably be expected to proceed to travel to B. By reachability, we mean direct reachability that does not require any intermediate motion components between A and B. However, this intuitive concept must be formalized and defined mathematically in order to be used. We formalize reachability as the conjunction of three conditions: $\bold {1)}$ The motion components A and B must be spatially located close together, $\bold {2)}$ The direction of flow vector of A should be similar to the direction of the vector $\rho_{AB}$, $\bold {3)}$ Direction of flow vectors for A and B should be similar. Intuitively, the second condition enforces flow vector of A to be aligned with shortest spatial path from A to B. To realize these conditions, we define the $\it proximity$, denoted by \math Pr$, of the \math m^{th}$ (as A) and the \math n^{th}$ (as B) motion components in Equation \ref{proximitydist} where $S_{mn}$ is the scale of the double-ellipse such that the \math n^{th}$ motion component falls on the boundary of the double-ellipse of the \math m^{th}$ motion component. Note that if $S_{mn}$ is smaller than 1, then the \math n^{th}$ motion component lies within the double-ellipse. Figures \ref{wedge-shape} and \ref{doubleellipse-shape}, respectively, show how the wedge and double-ellipse are located with respect to the \math m^{th}$ motion component. 

				\begin{figure*}[!ht]
							  \begin{center}
							  \subfigure[]{\label{pipeline1}
							\includegraphics[width=0.15\textwidth]{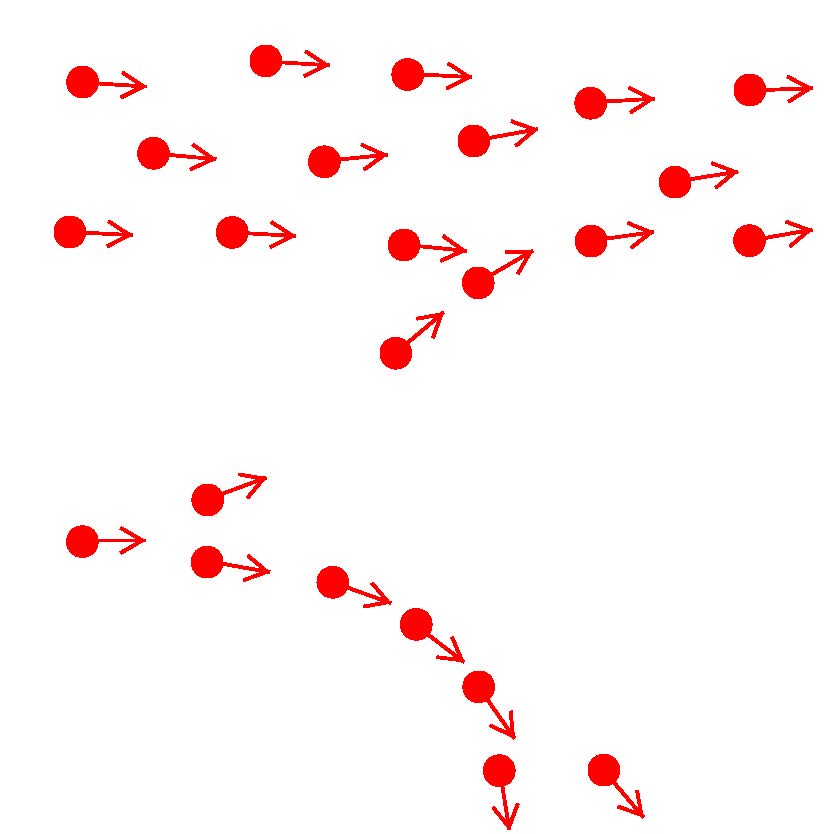}}
							  \subfigure[]{\label{pipeline2}
							\includegraphics[width=0.15\textwidth]{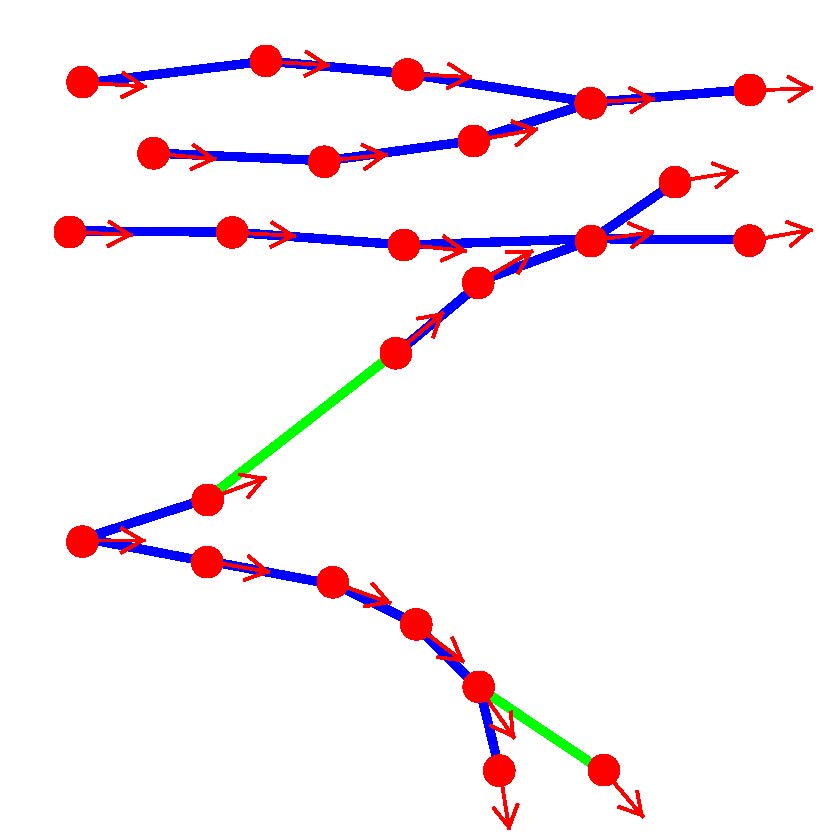}}
							  \subfigure[]{\label{pipeline3}
							\includegraphics[width=0.15\textwidth]{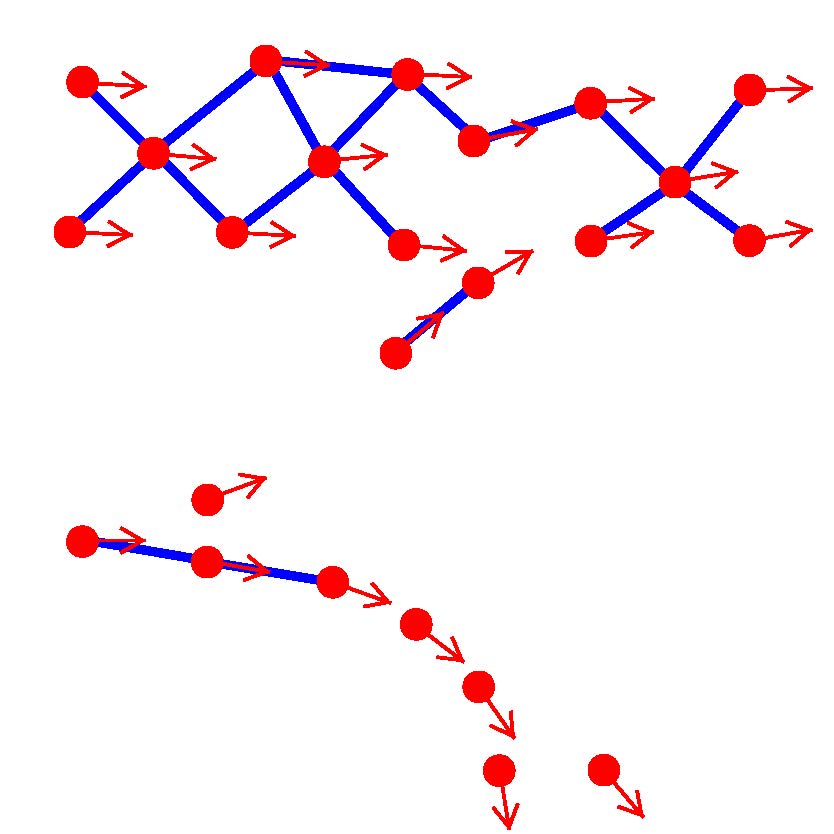}}
							  \subfigure[]{\label{pipeline4}
							\includegraphics[width=0.15\textwidth]{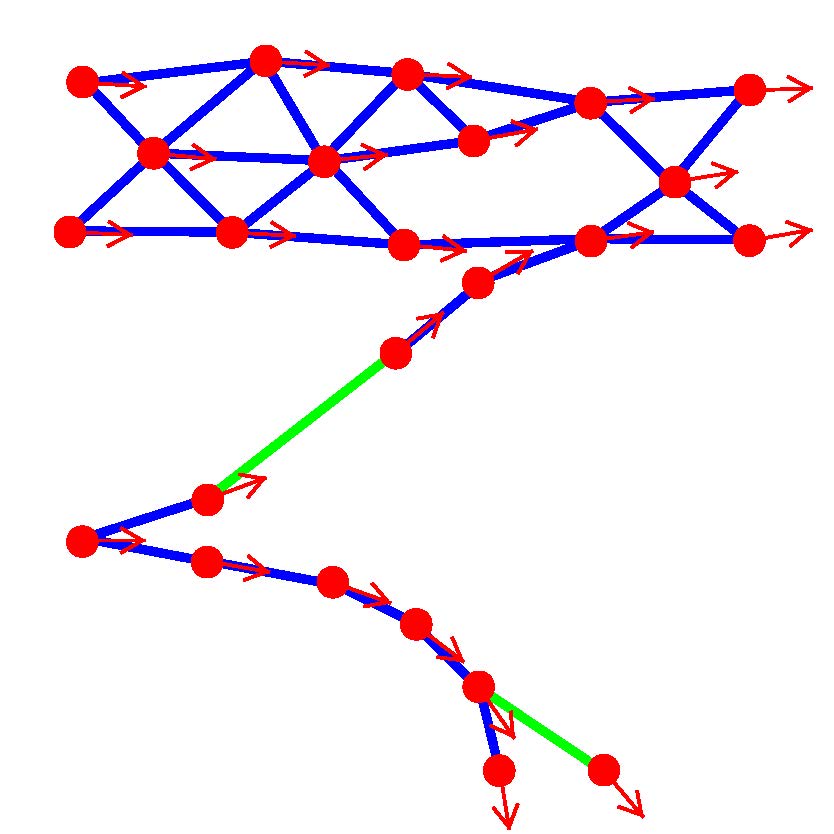}}
				 \subfigure[]{\label{pipeline5}
							\includegraphics[width=0.15\textwidth]{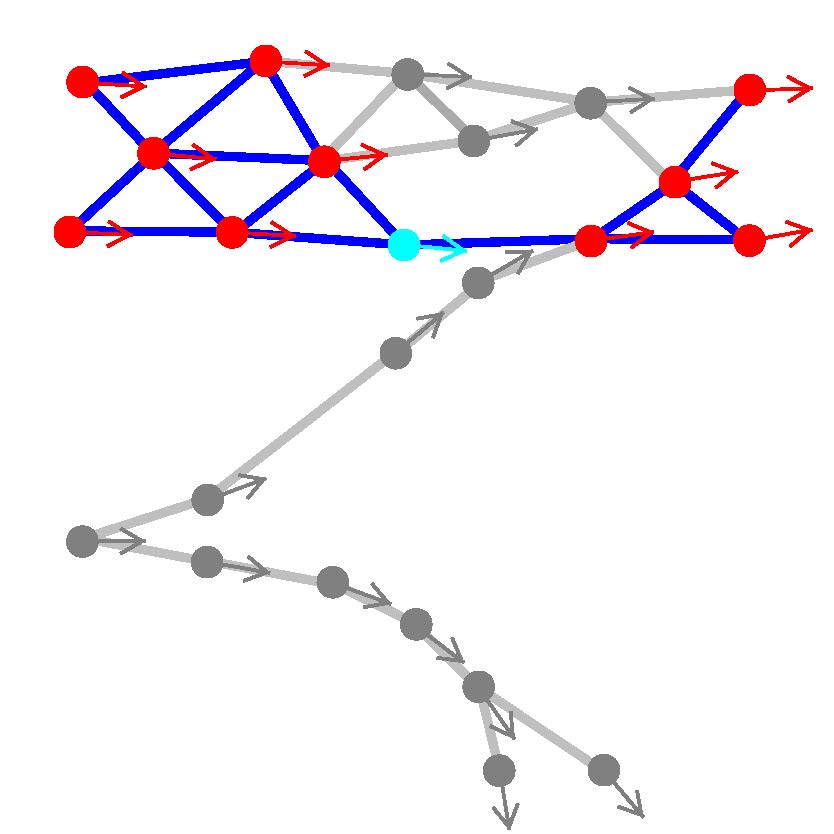}}
				 \subfigure[]{\label{pipeline6}
							\includegraphics[width=0.15\textwidth]{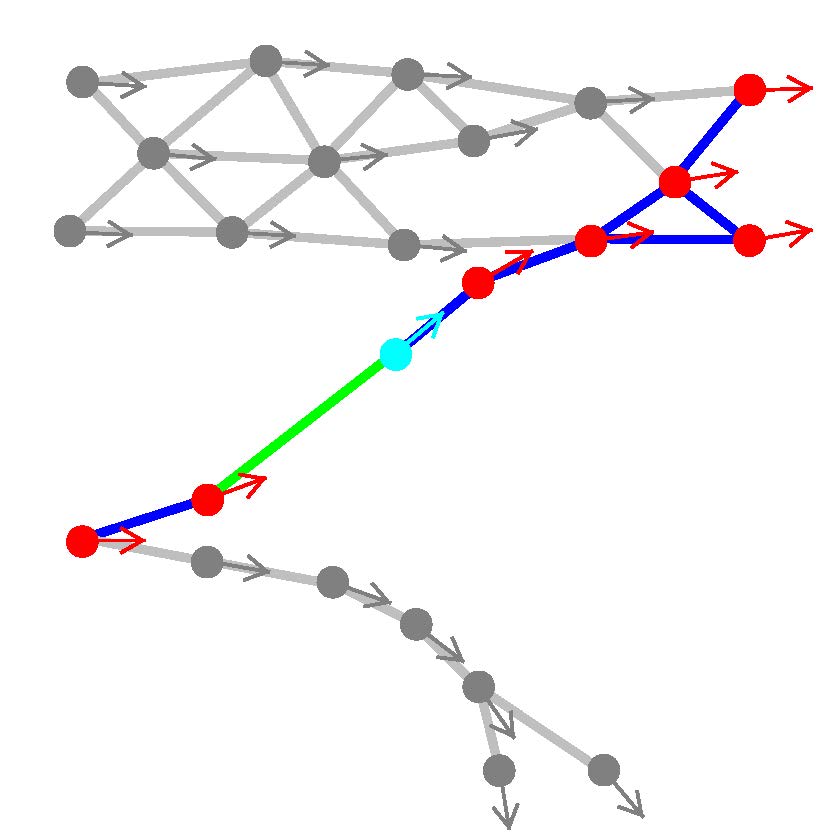}}
				 \subfigure[]{\label{pipeline7}
							\includegraphics[width=0.15\textwidth]{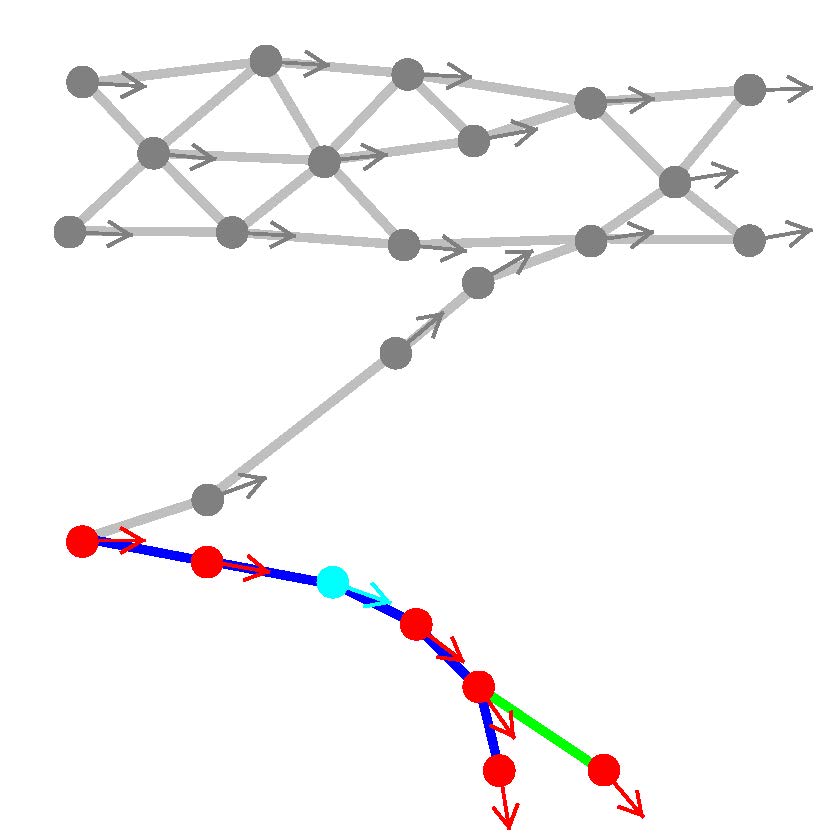}}
				 \subfigure[]{\label{pipeline8}
							\includegraphics[width=0.15\textwidth]{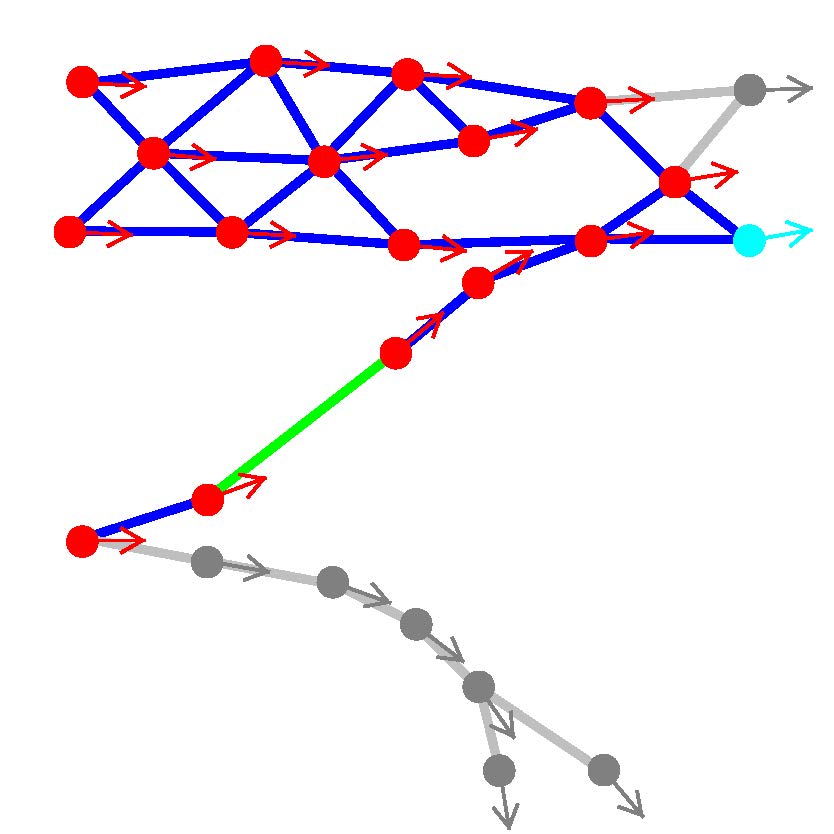}}
					                 \subfigure[]{\label{pipeline9}
							\includegraphics[width=0.15\textwidth]{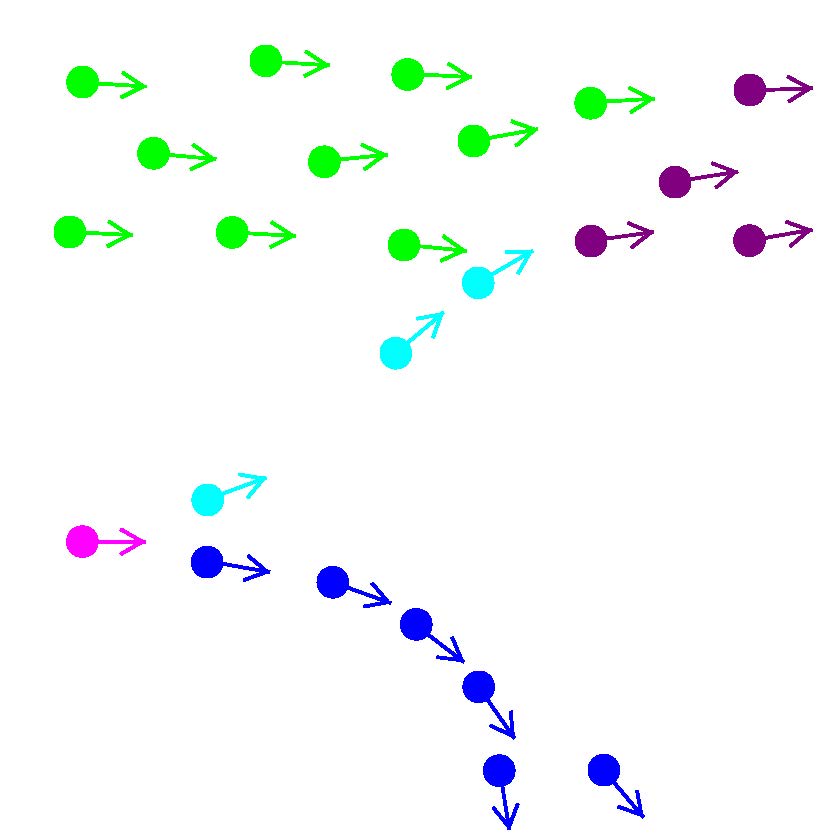}}
							  \caption{\footnotesize{\ref{pipeline1} shows motion components in red dots and the corresponding arrows refer to their flow direction. Using the double-ellipse and wedge conditions, reachable motion components are connected with blue lines respectively in \ref{pipeline2} and \ref{pipeline3}. Green lines are between pairs that happened to be reachable after the unblocking process. Although reachability is asymmetric while the lines are symmetric, the asymmetry should be clear from the flow direction of the motion components. \ref{pipeline4} illustrates all reachable pairs generated via the double-ellipse and wedge condition after applying the unblocking process. These pairs create the reachability set. Using the reachability set, the signature for each motion component is computed. \ref{pipeline5}, \ref{pipeline6}, \ref{pipeline7} and \ref{pipeline8} are four examples of the signature for the cyan motion components where the components not in the signature are grayed out. Finally, using the weighted Jaccard distance on motion components' signature, we form motion patterns. \ref{pipeline9} shows five motion patterns (indicated in green, violet, cyan, pink and blue) that our proposed method finds on this sample data.   
				}}
							\label{pipeline}
							 \end{center}
\vspace{-7mm}
							\end{figure*}
											\begin{equation}\label{proximitydist}
											Pr(M_m,M_n) = \left\{ 
											  \begin{array}{l l}
											    S_{mn} &   |\theta-\alpha\psi|<th_{\theta\psi}, |\theta|<th_{\theta}\\								
											inf &  o.w
											  \end{array} \right.
											\end{equation}
									$th_{\theta\psi}$ and $th_{\theta}$ are proximity parameters and will be discussed in detail in the implementation details section. Figure \ref{twoComp} illustrates how $\theta$ and $\psi$ are defined. Given the proximity measure in Equation \ref{proximitydist}, we define the \textit {reachability set} as $\left\{(M_{m},M_{n}) \big | \enskip Pr(M_m,M_n)<1\right\}$. Aiming to mine motion patterns, it is disadvantageous to have a motion component that cannot reach any other motion component (given $\it i$, $\forall{j}$ $:Pr(M_i,M_j)>1$) or no motion component is capable of reaching it  (given $\it i$, $\forall{j}$ $:Pr(M_j,M_i)>1$), a condition we refer to as a motion component being $\it blocked$. Such a situation would lead to breaking down motion patterns into smaller partitions, preventing us from extracting complete movement behaviors.
						In order to minimize blocked motion components, for each motion component $M_i$, we include in the reachability set the pair ($M_i,M_j$) where $Pr(M_i,M_j)$ is minimum with respect to j and less than \textit {search distance}. In the same fashion the pair ($M_z,M_i$) should be included. Finally, it is advantageous to be able to detect $\textit{short range semi-lateral}$ movements.  This property will allow us to merge similar but parallel motion 
									patterns together. In order to accommodate this, we include ($M_{m}$,$M_{n}$) pair in the reachability set if the $n^{th}$ motion component lies within a low radius-wide angle circular sector (\textit{wedge}) aligned with respect to $m^{th}$ motion component and their flow directions are similar. Formalizing the described condition, we include ($M_{m}$,$M_{n}$) pair in reachability set if $th^{w}_{\theta}, |\psi|<th^{w}_{\psi} $ and $ |\rho_{mn}|<th^{w}_{\rho}$.																
	We should note that in \cite{basharat2008learning}, \cite{saleemi2009} or any other method that learn the probability density function of the underlying dynamics of the entire data, reachability can be defined between any two points in the data space by employing approaches like MCMC or sequential sampling. On the other hand, methods like \cite{saleemi2010}, \cite{lasdas2012} and this work define reachability only between their intermediate representations. These intermediate representations are the GMM components, GP regression model of tracklets and motion components in aforementioned works, respectively. Our intermediate representations, motion components, are rough approximations of GMM components in \cite{saleemi2010} as we do not estimate the covariance matrices of the motion components.

						\subsection{Forming Motion Patterns}
We use two concepts of $\textit{path reachability}$ and $\textit{signature}$ to form motion patterns. The $i^{th}$ motion component is path reachable from the $j^{th}$ motion component if there is a chain of motion components from the $j^{th}$ to the $i^{th}$ motion component such that each $\textit {link}$ is in the reachability set. The signature of the $i^{th}$ motion component is the set of motion components that are either path reachable from the $i^{th}$ motion component or from which the $i^{th}$ motion component is path reachable. This is a novel representation that provides local and global properties of motion components simultaneously and therefore comparison between two motion components using their corresponding signatures would reflect both their local similarities and their contribution in global behaviors. Mathematically speaking, the reachability set can be shown as a directed graph while each node represents a motion component. An edge exists between two nodes if and only if the pair of motion components corresponding to those two nodes exists in the reachability set. Given a node, we can obtain the signature by applying depth first search on the graph and the reversed graph. Finally, the distance between two motion components is defined as the weighted Jaccard distance ($\it WJD$) between their associated signatures ($\it Sig$). The weighting is necessary because in practice, some areas have higher density of motion components (many motion components are located in a small region). This effect will skew our results since two motion components might be deemed similar by the unweighted Jaccard distance if their paths are quite different but they intersect in a region with many motion components. To counteract this effect, the Jaccard distance is weighted by a factor that assigns low values to the motion components that are located in dense regions and vice versa. The precise definition for the weighted Jaccard distance between the signatures of the $m^{th}$ and $n^{th}$ motion components is formalized in Equation \ref{WJD}. We extract $\textit{motion patterns}$ via agglomerative clustering with a distance cutoff, single linkage and WJD as distance metric. It is worth mentioning that when single linkage is used, the results will be the same as clustering via thresholding and forming weakly connected components. Despite this, our method is different from \cite{saleemi2010} as we use the weighted Jaccard distance between motion components' signatures as the distance metric. \cite{saleemi2010} forms motion patterns by finding weakly connected components on a graph that represents the reachability between GMM components where only local similarity of motion components is considered. Therefore, it would group globally different motion behaviors together if they share a common GMM component (note that GMM component in \cite{saleemi2010} have actual time information), a scenario that occurs in merging or diverging of motion patterns. \begin{eqnarray}
							\label{WJD1}\textit {Q}_\cap(m,n) =\{\textit {Sig}(M_m)\cap \textit {Sig}(M_n)\} \\
							\label{WJD2}\textit {Q}_\cup(m,n) =\{\textit {Sig}(M_m)\cup \textit {Sig}(M_n)\} \\
							\label{WJD}
											 \textit {WJD}_n^m =\frac{\sum_{i\in\textit {Q}_\cap(m,n)} (w_0+\sum_{j=1}^{K} e^{(-\frac{||\rho_{ij}||^{2}_{2}}{2\sigma^2})})^{-1}}
										{\sum_{i\in\textit {Q}_\cup(m,n)}(w_0+ \sum_{j=1}^{K} e^{(-\frac{||\rho_{ij}||^{2}_{2}}{2\sigma^2})})^{-1}}
											\end{eqnarray}												Given a sample set of motion components, Figure \ref{pipeline} illustrates the steps which our method takes to form motion patterns in a trajectory dataset.

						\section {Experimental Results}
										In this section, we first give a brief description of the datasets that were used for evaluation. Then, we show experimental results of our proposed trajectory clustering method on different datasets and finally compare them to the baseline methods. Due to the space limitation, the model parameters used to generate outputs associated with different datasets are provided in supplemental material.

										\subsection{Datasets}
	
										We used five different datasets to test the proposed method. These datasets are among the ones that are used by many papers in the literature and vary greatly in their properties, such as the number of trajectories, average number of points per trajectory, sampling density, spatial separation and complexity. Experimental results indicate that our proposed method is an effective solution for the trajectory clustering task regardless of the dataset properties.			
										\begin{itemize}
										
										\item {\textbf {Vehicle Motion Trajectory Dataset:} This dataset contains 1500 trajectories gathered by tracking vehicles at a traffic intersection. These trajectories are annotated manually; each trajectory is assigned to one of 15 trajectory classes. The mean number of points per trajectory is 96. This dataset is available at \cite{VMTD}}. 
										
										\item {\textbf {Atlantic Hurricane Dataset (HURDAT2):} This dataset is provided by the National Hurricane Service (NHS) and contains 1740 trajectories of Atlantic Hurricanes from 1851 through 2012, with trajectories containing 27 points on average. NHS also provides annotations of typical hurricane tracks for each month throughout the annual hurricane season that spans from June to November. In order to evaluate how close the motion patterns mined by our method are to the NHS annotations, we divided the Atlantic Hurricane Dataset into six subsets, one for each month. Trajectories that span more than one month were split to ensure each month includes only activity occurring within its span. This dataset is available at  \cite{AtlanticHurricanes}}.
										
										\item {\textbf {Swainson's Hawks Dataset:} This dataset contains 43 trajectories that trace the migration of Swainson's hawks. A description of the hawks' migration paths is provided in \cite{kochert2011migration}, which states that the hawks converge on the Gulf of Mexico coast, travel southward following a narrow path across the Andes in Colombia, then proceed along the east side of the Andes to central Argentina, where they spend the austral summer before returning north using largely the same route. The average number of points per trajectory in this dataset is 105. This dataset is available at \cite{Movebank}}.
										
										\item {\textbf {The Greek Trucks Dataset:} This dataset contains 1100 trajectories from 50 different trucks delivering concrete around Athens, Greece. As expected, the trucks follow highways giving the trajectories a distinctive appearance. The average number of points per trajectory is 86. This dataset is available at \cite{GTrucksData}}.
										
						\item {\textbf {The NGSIM Lankershim Dataset:} This dataset contains detailed vehicle trajectory data on Lankershim Boulevard in the Universal City neighborhood of Los Angles, CA on June 16, 2005. The dataset corresponds to two 15-minutes periods of 8:30 am to 8:45 am and 8:45 am to 9:00 am obtained by five video cameras. For our experiments, we extracted portions of trajectories captured by camera NO. 2 in 8:30 am to 8:45 am period as it is covering the busiest intersection in the dataset. This subset contains 1095 trajectories and the average number of points per trajectory is 305. The full Lankershim dataset and its annotations are available at \cite{lankershim}.}
										\end{itemize}	
\vspace{-2mm}
						\subsection{Pre-Processing}
							For all experiments, data was first normalized so that all the trajectories were in the bounding box determined by $x\in {[0,1000]}$ and $y\in {[0,1000]}$. The Atlantic Hurricane Dataset went through further pre-processing: after splitting the dataset by months, each of the resulting subsets was pruned by removing trajectories consisting solely of a single coordinate pair repeated one or more times.

						\subsection{Intrepreting Output}
						
\begin{figure}[!ht]
\vspace{-6mm}
							  \begin{center}
							  \subfigure[]{\label{mergedivergeOut}
							\includegraphics[width=0.08\textwidth]{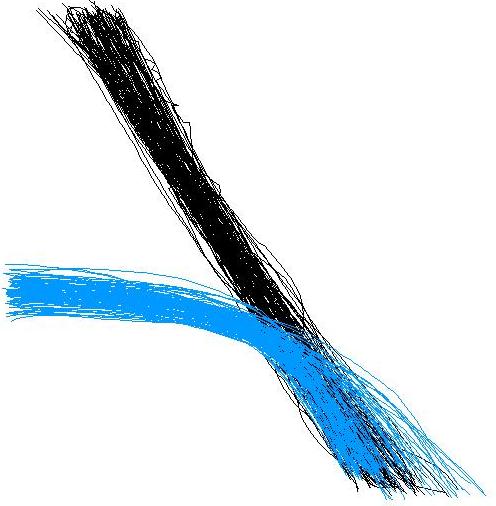}}
							  \subfigure[]{\label{mergedivergeOut1}
							\includegraphics[width=0.08\textwidth]{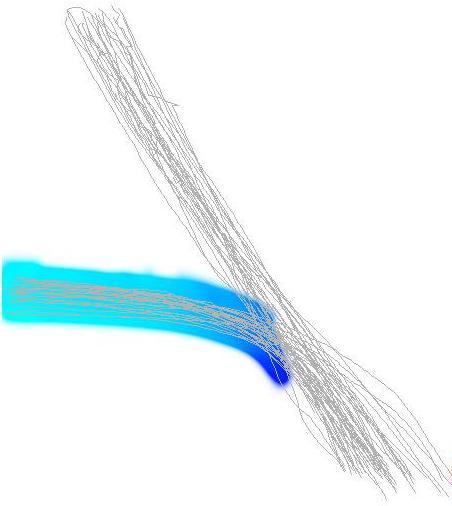}}
							  \subfigure[]{\label{mergedivergeOut2}
							\includegraphics[width=0.08\textwidth]{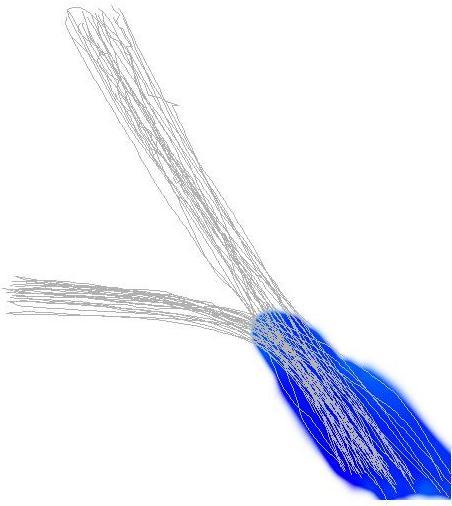}}
							  \subfigure[]{\label{mergedivergeOut3}
							\includegraphics[width=0.08\textwidth]{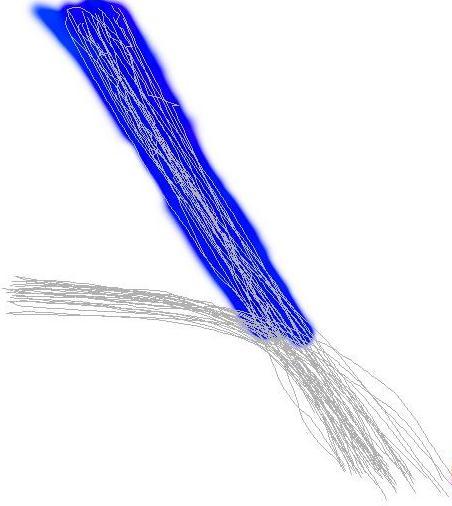}}
					                 \subfigure[]{\label{colorwheel1}
							\includegraphics[width=0.03\textwidth]{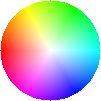}}
							  \caption{\footnotesize{An example for merging in \ref{mergedivergeOut} where the proposed method has found two separate motion patterns of \ref{mergedivergeOut1} and \ref{mergedivergeOut3} in addition to the common part \ref{mergedivergeOut2}. In \ref{mergedivergeOut}, colors of black and blue are used only to distinguish two classes of trajectories according to the annotations. Colorwheel is shown in \ref{colorwheel1}.}}
							 \end{center}
\vspace{-1mm}
							\end{figure}

						Before we discuss experimental results, we first briefly explain how to read and interpret the output of the proposed algorithm. Each discovered motion pattern is displayed on top of plotted trajectories. To avoid overcrowding the figures, only a random subset of trajectories are visualized. The color of the motion pattern denotes the direction of the motion. A color wheel that appears in each figure serves as the legend for translating color into direction of motion. Lastly, we must introduce the reasoning behind our handling of merging and diverging trajectories. An example of such a scenario is illustrated in Figure \ref{mergedivergeOut}, where two clusters of trajectories partially overlap. Our proposed method recovers three motion patterns in this case: two separate motion patterns for the distinct portions of the two clusters and a separate motion pattern for their merged portion. Therefore, it must be noted that a single trajectory may pass through several motion patterns. Using our proposed method, trajectories shown in Figure \ref{mergedivergeOut} produce motion patterns shown in Figures \ref{mergedivergeOut1}, \ref{mergedivergeOut2} and \ref{mergedivergeOut3}.
							
\vspace{-6mm}
						\subsection{Evaluations on Vehicle Motion Trajectory Dataset}
\begin{figure*}[!ht]

							  \begin{center}
 \subfigure[]{\label{vehicleGT}
							\includegraphics[width=0.11\textwidth]{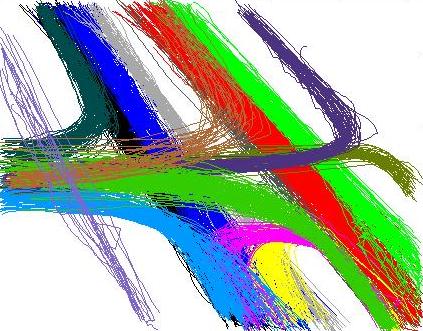}}
							  \subfigure[]{\label{vehicleGT1}
							\includegraphics[width=0.11\textwidth]{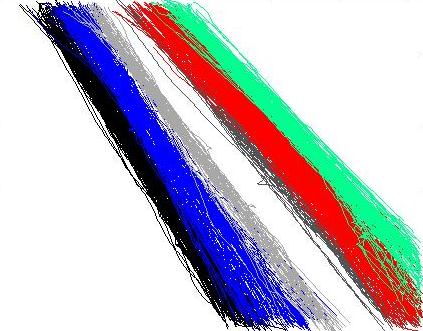}}
							  \subfigure[]{\label{vehicleGT2}
							\includegraphics[width=0.11\textwidth]{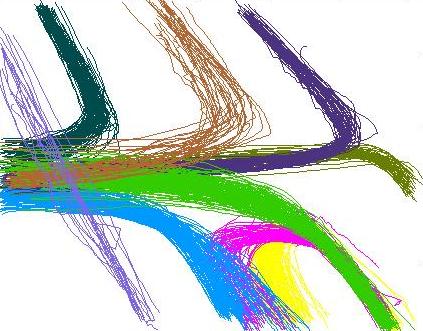}}
							  \subfigure[]{\label{intersection1}
							\includegraphics[width=0.11\textwidth]{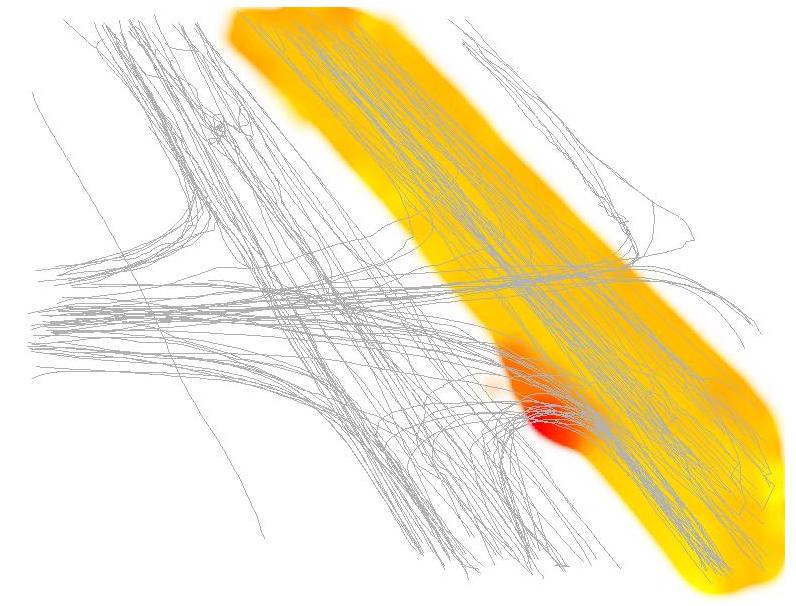}}
							  \subfigure[]{\label{intersection2}
							\includegraphics[width=0.11\textwidth]{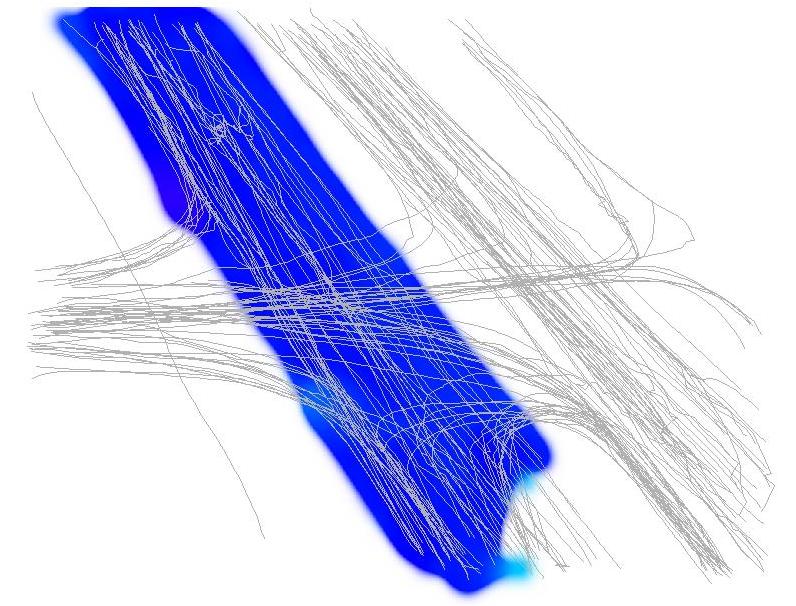}}
							  \subfigure[]{\label{intersection3}
							\includegraphics[width=0.11\textwidth]{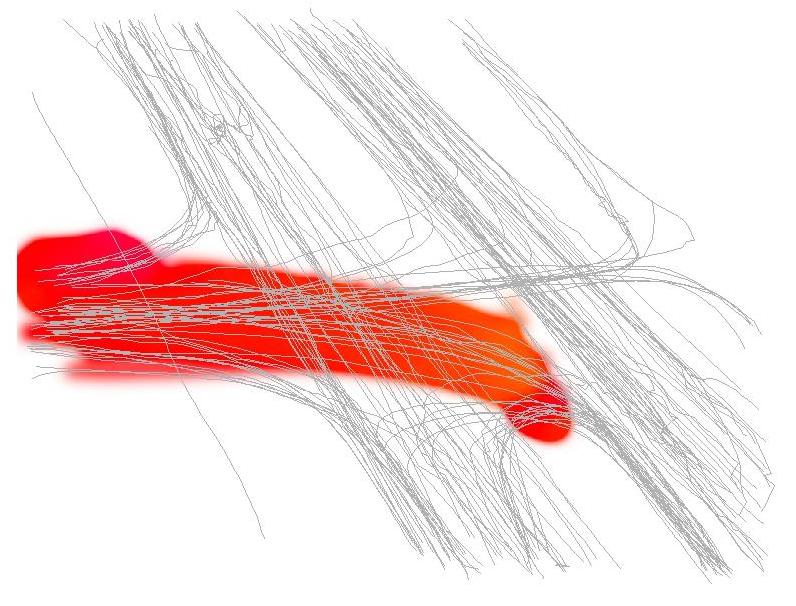}}
							 \subfigure[]{\label{intersection4}
							\includegraphics[width=0.11\textwidth]{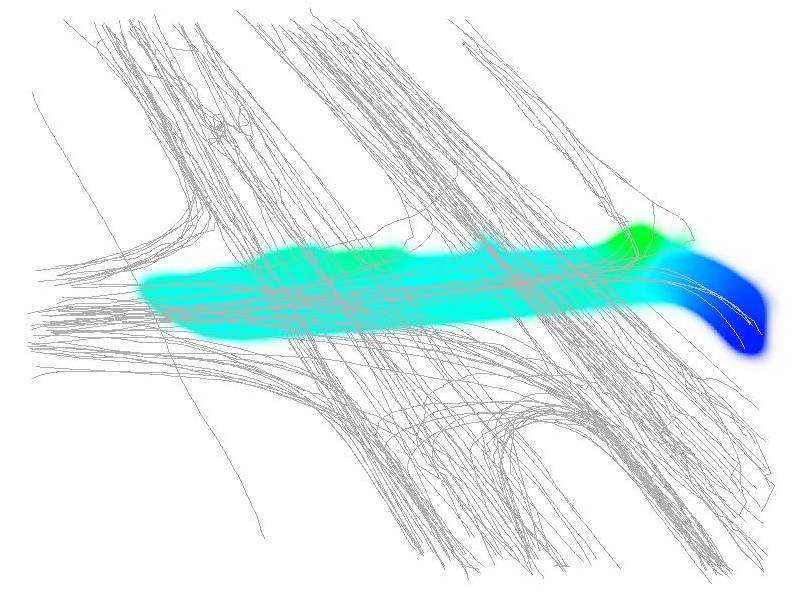}}
							  \subfigure[]{\label{intersection5}
							\includegraphics[width=0.11\textwidth]{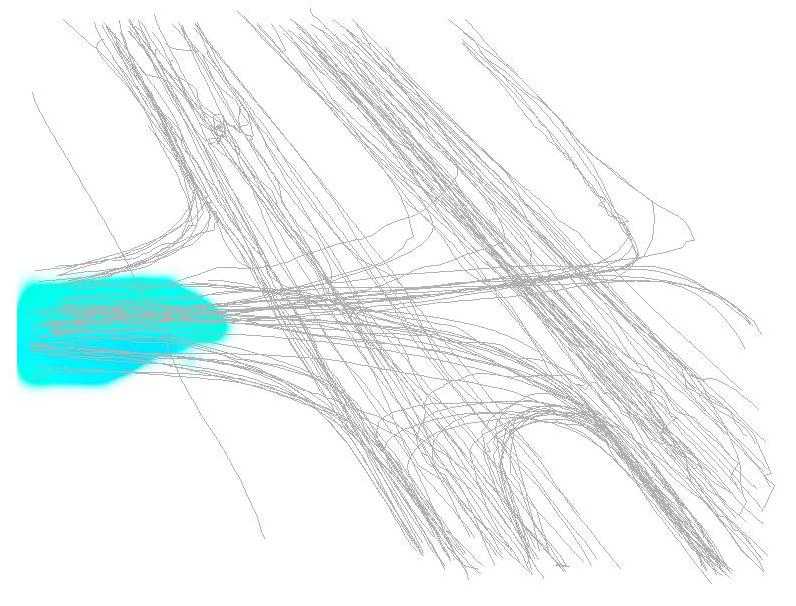}}
							  \subfigure[]{\label{intersection6}
							\includegraphics[width=0.11\textwidth]{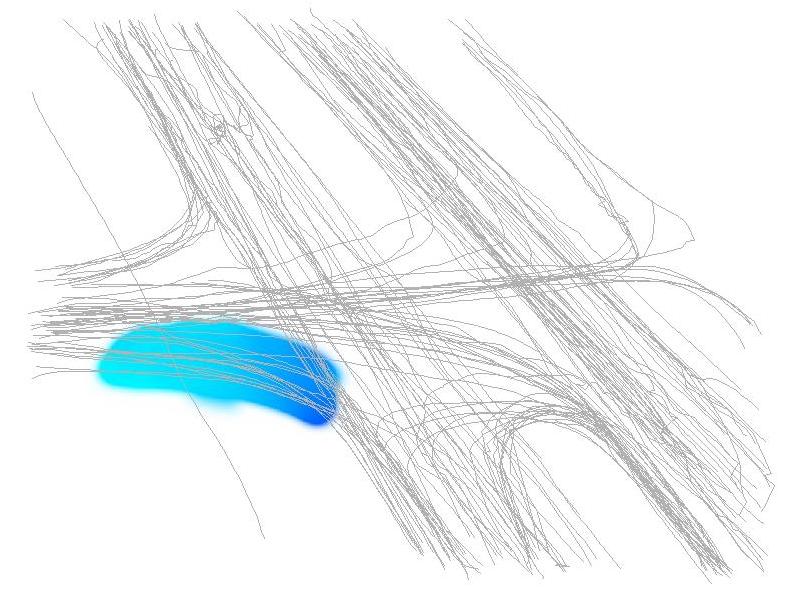}}
							 \subfigure[]{\label{intersection7}
							\includegraphics[width=0.11\textwidth]{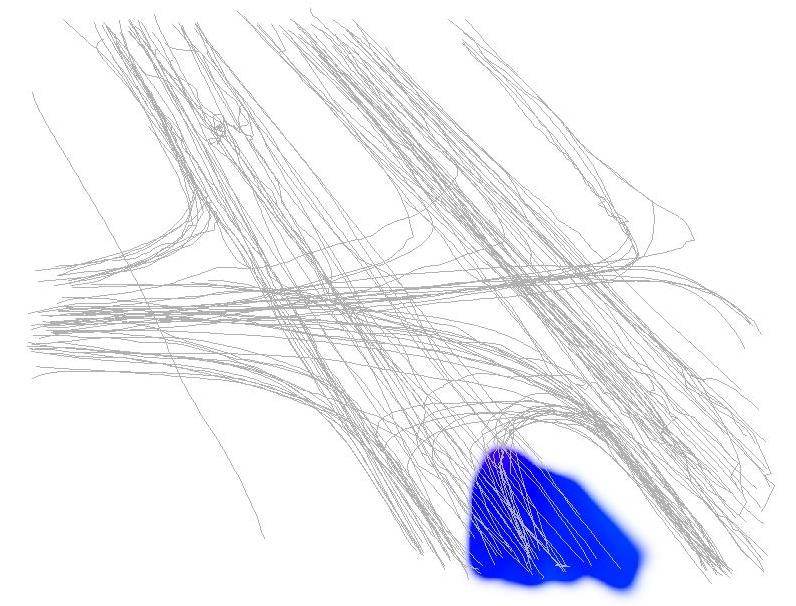}}
							  \subfigure[]{\label{intersection8}
							\includegraphics[width=0.11\textwidth]{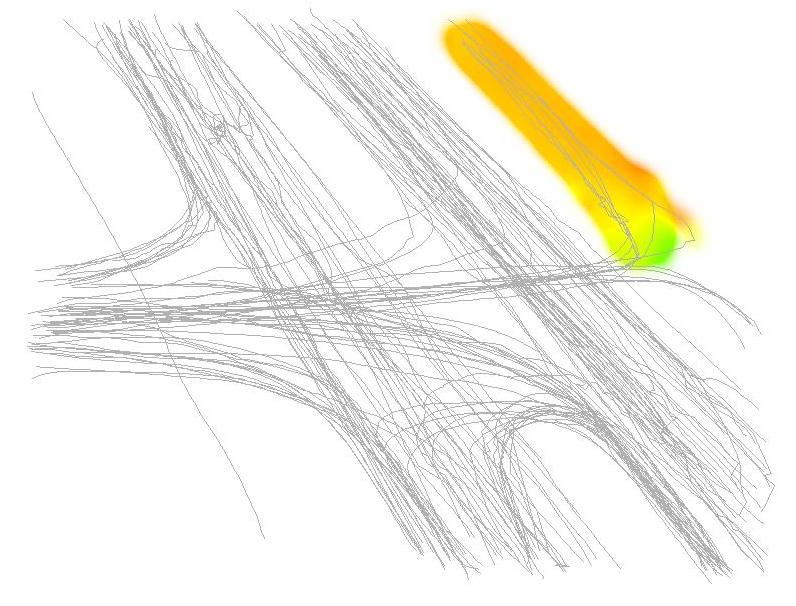}}
							  \subfigure[]{\label{intersection9}
							\includegraphics[width=0.11\textwidth]{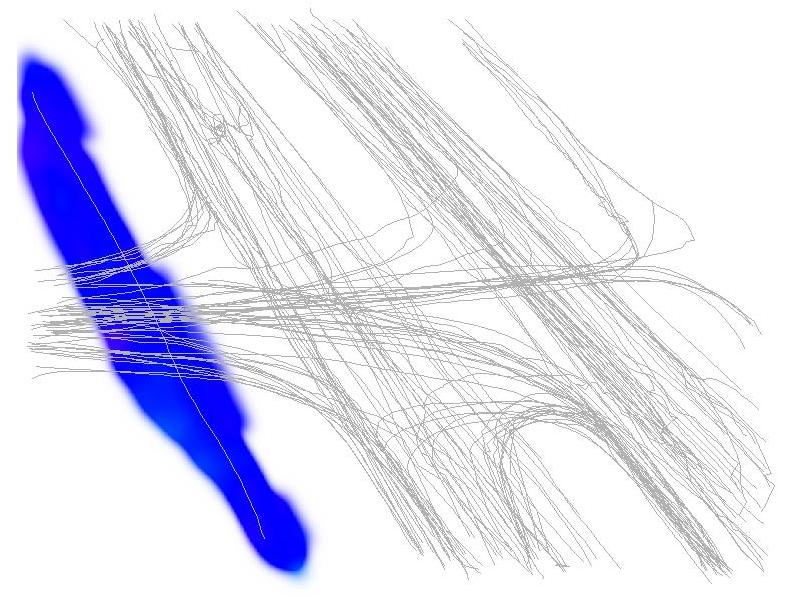}}
							 \subfigure[]{\label{intersection10}
							\includegraphics[width=0.11\textwidth]{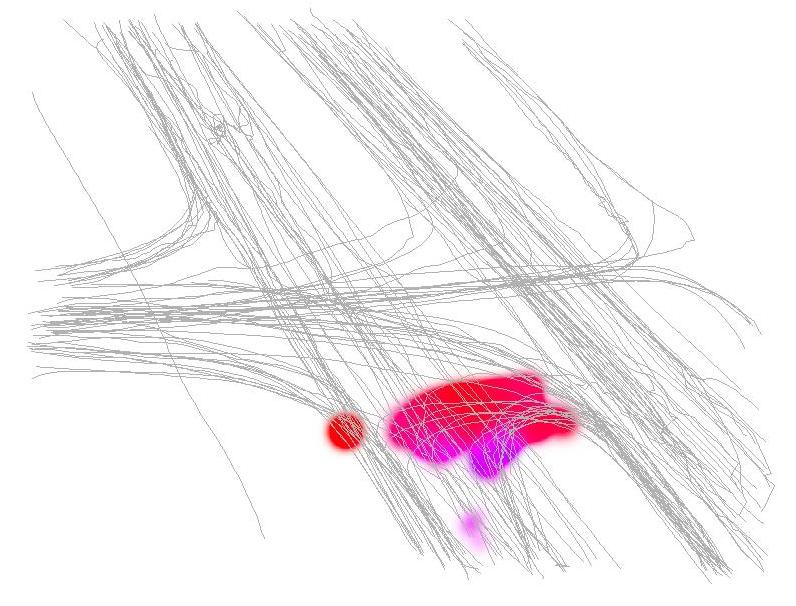}}
							  \subfigure[]{\label{intersection11}
							\includegraphics[width=0.11\textwidth]{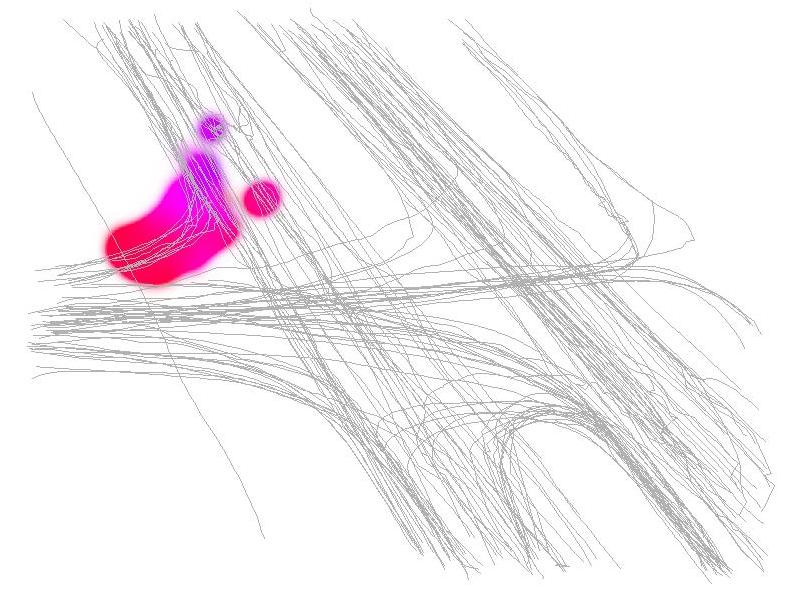}}
							  \subfigure[]{\label{intersection12}
							\includegraphics[width=0.11\textwidth]{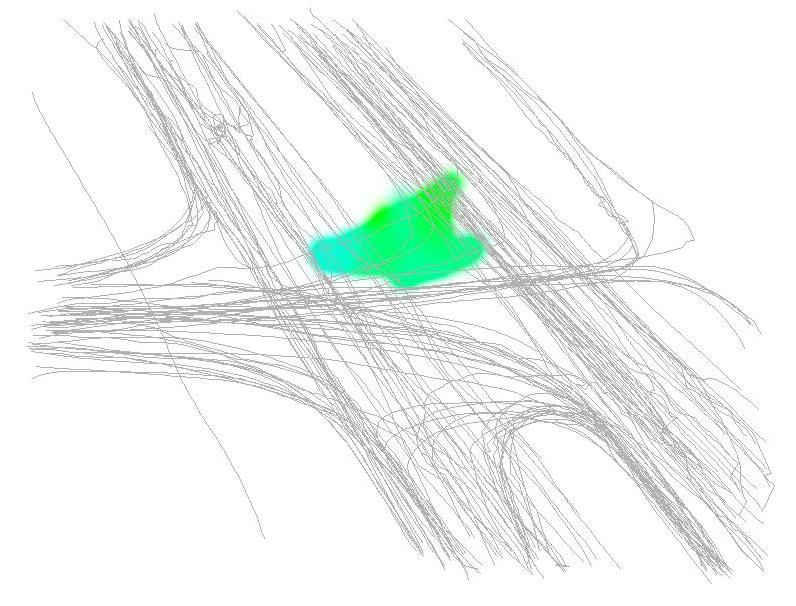}}
				\subfigure[]{\label{colorwheel2}
							\includegraphics[width=0.03\textwidth]{figures/colorwheel.jpg}}
							  \caption{\footnotesize{15 different annotated classes of trajectories in Vehicle Motion Trajectory Dataset are shown in \ref{vehicleGT} with different colors. To have a more clear visualization, we illustrate classes belong to main highways in \ref{vehicleGT1} and the rest in \ref{vehicleGT2}. Extracted motion patterns from Vehicle Motion Trajectory Dataset are illustrated in \ref{intersection1} to \ref{intersection12}. Colorwheel is shown in \ref{colorwheel2}.}}
							\label{intersectOutput}
							 \end{center}
\vspace{-5mm}
							\end{figure*}

							Examining the outputs of the proposed method versus annotations shown in Figure \ref{vehicleGT}, we see that the proposed algorithm recovers the motion patterns present in the dataset, with a few differences from the annotated version. First of all, the annotations categorize the traffic in each of the parallel lanes in Figure \ref{vehicleGT1} into separate clusters. Because the motion in the left group of three lanes is very similar, the proposed algorithm recovers it as a single motion pattern. The same applies to the three-lane group on the right. Second, for reasons stated above, the turning trajectories shown in Figure \ref{vehicleGT2} are segmented to differentiate their distinct portions. 
							Figures \ref{intersection1} and \ref{intersection2} correspond to the two directions of traffic on the main highways. Figure \ref{intersection3} corresponds to one of the left turns and is especially large since it contains the bulk of the left-side outgoing traffic. Figures \ref{intersection4}, \ref{intersection5}, and \ref{intersection6} correspond to the incoming traffic from the left side. This is an example of the diverging behavior where the motion pattern from Figure \ref{intersection5} diverges into the motion pattern turning right in Figure \ref{intersection6} and the motion pattern going straight in Figure \ref{intersection4}. Motion patterns in Figures \ref{intersection7} and \ref{intersection10} reflect a U-turn, where the former is a portion of the U-turn that merges with the traffic shown in Figure \ref{intersection2} and the latter is the distinct part of the U-turn. Figures \ref{intersection8} and \ref{intersection9} show two different access roads. Figure \ref{intersection11} shows a right turn and Figure \ref{intersection12} shows a rarely taken left turn. 

								\subsection{Evaluations on Atlantic Hurricane Dataset}
\begin{figure}[!ht]

							  \begin{center}
							  \subfigure[]{\label{june}
							\includegraphics[width=0.15\textwidth]{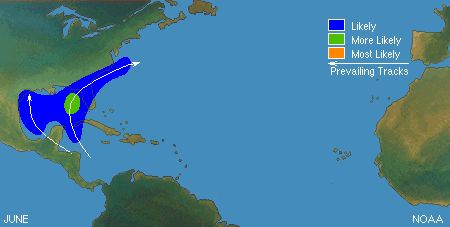}}
							  \subfigure[]{\label{june1}
							\includegraphics[width=0.15\textwidth]{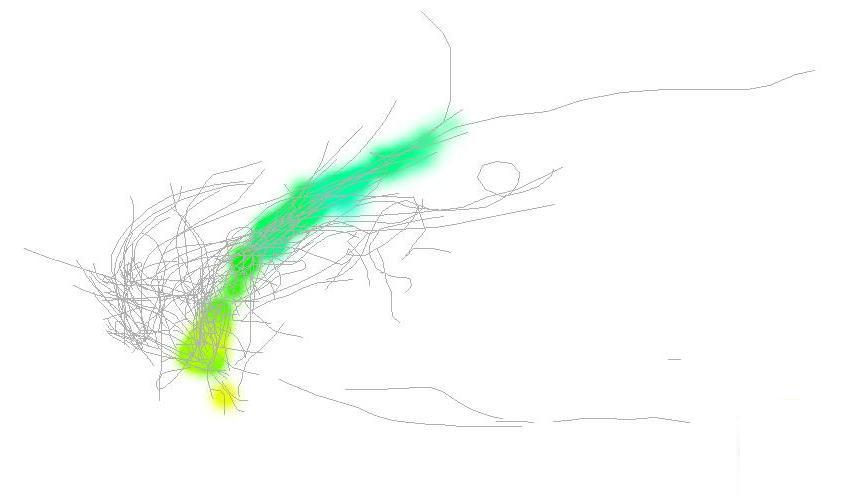}}
							  \subfigure[]{\label{june2}
							\includegraphics[width=0.15\textwidth]{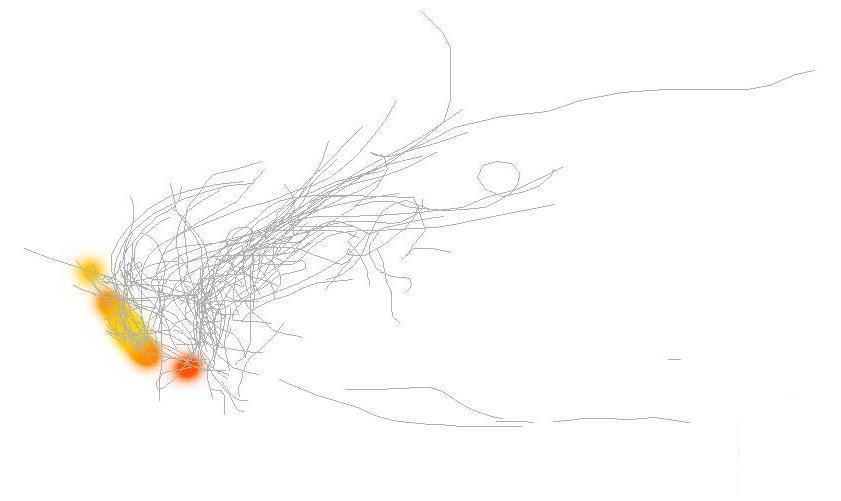}}
							  \subfigure[]{\label{july}
							\includegraphics[width=0.15\textwidth]{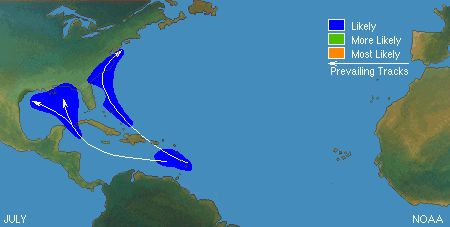}}
							  \subfigure[]{\label{july1}
							\includegraphics[width=0.15\textwidth]{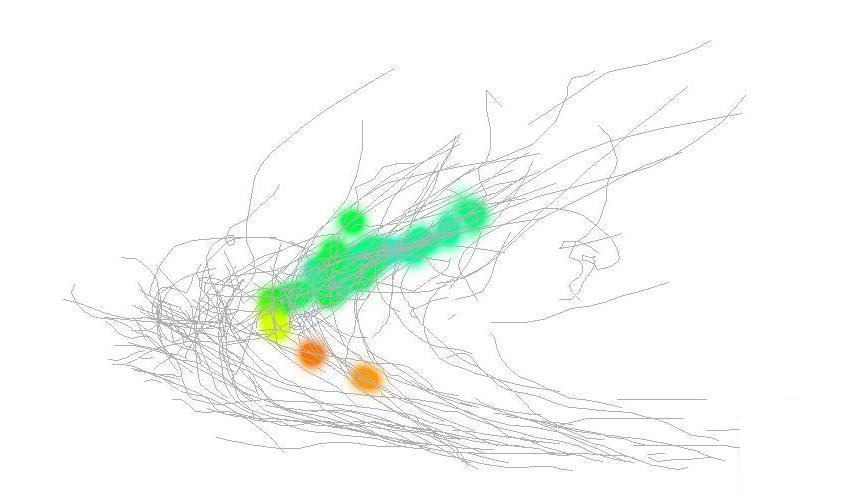}}
							  \subfigure[]{\label{july2}
							\includegraphics[width=0.15\textwidth]{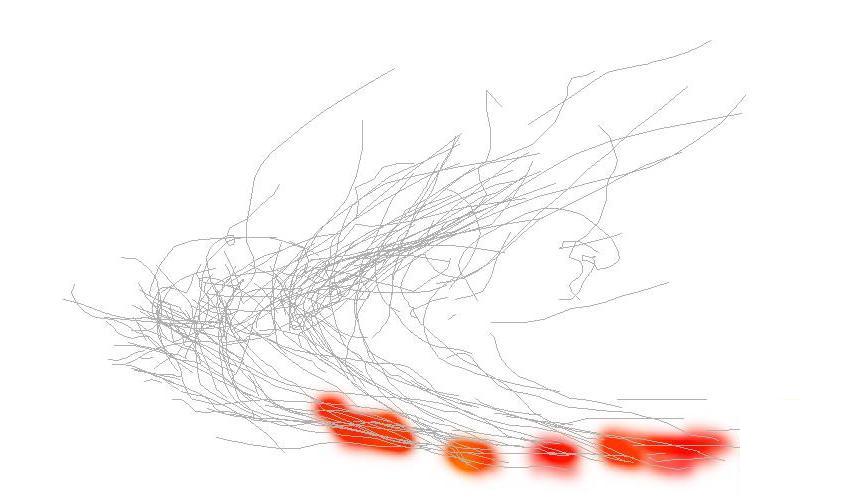}}
							  \subfigure[]{\label{july4}
							\includegraphics[width=0.15\textwidth]{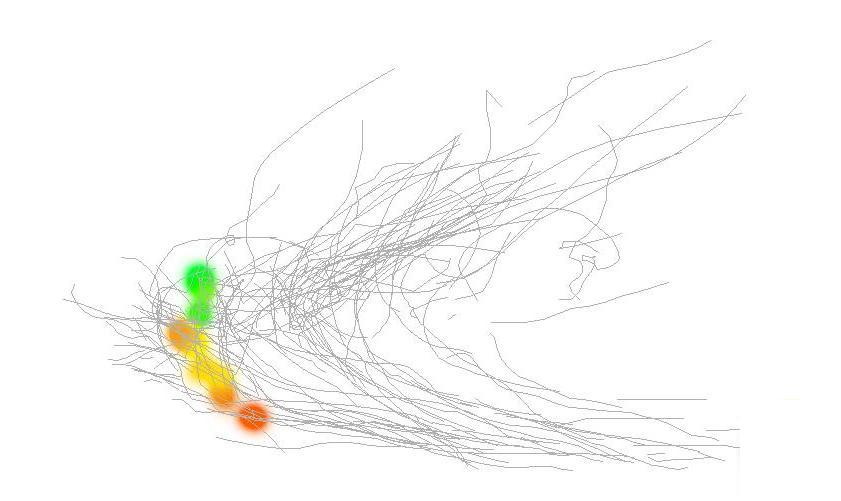}}
						        \subfigure[]{\label{july6}
							\includegraphics[width=0.15\textwidth]{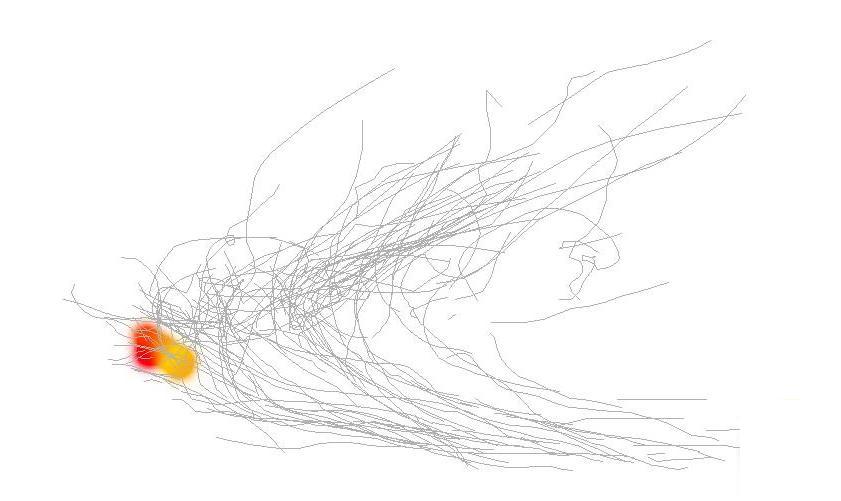}}	
							\subfigure[]{\label{aug}
							\includegraphics[width=0.15\textwidth]{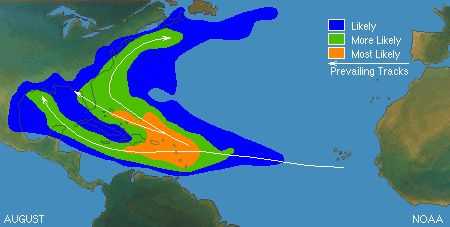}}
							  \subfigure[]{\label{aug1}
							\includegraphics[width=0.15\textwidth]{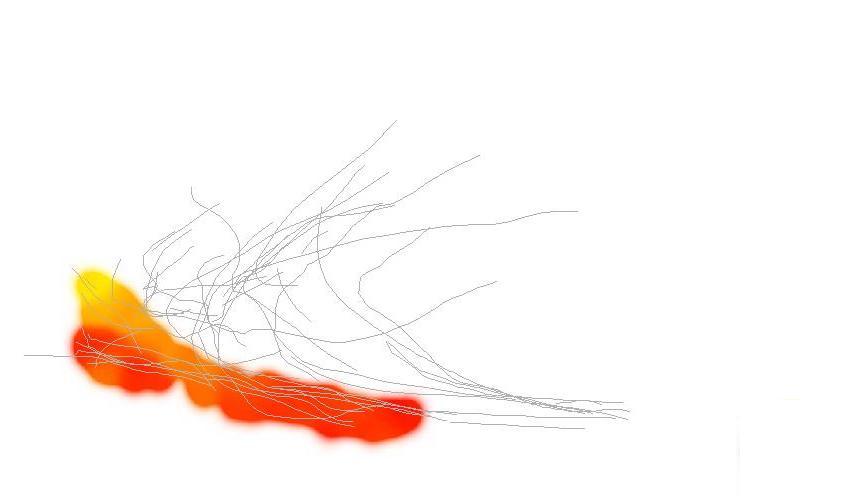}}
							  \subfigure[]{\label{aug2}
							\includegraphics[width=0.15\textwidth]{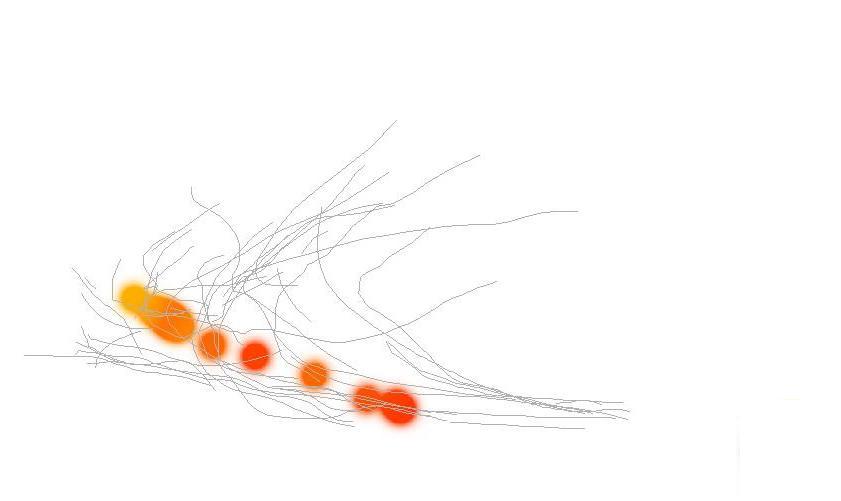}}
							  \subfigure[]{\label{aug3}
							\includegraphics[width=0.15\textwidth]{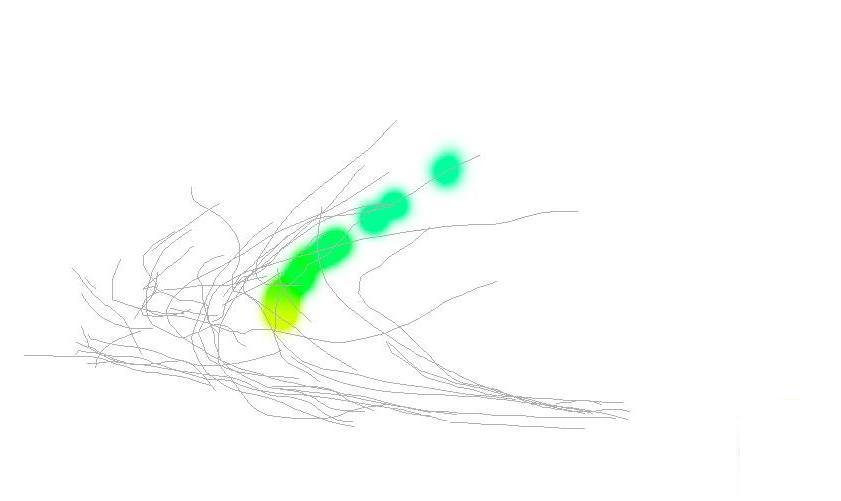}}	
							\subfigure[]{\label{sep}
							\includegraphics[width=0.15\textwidth]{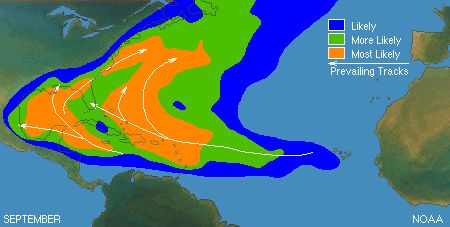}}
							  \subfigure[]{\label{sep1}
							\includegraphics[width=0.15\textwidth]{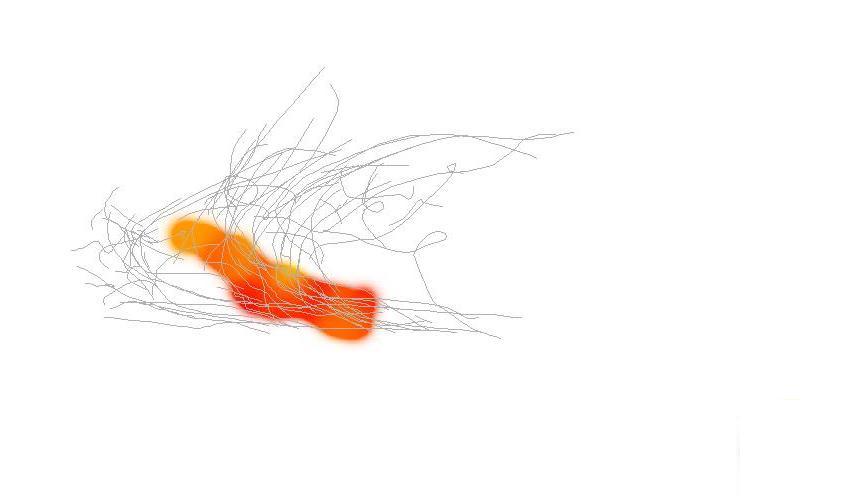}}
							  \subfigure[]{\label{sep2}
							\includegraphics[width=0.15\textwidth]{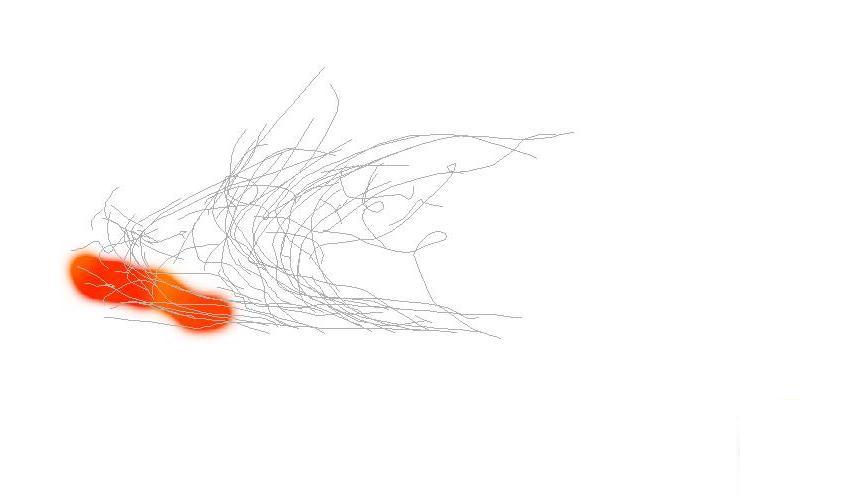}}
							  \subfigure[]{\label{sep4}
							\includegraphics[width=0.15\textwidth]{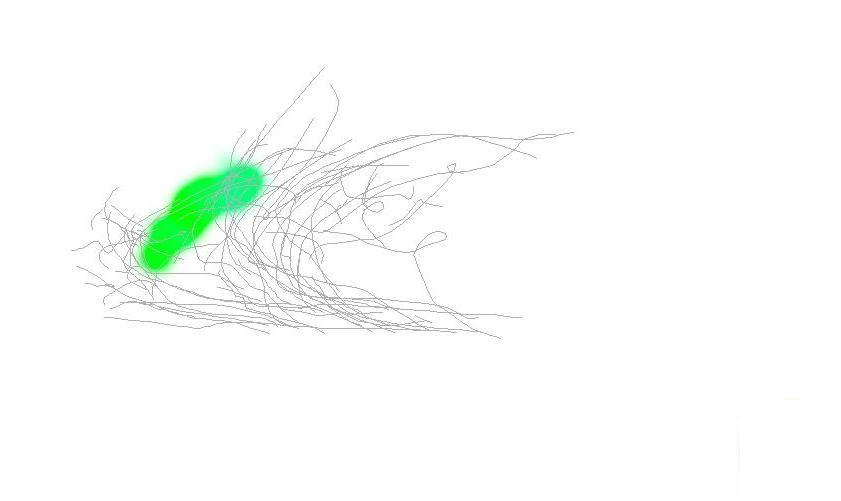}}
							\subfigure[]{\label{sep5}
							\includegraphics[width=0.15\textwidth]{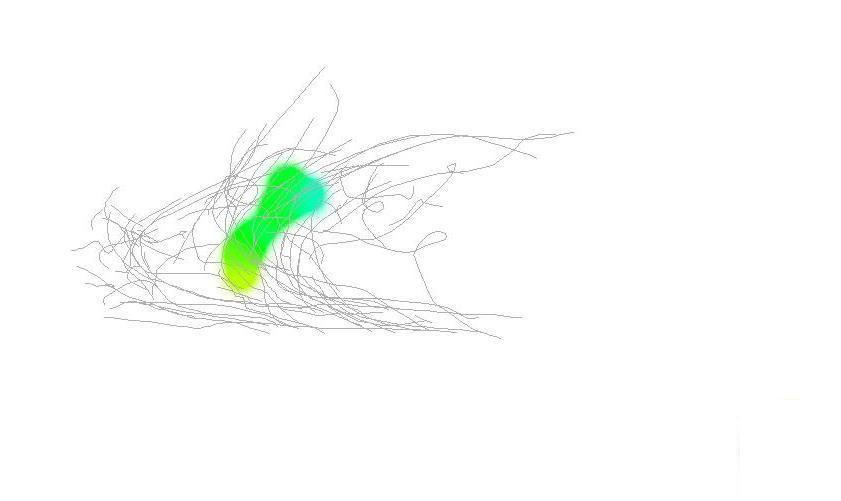}}
							\subfigure[]{\label{sep6}
							\includegraphics[width=0.15\textwidth]{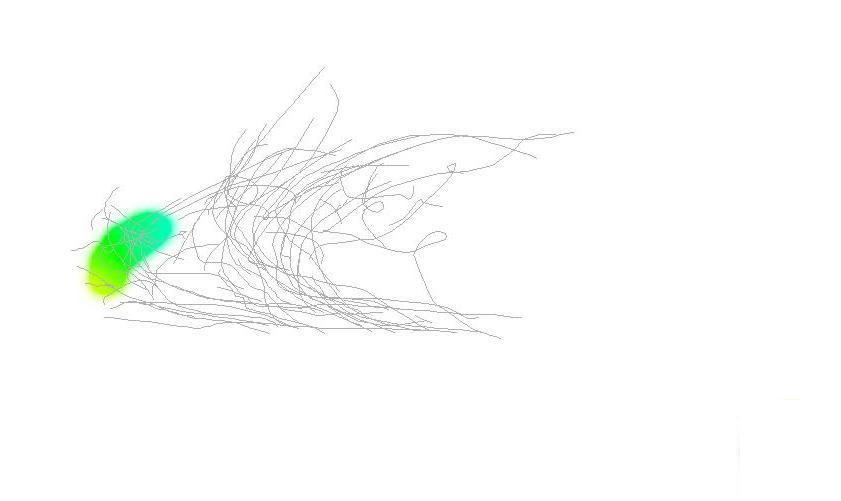}}	
							\subfigure[]{\label{oct}
							\includegraphics[width=0.15\textwidth]{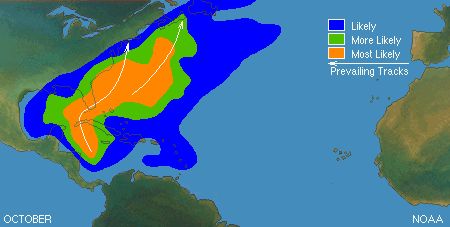}}
							  \subfigure[]{\label{oct1}
							\includegraphics[width=0.15\textwidth]{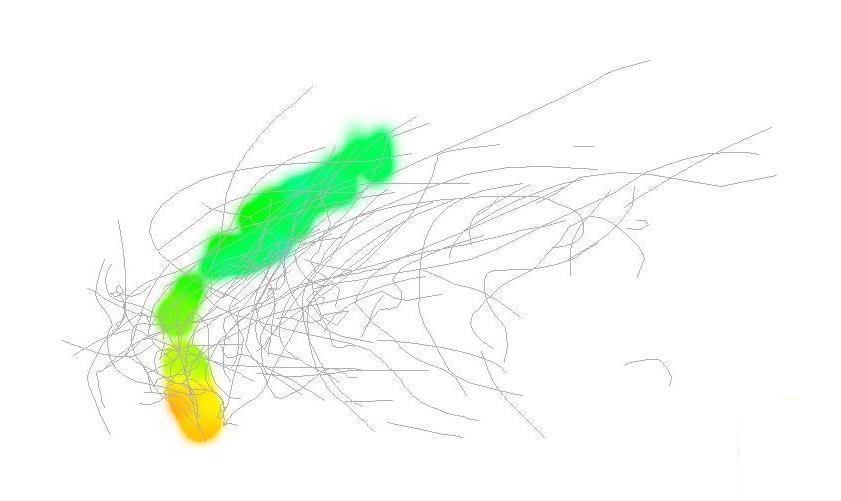}}
							  \subfigure[]{\label{oct3}
							\includegraphics[width=0.15\textwidth]{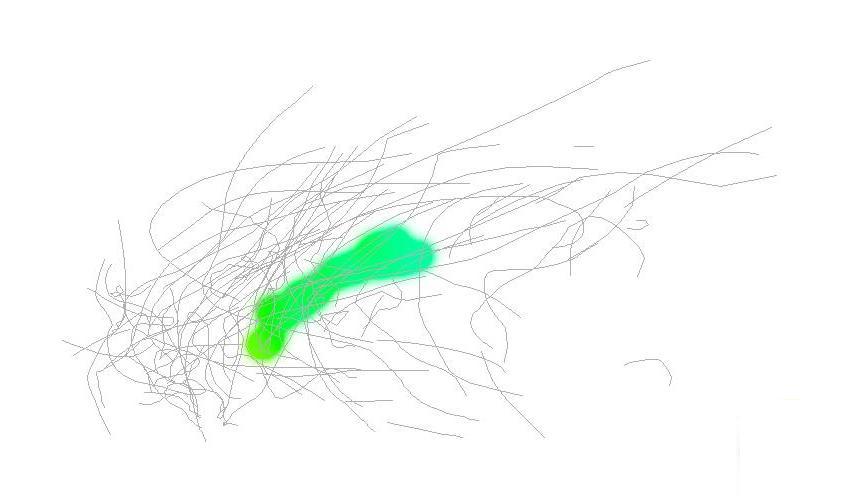}}	  
							\subfigure[]{\label{nov}
							\includegraphics[width=0.15\textwidth]{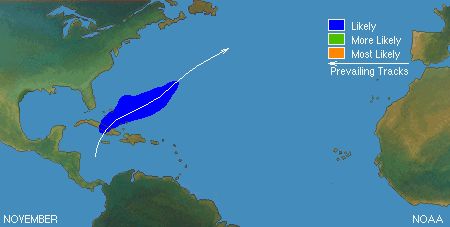}}
							\subfigure[]{\label{nov1}
							\includegraphics[width=0.15\textwidth]{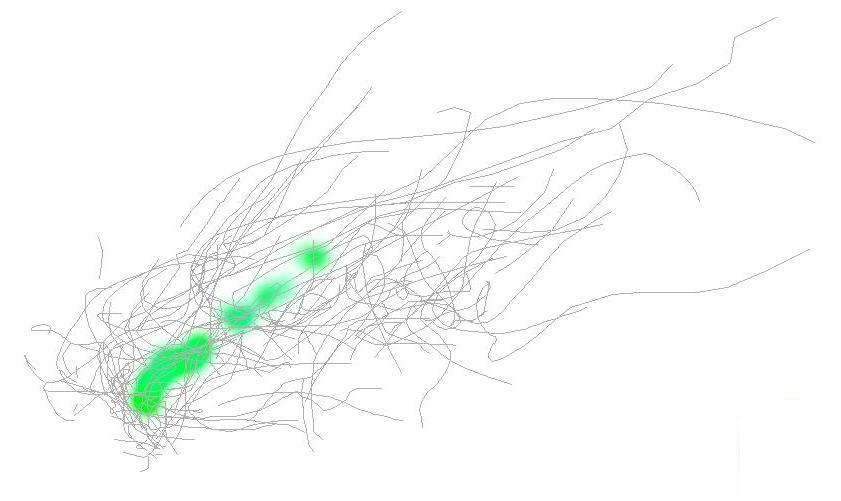}}
				\subfigure[]{\label{colorwheel3}
							\includegraphics[width=0.03\textwidth]{figures/colorwheel.jpg}}
							  \caption{\footnotesize{Extracted motion patterns from Atlantic Hurricane Dataset. Colorwheel is shown in \ref{colorwheel3}.}}
							\label{Hurricane}
							 \end{center}
\vspace{-5mm}
							\end{figure}

							We evaluated the performance of the proposed method on the Atlantic Hurricane Dataset by comparing its output with the annotations provided by the National Hurricane Service (NHS). Prevailing hurricane tracks for each month are indicated by white arrows in Figures \ref{june}, \ref{july}, \ref{aug}, \ref{sep}, \ref{oct} and \ref{nov}. Examining the output for each month, we see that the proposed algorithm recovers motion patterns closely resembling the prevailing hurricane tracks for June in \ref{june1} and \ref{june2}, for October in \ref{oct1} and \ref{oct3}, and for November in \ref{nov1}.  For July, \ref{july1} reflects the rightmost track, while \ref{july2} reflects the shared path of the leftmost tracks, and \ref{july4} and \ref{july6} serve as the distinct portions of the two leftmost paths. For August, \ref{aug1} reflects the bottom left prevailing track, while \ref{aug2} and \ref{aug3} correspond to the splitting tracks on the top right. September hurricane tracks in \ref{sep} are split into two groups of three. The left group of three is output as motion patterns \ref{sep2}, \ref{sep6}, \ref{sep4}, while the right group of three is output as \ref{sep1} for the arrow pointing northwest and \ref{sep5} for the two arrows turning towards northeast.  There is a considerable variation in trajectory density across different months in the dataset. As a general rule, months with fewer hurricanes, such as June, July and November, require more relaxed reachability conditions (larger double-ellipse and wedge radius), while months with high number of hurricanes, such as August, September and October, require stricter ones.
							
						\begin{figure}[!ht]
							  \begin{center}
							  \subfigure[]{\label{hawks1}
							\includegraphics[width=0.15\textwidth]{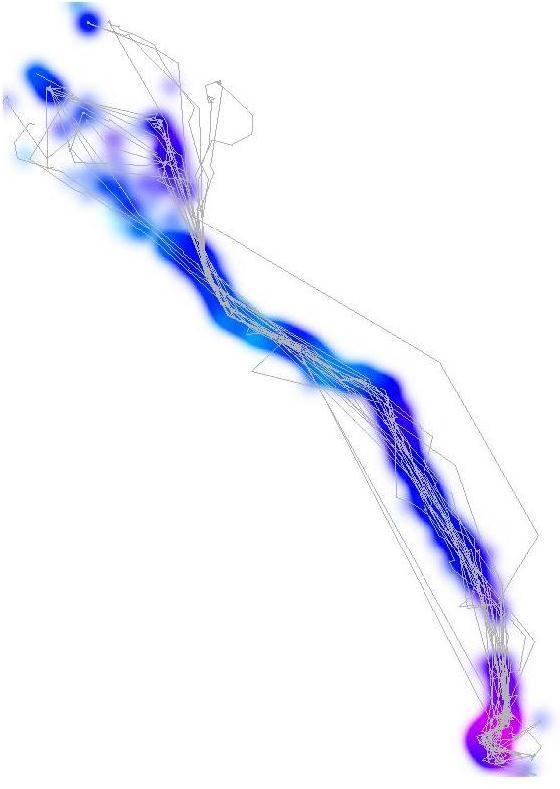}}
							  \subfigure[]{\label{hawks2}
							\includegraphics[width=0.15\textwidth]{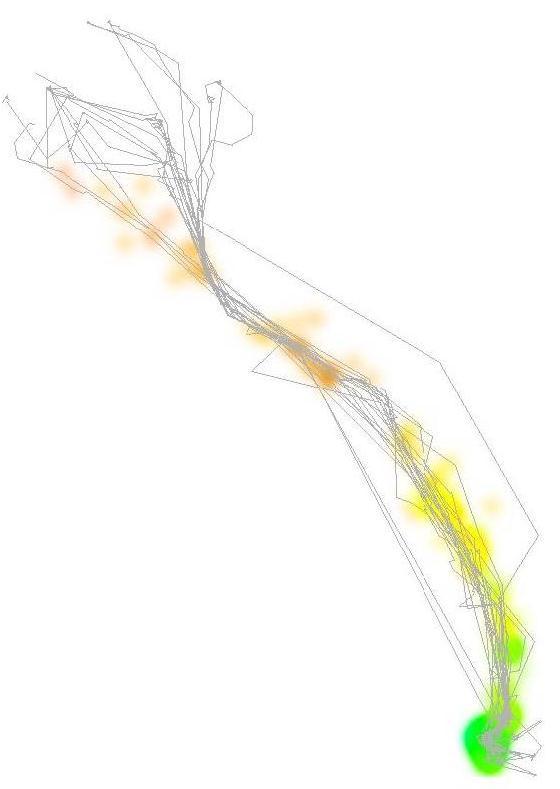}}
				\subfigure[]{\label{colorwheel4}
							\includegraphics[width=0.03\textwidth]{figures/colorwheel.jpg}}
							  \caption{\footnotesize{Extracted motion patterns from Swainson's Hawks Dataset. Colorwheel is shown in \ref{colorwheel4}.}}
							\label{Hawks}
							 \end{center}
\vspace{-4mm}
							\end{figure}
						
						\begin{figure}[!ht]

							  \begin{center}
							  \subfigure[]{\label{truck1}
							\includegraphics[width=0.12\textwidth]{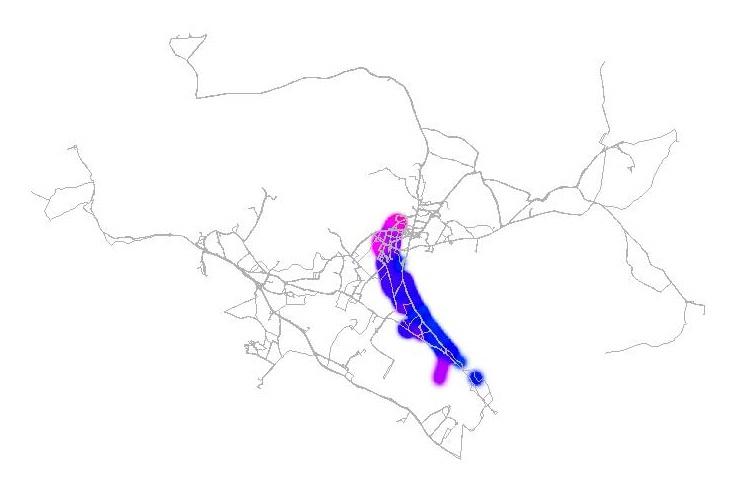}}
							  \subfigure[]{\label{truck2}
							\includegraphics[width=0.12\textwidth]{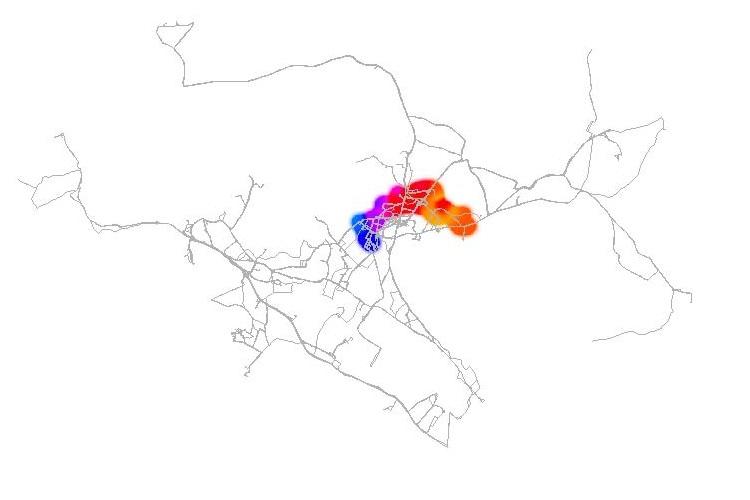}}
							\subfigure[]{\label{truck3}
							\includegraphics[width=0.12\textwidth]{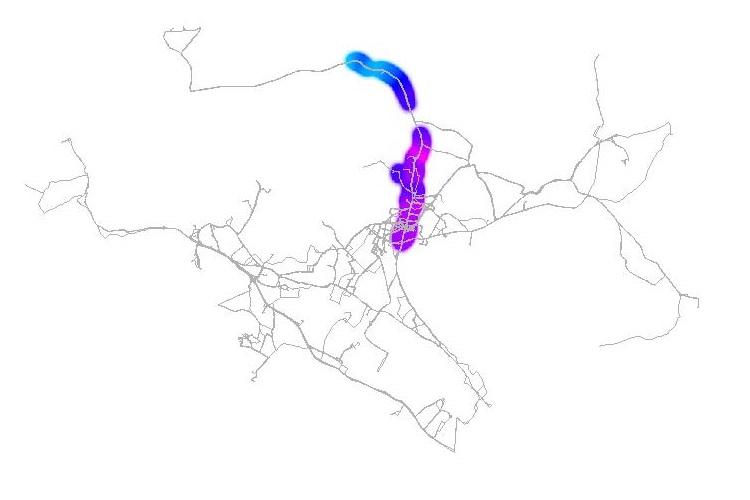}}
							  \subfigure[]{\label{truck4}
							\includegraphics[width=0.12\textwidth]{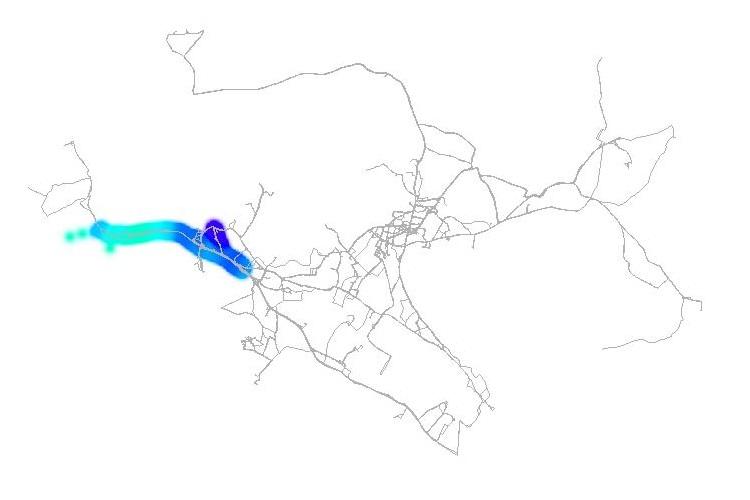}}
							\subfigure[]{\label{truck5}
							\includegraphics[width=0.12\textwidth]{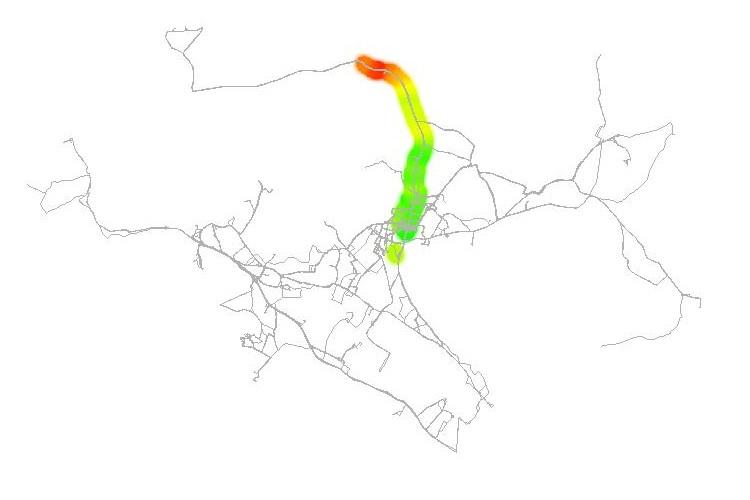}}
							  \subfigure[]{\label{truck6}
							\includegraphics[width=0.12\textwidth]{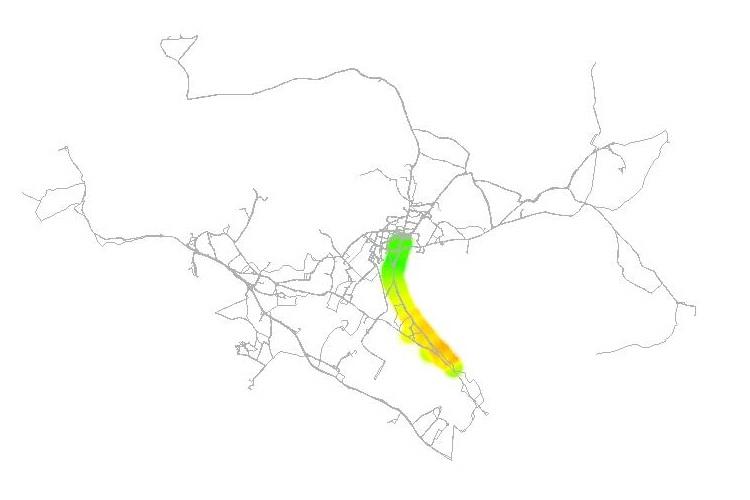}}
							\subfigure[]{\label{truck7}
							\includegraphics[width=0.12\textwidth]{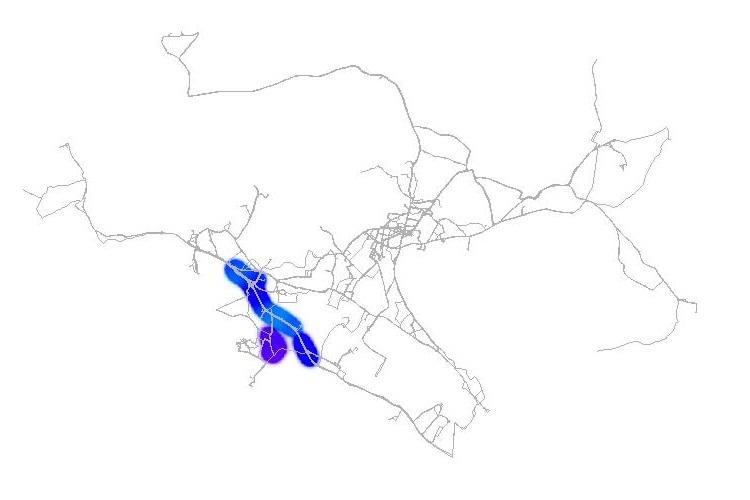}}
							  \subfigure[]{\label{truck11}
							\includegraphics[width=0.12\textwidth]{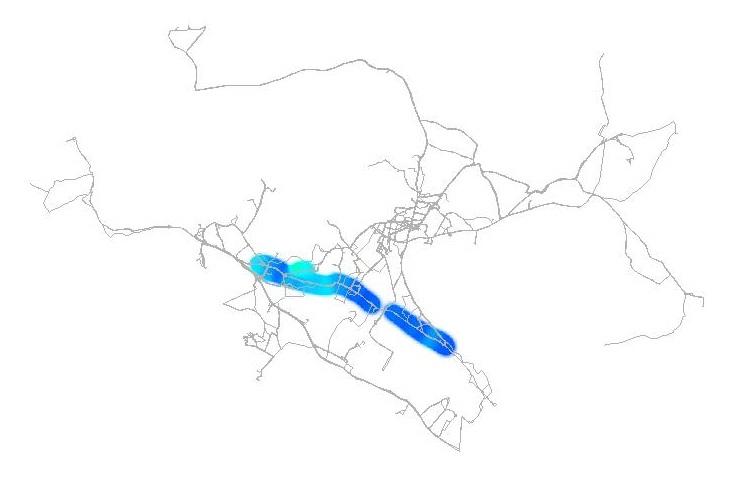}}
							\subfigure[]{\label{truck13}
							\includegraphics[width=0.12\textwidth]{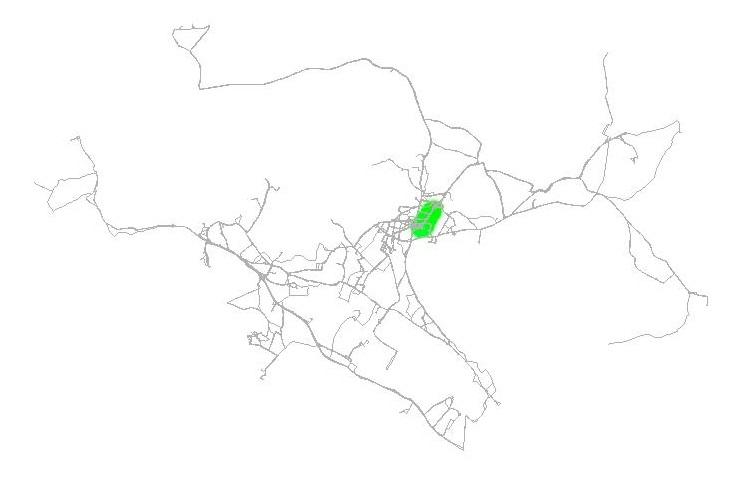}}
							  \subfigure[]{\label{truck14}
							\includegraphics[width=0.12\textwidth]{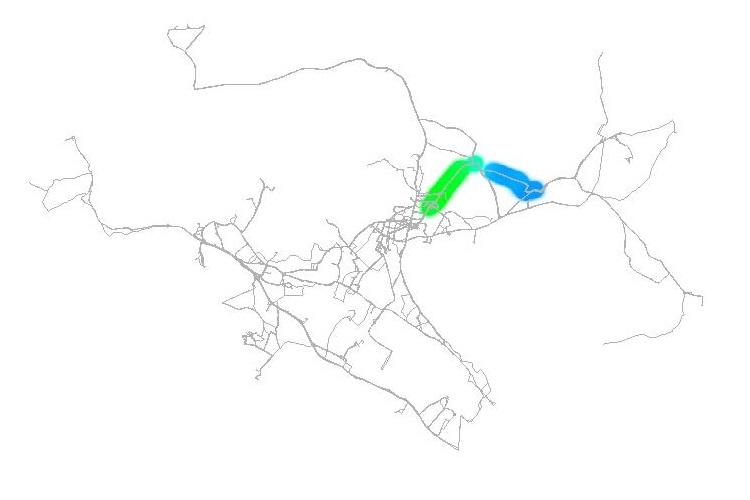}}
				\subfigure[]{\label{colorwheel5}
							\includegraphics[width=0.03\textwidth]{figures/colorwheel.jpg}}
							  \caption{\footnotesize{Extracted motion patterns from The Greek Trucks Dataset. Colorwheel is shown in \ref{colorwheel5}.}}
							\label{Trucks}
							 \end{center}
\vspace{-3mm}
							\end{figure}
			
			                                 \begin{figure}[!ht]
							  \begin{center}
							  \subfigure[]{\label{lankershim7}
							\includegraphics[width=0.08\textwidth]{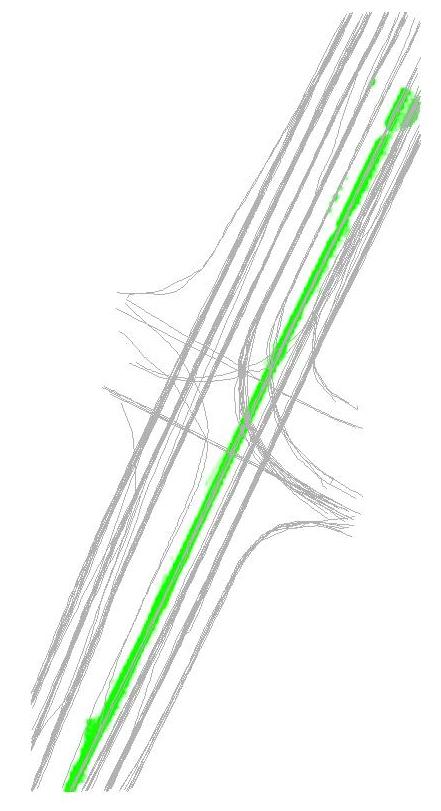}}
							  \subfigure[]{\label{lankershim4}
							\includegraphics[width=0.08\textwidth]{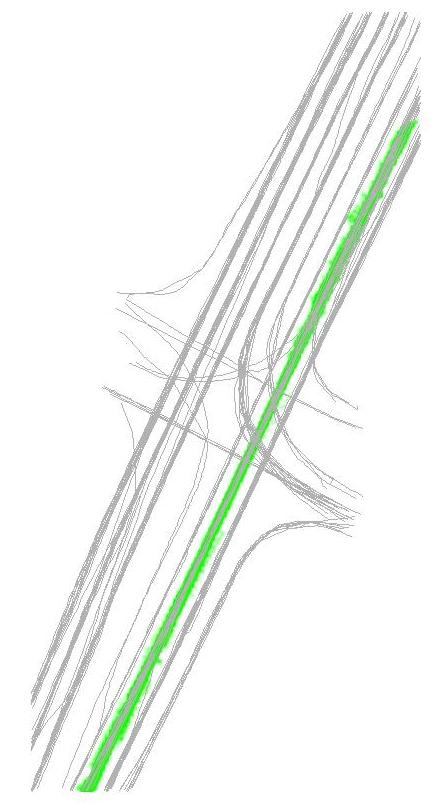}}
							\subfigure[]{\label{lankershim2}
							\includegraphics[width=0.08\textwidth]{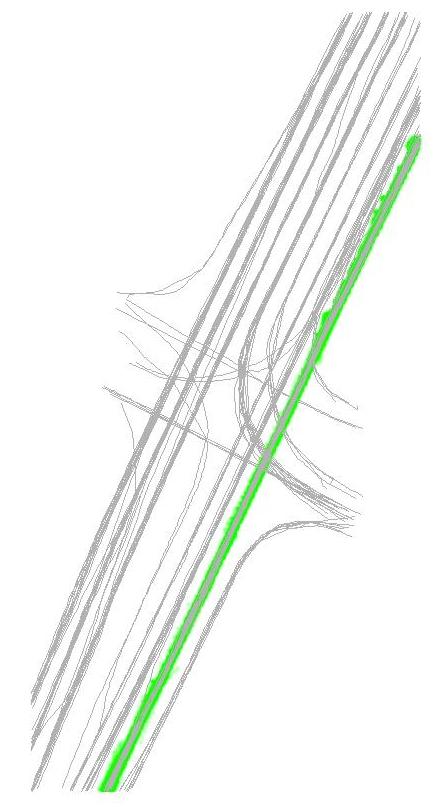}}
							  \subfigure[]{\label{lankershim3}
							\includegraphics[width=0.08\textwidth]{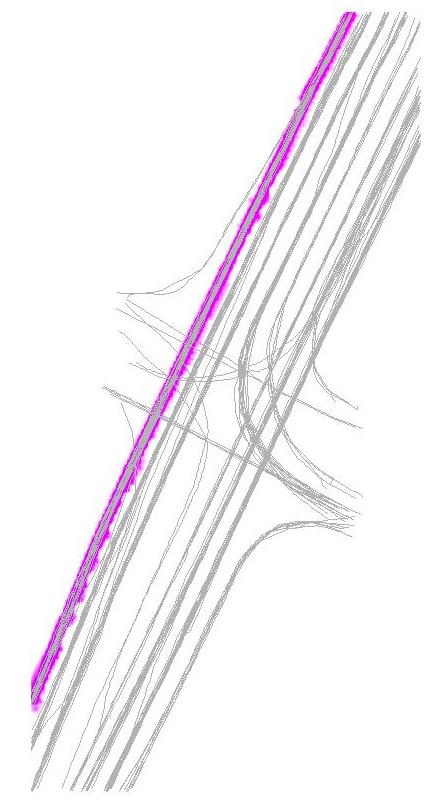}}
							\subfigure[]{\label{lankershim5}
							\includegraphics[width=0.08\textwidth]{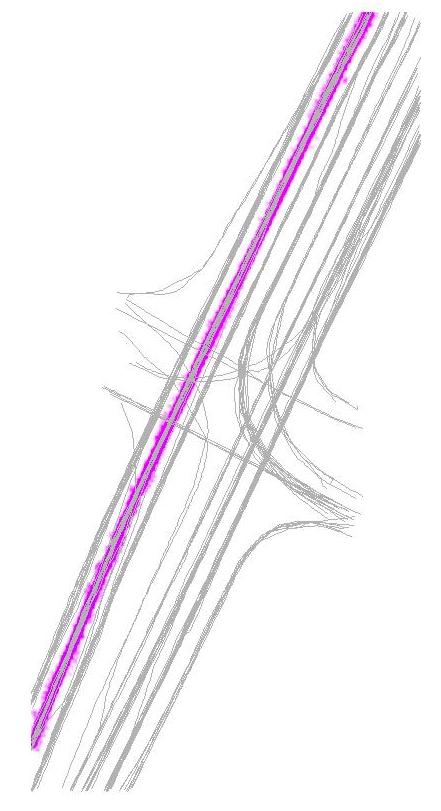}}
							  \subfigure[]{\label{lankershim1}
							\includegraphics[width=0.08\textwidth]{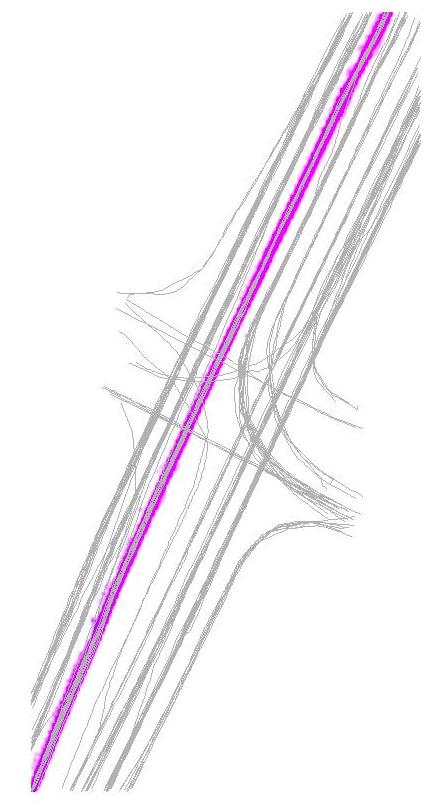}}
			                                \subfigure[]{\label{lankershim6}
							\includegraphics[width=0.08\textwidth]{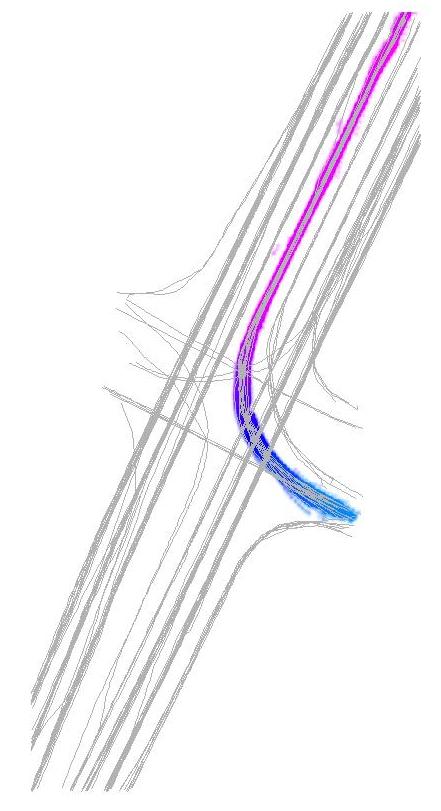}}
							  \subfigure[]{\label{lankershim9}
							\includegraphics[width=0.08\textwidth]{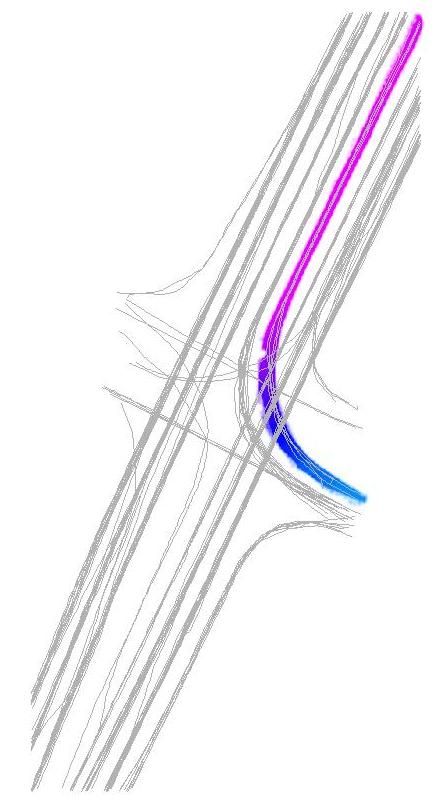}}
							\subfigure[]{\label{lankershim8}
							\includegraphics[width=0.08\textwidth]{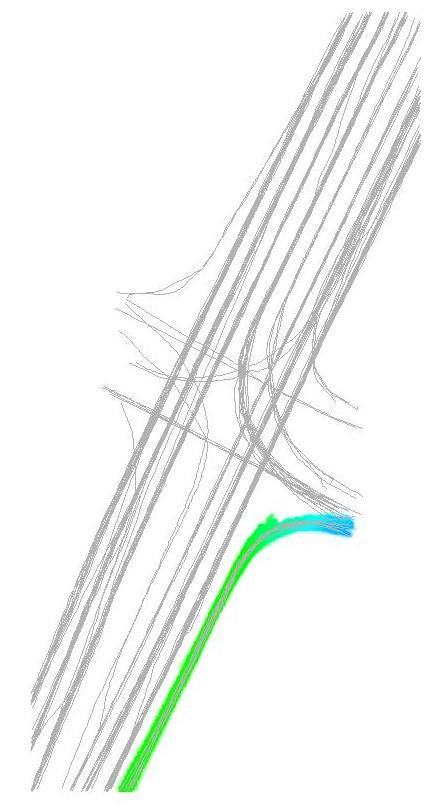}}
							  \subfigure[]{\label{lankershim11}
							\includegraphics[width=0.08\textwidth]{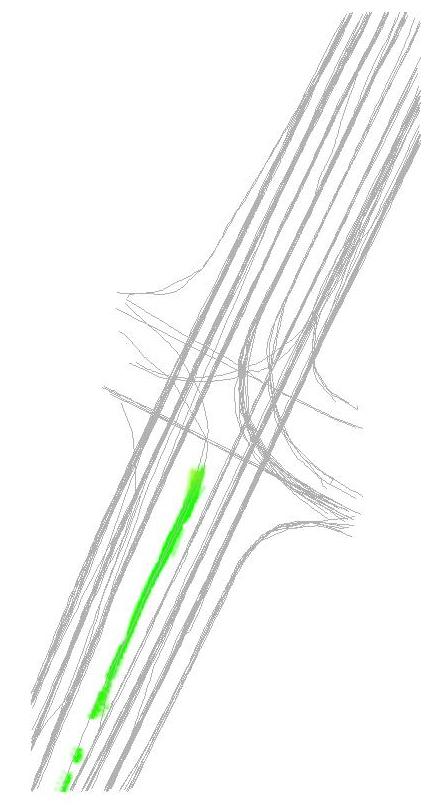}}
							\subfigure[]{\label{lankershim12}
							\includegraphics[width=0.08\textwidth]{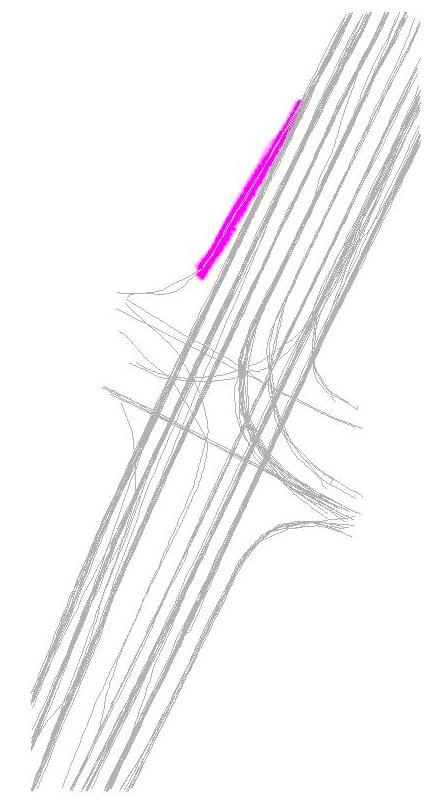}}
							  \subfigure[]{\label{lankershim13}
							\includegraphics[width=0.08\textwidth]{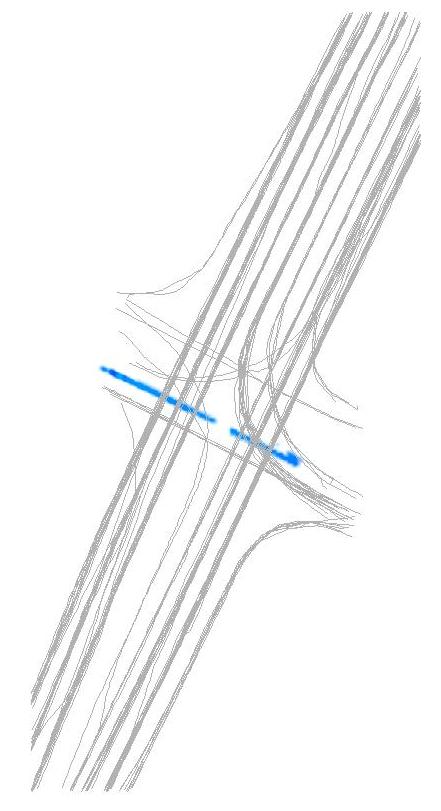}}
							  \subfigure[]{\label{lankershim14}
							\includegraphics[width=0.08\textwidth]{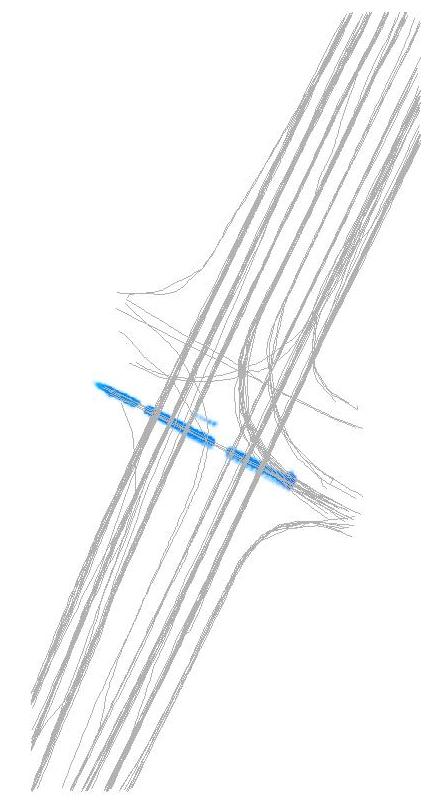}}
							  \subfigure[]{\label{lankershim15}
							\includegraphics[width=0.08\textwidth]{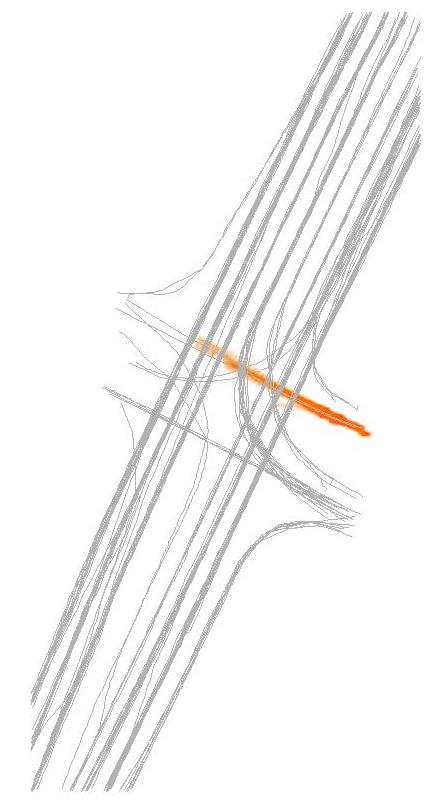}}
							 \subfigure[]{\label{colorwheel6}
							\includegraphics[width=0.03\textwidth]{figures/colorwheel.jpg}}
							  \caption{\footnotesize{Extracted motion patterns from the NGSIM Lankershim Dataset. Colorwheel is shown in \ref{colorwheel6}.}}
							\label{NGSIMLankershim}
							 \end{center}
\vspace{-6mm}
							\end{figure}
						
							\subsection{Evaluation on Swainson's Hawks Dataset}
							We evaluated the proposed method's performance on the Swainson's Hawks Dataset by comparing the algorithm's output with the study summary provided in \cite{kochert2011migration}. In particular, we have been able to retrace the southward and northward migration paths of the birds in Figures \ref{hawks1} and \ref{hawks2} respectively.

							\subsection{Evaluation on Greek Trucks Dataset}
							
							Given the dataset, we find the major highways that are taken by trucks. As previously noted, in the case of two routes merging, the algorithm will find three patterns: two for the routes before they merge, and one for the combined routes after they merge. The patterns in Figures \ref{truck2} and \ref{truck3} merge into the pattern in Figure \ref{truck1}. Similarly, the pattern in Figure \ref{truck4} diverges into the patterns shown in Figures \ref{truck7} and \ref{truck11}. Finally, the patterns shown in Figures \ref{truck6}, \ref{truck13}, and \ref{truck14} merge to create pattern shown in Figure \ref{truck5}.

			\subsection{Evaluation on NGSIM Lankershim Dataset}
							
							The traffic lanes traveling upward are reflected in \ref{lankershim7}, \ref{lankershim4} and \ref{lankershim2}. The three downward lanes are shown in \ref{lankershim3}, \ref{lankershim5} and \ref{lankershim1}. Figures \ref{lankershim6}, \ref{lankershim9} and \ref{lankershim11} reflect the left turns, while \ref{lankershim8} and \ref{lankershim12} reflect right turns. Traffic crossing Lankershim boulevard is reflected in \ref{lankershim13}, \ref{lankershim14} and \ref{lankershim15}.		

						\subsection{Experimental Comparison}
\begin{figure}[!ht]
							  \begin{center}
\vspace{-4mm}
							  \subfigure[]{\label{traclus_intersect}
							\includegraphics[width=0.23\textwidth]{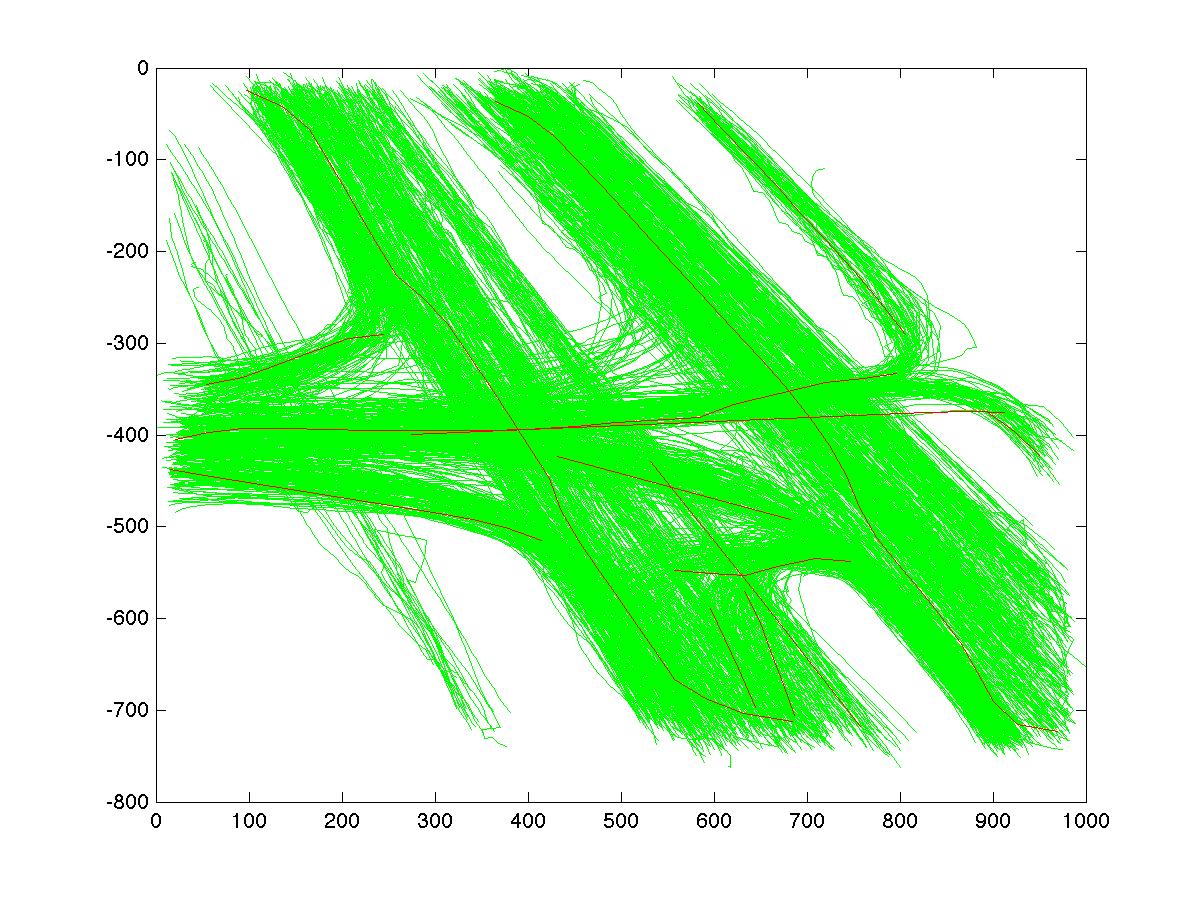}}
							  \subfigure[]{\label{traclusDensityIssue}
							\includegraphics[width=0.23\textwidth]{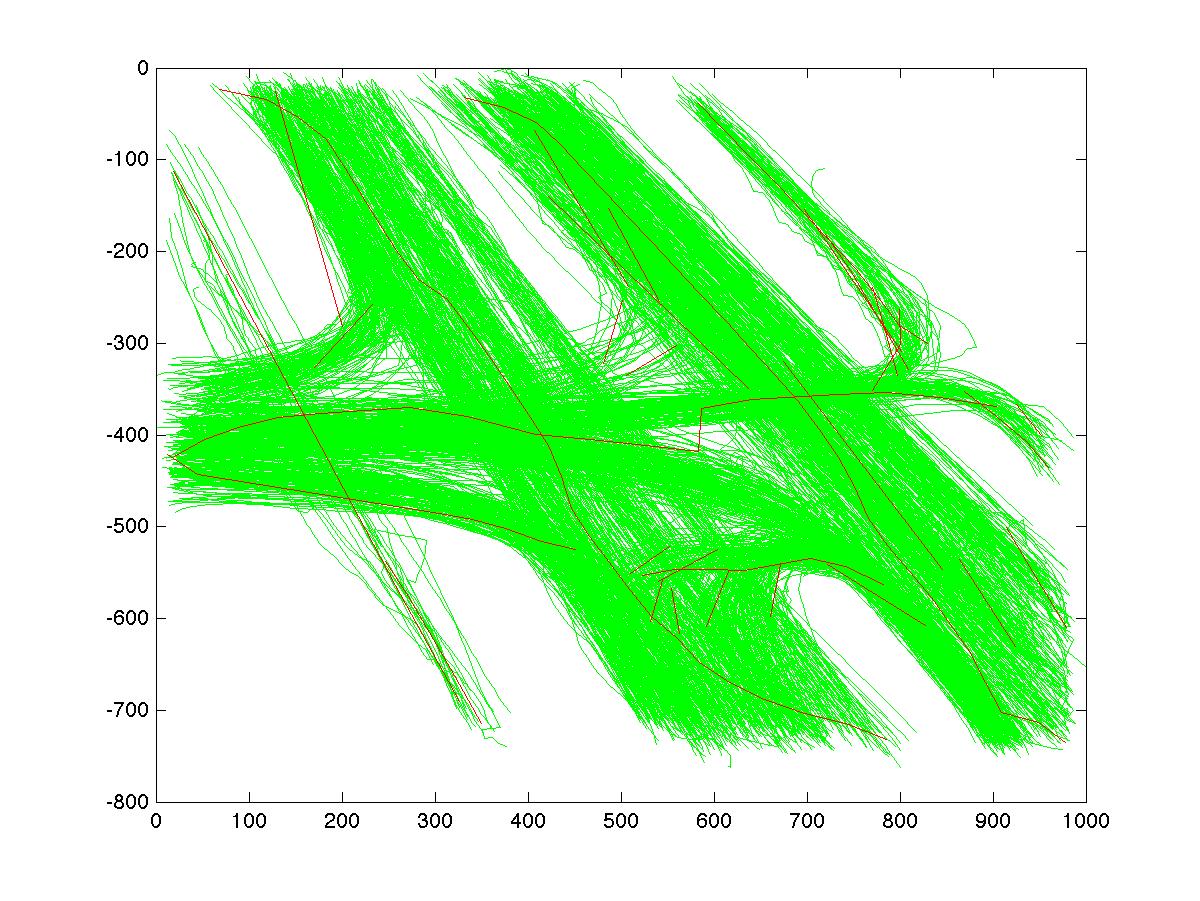}}
							  \caption{\footnotesize{TRACLUS parameters ($\varepsilon$,$\it{MinLns}$) used in \ref{traclus_intersect} and \ref{traclusDensityIssue} are respectively as follows: (20,12), (20,3).}}
							\label{traclusFalse}
\vspace{-4mm}
							 \end{center}
							\end{figure}

						Various quantitative measures for evaluating the quality of clusters exist in the literature. \cite{yang2012trajectory} ,\cite{zhang2006comparison}, \cite{Morris} and \cite{ricci2010learning} computed Correct Clustering Rate (CCR) as a measure to evaluate clustering results against the ground truth or manual annotations. \cite{Atev} used an information theoretic criterion proposed by Meil{\u{a}} \cite{meilua2007}, Variation of Information (VI), to validate obtained clusters. VI determines the amount of information lost and gained between two different clusterings of data. Masciari compared inter-class and intra-class similarity of clusters in \cite{masciari2009},\cite{masciari2012}. All of these and similar measures, however, were used to evaluate clusters of entire trajectories. Since this does not align with the goals of our work, an attempt to use the aforementioned metrics would not yield a meaningful comparison.	
						Rather than clustering entire trajectories, the proposed algorithm mines regional trends in trajectory data like \cite{kang2010mining}, \cite{zhang2009},  \cite{Traclus}, \cite{Divclust}, \cite{ulm} and \cite{giannotti}.  \cite{kang2010mining} evaluates obtained results by visual comparison with GSP \cite{srikant1996mining} and PrefixSpan \cite{han2001prefixspan}. \cite{zhang2009} offers a visual comparison of results with TRACLUS \cite{Traclus}. TRACLUS authors, in turn, state that there is no well-defined measure for density-based clustering methods, and suggest using a metric consisting of sum of squared error and noise penalty to, in authors' own words, "get a hint of the clustering quality". This metric cannot be easily adapted to methods other than variations of TRACLUS. Ferreira $\it{et. al.}$\cite{vector_field}, who use the subset of the Atlantic Hurricane Dataset for their evaluations, report that their results are visually consistent with expected hurricane behavior, as does \cite{guan2013}. \cite{giannotti} provides a visual analysis and suggests a semantic interpretation of their results on the Greek Trucks Dataset. Staying close to the spirit of the works that mine regional trends, we will demonstrate the effectiveness of the proposed method by visually comparing our results with those of TRACLUS \cite{Traclus}, DivCluST \cite{Divclust}, \cite{giannotti} and \cite{ulm}. Due to the space limitation in the paper, we were not able to provide all the visualizations for each of the baseline methods. Instead, we illustrated cases where they could not handle the difficulties provided by the datasets. Compared to the baseline methods, our algorithm can successfully face those challenges.											
						The goal of TRACLUS \cite{Traclus} is to detect similar portions of trajectories, which semantically correspond to the output of our proposed method. More formally, given a set of trajectories, TRACLUS partitions trajectories, clusters them, and outputs a representative trajectory for each cluster. We used TRACLUS implementation available at \cite{TraclusImplementation}. It contains a program that estimates optimum parameter values. According to \cite{Traclus}, these values are approximate and may differ slightly from the true optimal values. Therefore, we ran TRACLUS on each dataset multiple times, tuning each of the parameters. This is not unlike the approach to parameter selection that TRACLUS authors take. Lee $\it{et. al.}$ \cite{Traclus} suggest a heuristic approach to parameter selection, where optimal $\varepsilon$ is selected to minimize the entropy of obtained clusters. Further, the quality of clusters is estimated using the sum of squared error and the noise penalty. Finally, \cite{Traclus} points out that the final selection of optimal parameters for TRACLUS is made using visual inspection and domain knowledge. One of the challenges that TRACLUS faces, as \cite{zhang2009} also mentions, is that it is not capable of simultaneously extracting  both dense and sparse trajectory clusters because its parameters are optimized globally. Modifying its parameters such that it finds sparser clusters leads to redundant clusters in denser regions. Figure \ref{traclus_intersect} shows clusters found by TRACLUS in Vehicle Motion Trajectory, when optimized $\varepsilon$ and $\it{MinLns}$ are used. We tuned these parameters to allow TRACLUS to extract clusters in both sparse and dense regions, but it results in many short, local clusters and redundant ones shown in Figure \ref{traclusDensityIssue}. Our proposed method can  handle variation in density as is illustrated in Figure \ref{intersectOutput}.
						
\begin{figure}[!ht]
							  \begin{center}
\vspace{-4mm}
						 \subfigure[]{\label{Divclust_trucks1}
							\includegraphics[width=0.3\textwidth]{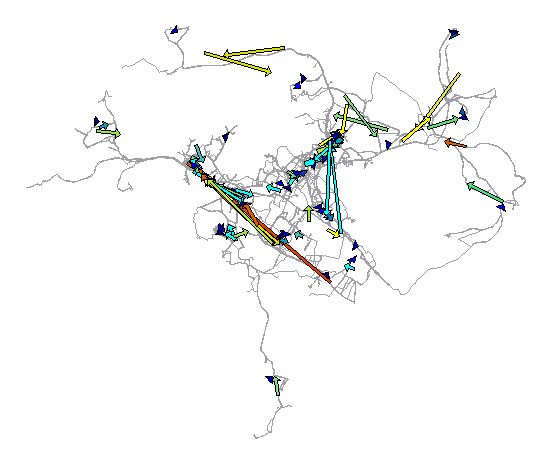}}
							  \subfigure[]{\label{Divclust_trucks2}
							\includegraphics[width=0.3\textwidth]{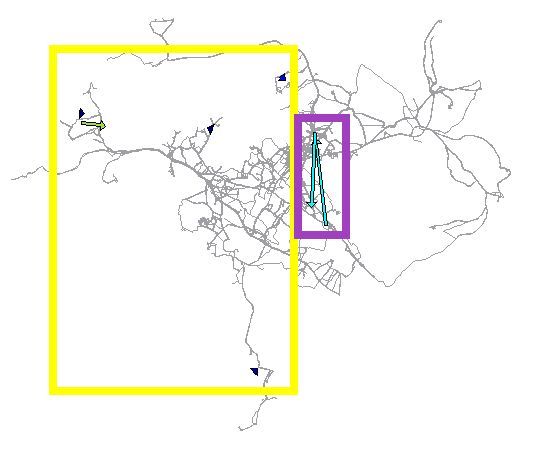}}
							  \caption{\footnotesize{DivCluST on the Greek Trucks Dataset with parameters $\it th_{len}$=0.17, $\it th_{spd}$=0.50, and $\it k$=95 is shown in \ref{Divclust_trucks1}. Yellow box in \ref{Divclust_trucks2} indicates a region with low trajectory density while the purple box contains an example of curvy motion patterns that are modeled as straight lines by DivCluST.}}
							\label{Divclust_trucks}
\vspace{-4mm}
							 \end{center}
							\end{figure}

						DivCluST \cite{Divclust} is an algorithm that seeks to find regional typical moving styles in the form of mean lines. Performance of DivCluST on the Greek Trucks Dataset is shown in Figure \ref{Divclust_trucks1} where  parameters are optimized according to the method described in the paper. Each arrow represents a mean line where the thickness refers to the frequency of that style and the color corresponds to the speed of that style. Warmer, reddish colors are for faster styles while cooler, bluish colors are for slower styles. This algorithm has trouble when there is large variation in the trajectory density. The Greek Trucks Dataset has this property which causes problems. In the high density regions, the mean lines are very cluttered and overlapping. Due to the Kmeans-type model, there are very similar mean lines. This is because if the random initial clusters are close together in a high density region, a Kmeans-type algorithm will often keep them close together. In the low density regions, there are mean lines that are not representative, such as the mean lines within the yellow box in Figure \ref{Divclust_trucks2}. This is because in low density regions, quite different representative segments are clustered together, producing mean lines that are dissimilar from all the representative segments. Additionally, because the model is restricted to straight mean lines rather than curves, motion that would be better described as a curve is instead required to be described as one long mean line or a sequence of short mean lines. An example of this are the long cyan arrows within the purple box in Figure \ref{Divclust_trucks2}. Our algorithm deals much better with variation in density and has the ability to find curved motion patterns.
						
\begin{figure}[!ht]	
							  \begin{center}

							  \subfigure[]{\label{Ginter1}
							\includegraphics[width=0.15\textwidth]{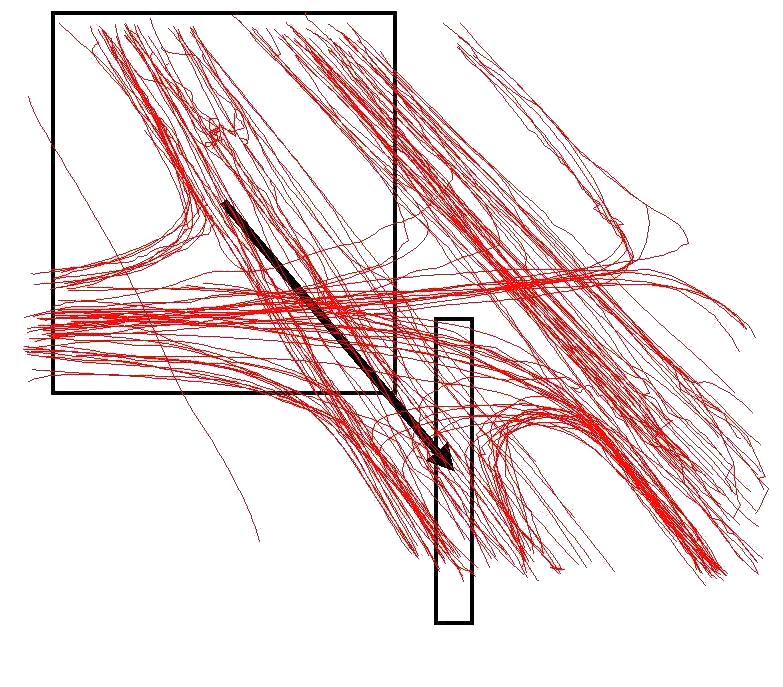}}
							  \subfigure[]{\label{Ginter2}
							\includegraphics[width=0.15\textwidth]{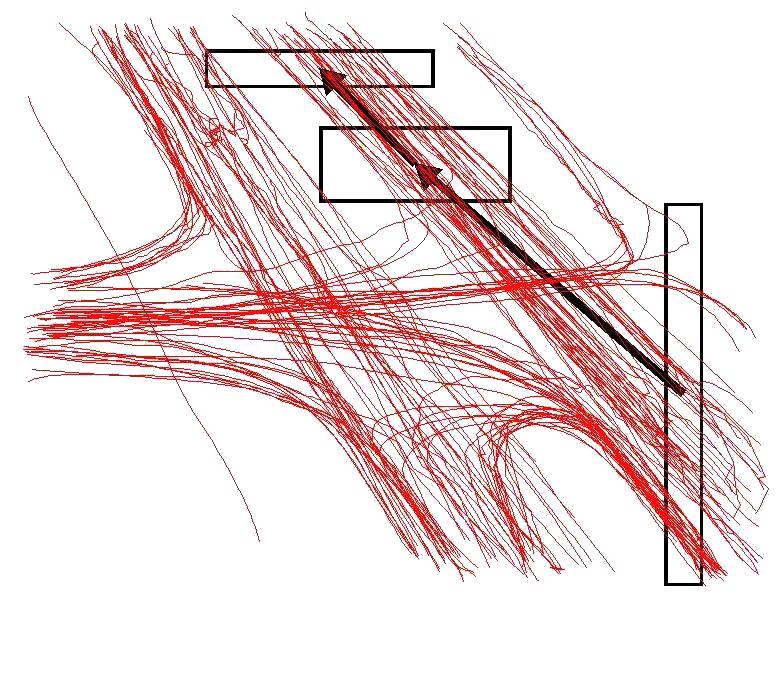}}
						 \subfigure[]{\label{Ginter3}
							\includegraphics[width=0.15\textwidth]{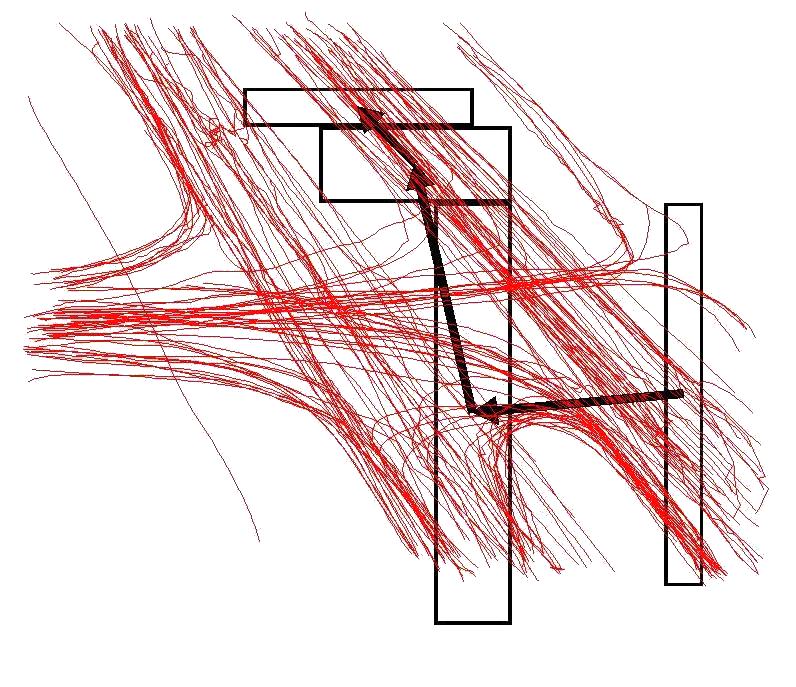}}
						 \subfigure[]{\label{Ginter4}
							\includegraphics[width=0.15\textwidth]{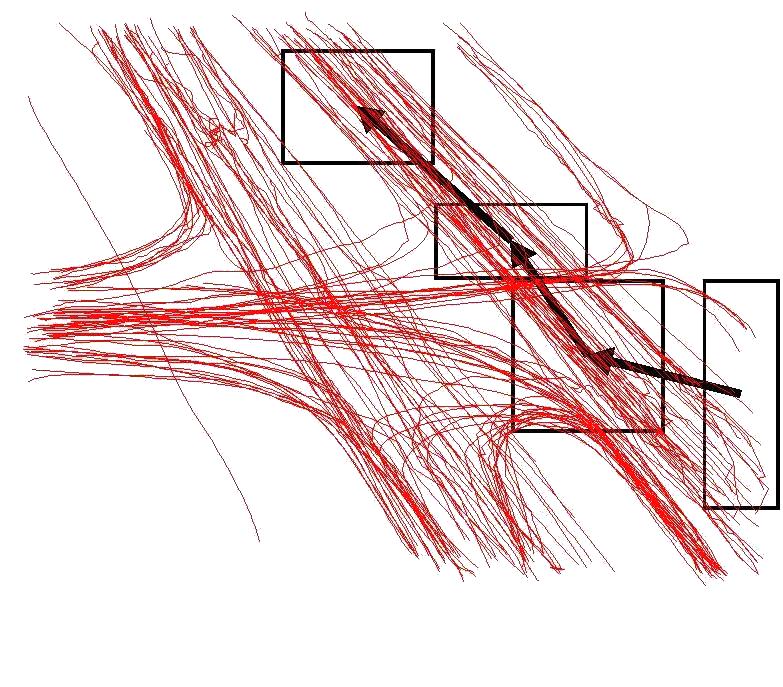}}
						 \subfigure[]{\label{Ginter5}
							\includegraphics[width=0.15\textwidth]{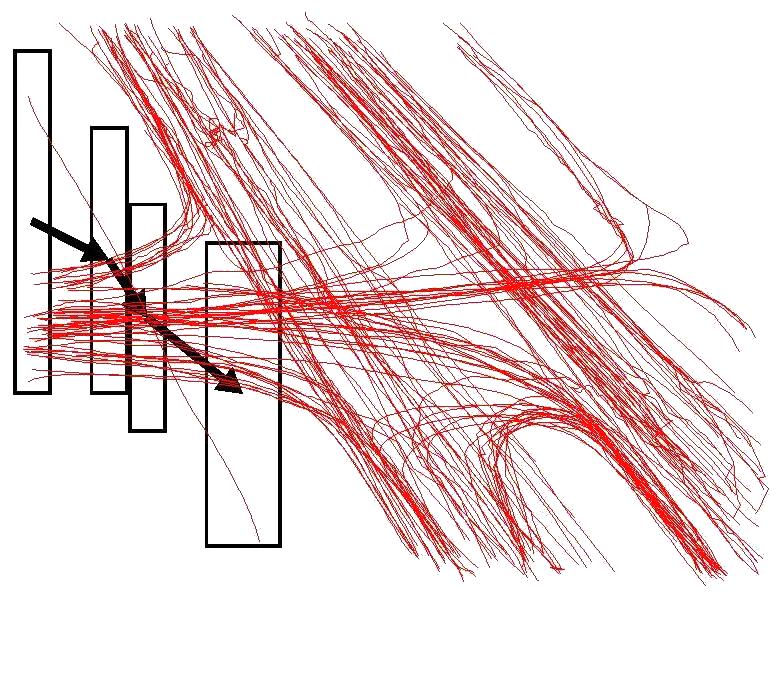}}
						 \subfigure[]{\label{Ginter6}
							\includegraphics[width=0.15\textwidth]{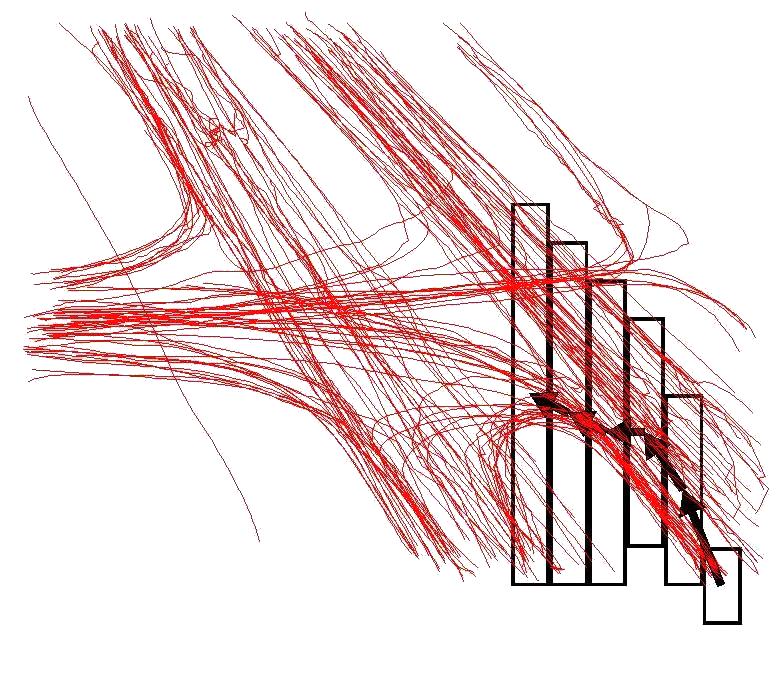}}
						
							 \caption{\footnotesize{ Parameters that \cite{giannotti} has used for Vehicle Motion Trajectory Dataset are as: $S_{min}$=0.17, $\delta$=0.17, $\tau$=100 and $\varepsilon$=25.}}
						\label{gionnati_intersect}
							 \end{center}
\vspace{-6mm}
							\end{figure}

						Giannotti $\it{et. al.}$\cite{giannotti} proposed an algorithm that seeks to find aggregate motion behaviors from trajectories. These behaviors are defined as sequences of rectangular regions. In Figure \ref{gionnati_intersect}, some of the aggregated motion behaviors are illustrated where each is represented as a sequence of rectangular black regions. The order of the sequence is shown by the black arrows. First, it is important to note that this algorithm gives very redundant results. Many of the motion behaviors are very similar and some are even subsets of each other. This requires digging through many patterns to find the distinguishable ones. Although there are 80 generated patterns in total for the Vehicle Motion Trajectory Dataset, only 6 representative ones are shown in Figure \ref{gionnati_intersect}. In other words, of the patterns not shown, each of them has the same shape or is a subset of the shown patterns. Requiring regions to be rectangular restricts the shape of extracted patterns. For instance, in Figure \ref{Ginter1}, the upper right part of the larger box covers a lane of traffic that should not be included in the pattern.  Since the rectangular regions are built only based on density of trajectories without considering motion properties, they are not always suitable to represent motion behaviors. For instance, in Figures \ref{Ginter2}, \ref{Ginter3}, and \ref{Ginter4}, the rightmost box includes a different lane of traffic in the upper portion of the box. Finally, multiple traffic behaviors are sometimes combined into a single pattern. For instance, Figure \ref{Ginter1} shows the traffic turning right and going straight combined, Figure \ref{Ginter3} shows the traffic going straight and making U-turns combined, Figure \ref{Ginter5} shows the access road and incoming traffic from the left combined and Figure \ref{Ginter6} shows the U-turn and straight traffic combined. Unlike this method, the approach which we have proposed in this paper does not generate redundant or overlapping motion patterns. Our motion components are more flexible than rectangular regions used in \cite{giannotti} and our method does not mix multiple patterns of traffic in a motion pattern as can be seen in Figure \ref{intersectOutput}.
						
\begin{figure}[!ht]	
							  \begin{center}
\vspace{-3mm}
							  \subfigure[]{\label{Ulm1}
							\includegraphics[width=0.2\textwidth]{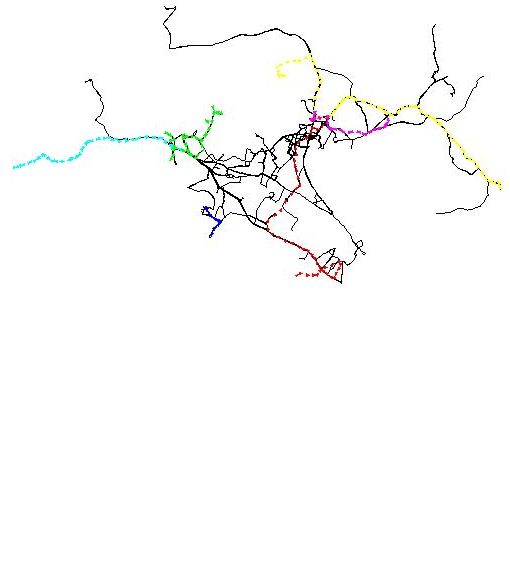}}
							  \subfigure[]{\label{Ulm2}
							\includegraphics[width=0.2\textwidth]{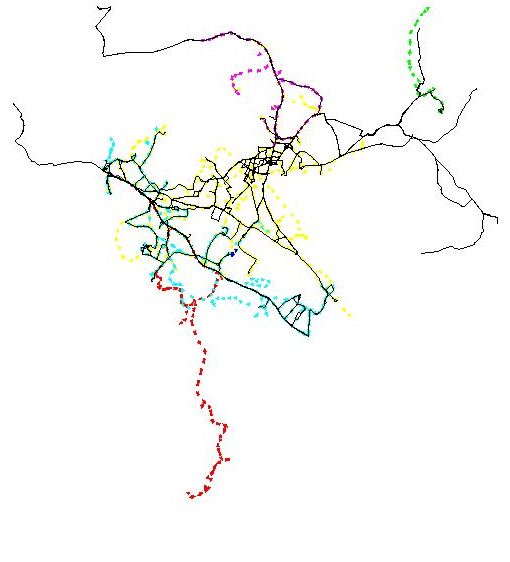}}
						 \subfigure[]{\label{Ulm3}
							\includegraphics[width=0.2\textwidth]{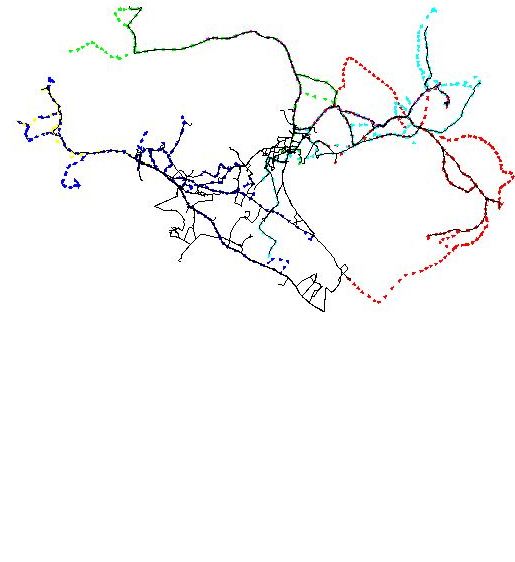}}
						 \subfigure[]{\label{Ulm4}
							\includegraphics[width=0.2\textwidth]{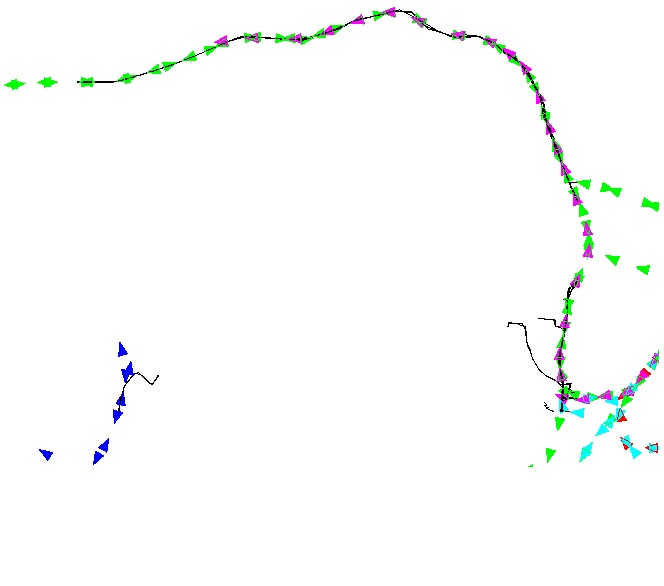}}
							 \caption{\footnotesize{\cite{ulm} is evaluated on the Greek Trucks Dataset with the cluster threshold of 110, the prototype threshold of 100, the position weight of 1 and the direction weight of 100. All other parameters were set to the default values. Figures \ref{Ulm1}, \ref{Ulm2}, \ref{Ulm3} and \ref{Ulm4} are the clusters that \cite{ulm} has found.}}
						\label{Ulm_truck}
\vspace{-7mm}
							 \end{center}
							\end{figure}

						Ulm $\it{et. al.}$\cite{ulm} proposed an algorithm that seeks to find clusters in the form of vector fields defined on a connected spatial set. While this algorithm performs well on datasets that are well structured such as Vehicle Motion Trajectory dataset, it is not well suited for those that lack this property. This is partially due to the fact that it behaves much like the algorithms that cluster trajectories as a whole. It can be seen in Figure \ref{Ulm4} that the vector fields can include directions going two opposite ways on a road. This can be avoided by changing the weights, but this results in vector fields that point perpendicular to the roads. This issue is not present in our algorithm since it can find two spatially overlapping motion patterns with different directions. Secondly, the clusters are not homogeneous in movement. Some of the clusters are very sprawling while others have many small, noisy portions. This is not the effect of using too few clusters since there is a large amount of overlap among them. Instead, this is the effect of using whole trajectories to create vector fields.
						
										\section{Implementation Details}
							
							In this section, the parameters will be discussed. This algorithm includes 16 parameters in total that influence the output of the method. The intuitive interpretation of each of the parameters will be discussed and an approach for fine-tuning them in independent groups will be described.
			
			\begin{figure}[!ht]	
							  \begin{center}

							  \subfigure[]{\label{flowWeight_1000_3}
							\includegraphics[width=0.08\textwidth]{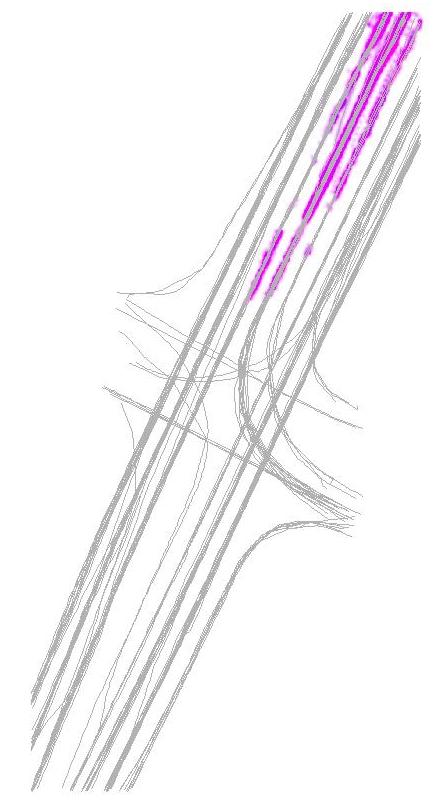}}
							  \subfigure[]{\label{ellipse_width_251}
							\includegraphics[width=0.08\textwidth]{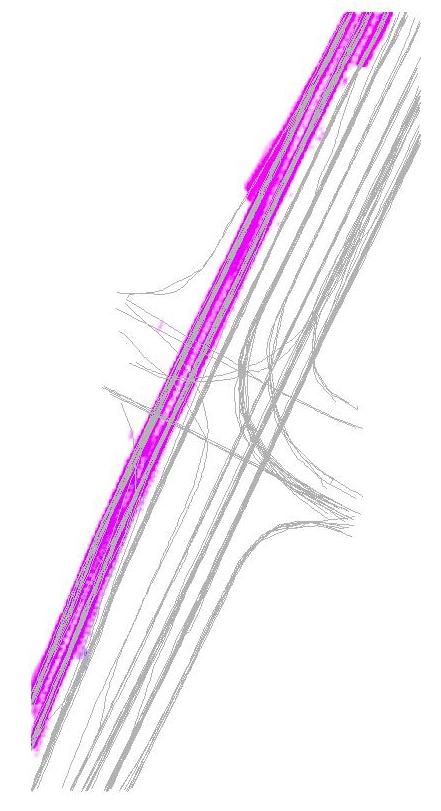}}
						 \subfigure[]{\label{th_flow_120_4}
							\includegraphics[width=0.08\textwidth]{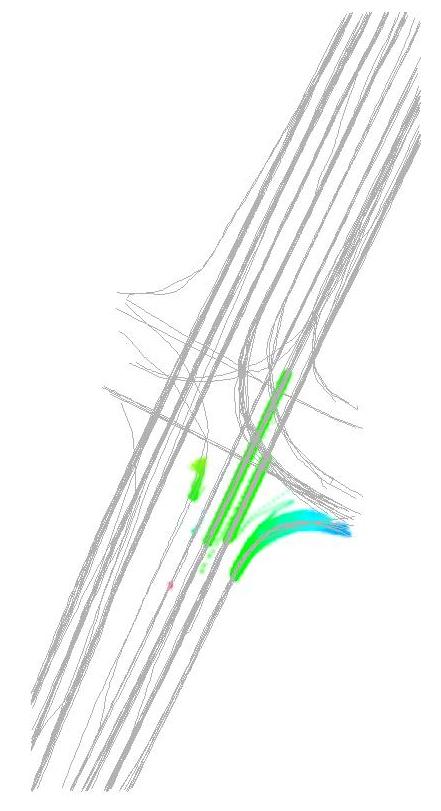}}
			 \subfigure[]{\label{th_flow_2_9}
							\includegraphics[width=0.08\textwidth]{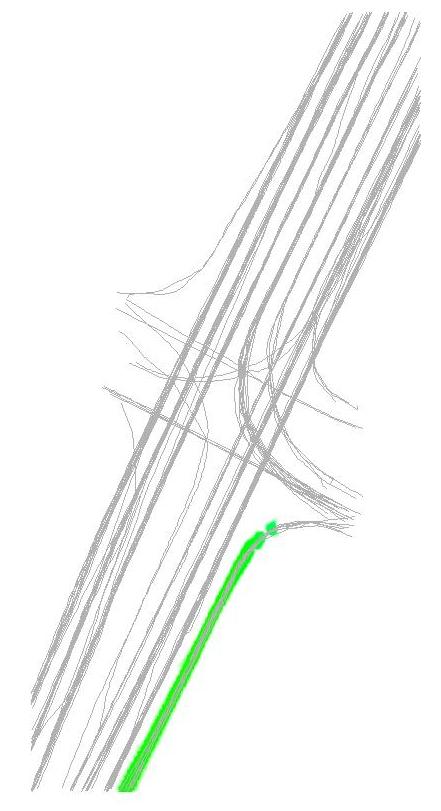}}
		 \subfigure[]{\label{th_flow_2_28}
							\includegraphics[width=0.08\textwidth]{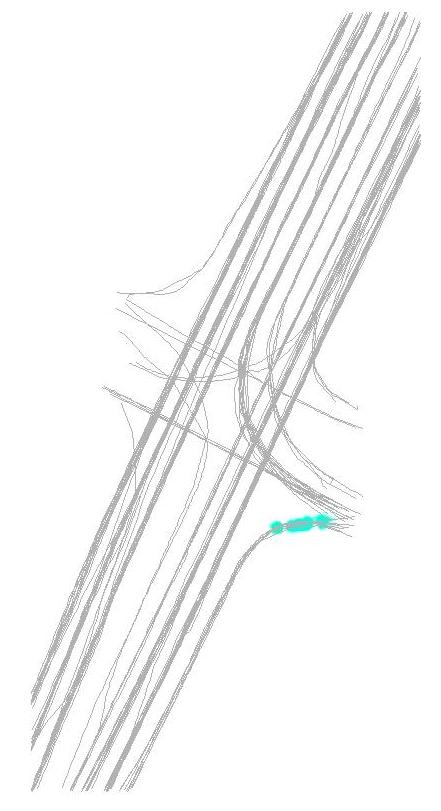}}
						 \subfigure[]{\label{th_flow_2_36}
							\includegraphics[width=0.08\textwidth]{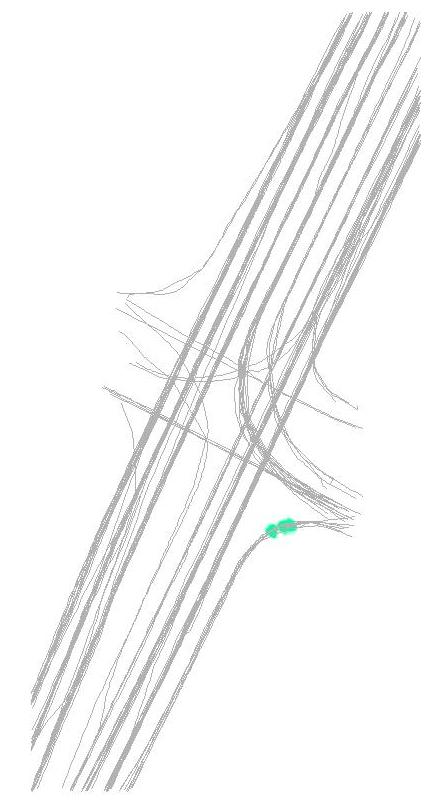}}
			 \subfigure[]{\label{wedge_radius_20_angle_180_2}
							\includegraphics[width=0.08\textwidth]{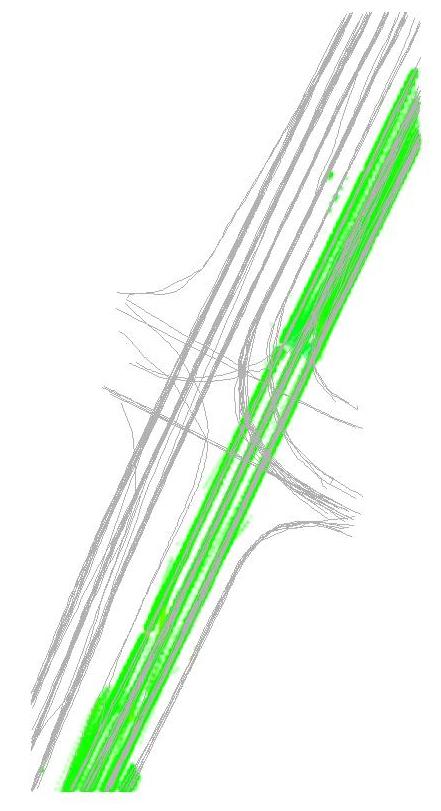}}
	\subfigure[]{\label{colorwheel7}
							\includegraphics[width=0.03\textwidth]{figures/colorwheel.jpg}}
							 \caption{\footnotesize{A subset of NGSIM Lankershim Dataset is used for visualizing the effects of parameters tuning. The effects of extremely large $\beta$, double-ellipse parameters and so large $th_{\theta\psi}$ are illustrated in \ref{flowWeight_1000_3}, \ref{ellipse_width_251} and \ref{th_flow_120_4}, respectively. Figures \ref{th_flow_2_9}, \ref{th_flow_2_28} and \ref{th_flow_2_36} show the effect of using too small $th_{\theta\psi}$. Finally, the effect of too large $th^{w}_{\rho}$ is shown in \ref{wedge_radius_20_angle_180_2}.}}
						\label{Discussion_vis}
\vspace{-4mm}
							 \end{center}
							\end{figure}
		
		                                         \begin{figure}[!ht]	
							  \begin{center}
							  \subfigure[]{\label{noWedgeSignatureZoomedIn}
							\includegraphics[width=0.22\textwidth]{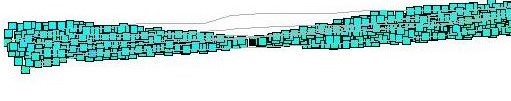}}
							  \subfigure[]{\label{WedgeSignatureZoomedIn}
							\includegraphics[width=0.22\textwidth]{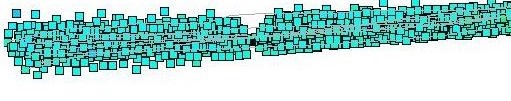}}
											 \caption{\footnotesize{\ref{noWedgeSignatureZoomedIn} shows motion components that form the signature for a motion component depicted in black. This signature is computed without using the wedge. Figure \ref{WedgeSignatureZoomedIn} shows the signature for the same motion component that is computed with the wedge parameters set to $th^{w}_{\psi}$ = 120, $th^{w}_{\rho}$ = 25, $th^{w}_{\theta}$ = 15.}}
						\label{wedgeEffect_discussion}
\vspace{-5mm}
							 \end{center}
							\end{figure}

			The first parameter is the number of clusters \math K$, which determines the number of motion components. It is important for the total number of motion components to be large in order to have a good resolution for the patterns. However, if the number is too large, some of the motion components may end up with only a few flow vectors. These motion components may be dramatically affected by noise and in some cases even form false motion patterns. It must also be noted that clustering is a computationally intensive task, hence the selection of \math K$ will also be influenced by the desired running time. Generally, datasets that are not expected to contain much noise will allow for arbitrarily large values of \math K$.

							The second parameter, $\beta$, is also used in clustering. This parameter controls how much the velocity of the flow vector affects the formation of motion components. If $\beta$ is too small, the spatial proximity will be the primary factor affecting the formation of the motion components, and each motion component's flow will be the average flow of its constituent flow vectors. Whenever this is the case, the algorithm will likely miss overlapping motion patterns with different flow directions, such as those seen in the Swainson's Hawks dataset. The larger $\beta$ becomes, the larger role the flow direction plays in clustering. A large $\beta$ will make the algorithm sensitive to smaller deviations in flow direction, as well as noise.  Extremely large $\beta$ will cause the flow direction to bias the clustering process, forming motion components from vectors with similar flow directions and dissimilar spatial coordinates, as illustrated in \ref{flowWeight_1000_3}, where $\beta$ is set to 1000, which is much higher than $\beta$ equal to 45, used for other results. In general, $\beta$ should be as small as possible while still allowing any meaningful overlap of patterns to exist. These two parameters completely determine the formation of motion components. Thus, these parameters can be fine-tuned without running the entire model but only visualizing the motion components. This will make the parameter selection process far simpler. 
						
			The next parameters are \math a_1$, \math b_1$, \math a_2$, and \math b_2$. These parameters determine the shape and size of the double-ellipse. In general, the double-ellipse should be small enough to avoid denoting motion components as reachable that belong to different motion patterns, yet large enough to \textit{jump} gaps in the motion component distribution within motion patterns. Example in \ref{ellipse_width_251} illustrates the outcome of setting the \math a_1$ and \math b_2$ of the ellipses too high. The "wider" ellipses \textit{jump} the gap between lanes.  Similar logic applies to manipulating \math a_2$, \math b_1$ values.
			The parameter $\textit th_{\theta\psi}$ should be chosen in a similar way except that this parameter deals with the flow direction rather than spatial position. $\textit th_{\theta\psi}$ should be large enough to capture deviations within motion patterns but not so large to make motion components reachable which belong to different motion patterns, as is demonstrated in \ref{th_flow_120_4}, where $\textit th_{\theta\psi}$ was set to 120 from its usual 12. Too low value, on the other hand, will yield results where even small differences in flow direction will make motion components unreachable. This effect is demonstrated in Figures \ref{th_flow_2_9}, \ref{th_flow_2_36} and \ref{th_flow_2_28}, where the right turn is spit into three motion patterns.
			
						The parameter $\alpha$ also deals with the flow direction of motion components. If this parameter is high, the algorithm expects well-formed curves, and if it is small, the algorithm will look for similar flow directions between reachable motion components rather than well-formed curves. If this parameter is set too high, the algorithm will be sensitive to noisy data and find curves where there are none. On the other hand, if it is set too low, the algorithm will have trouble finding curves - leading to the same problem as the one demonstrated in \ref{th_flow_2_9}, \ref{th_flow_2_36} and \ref{th_flow_2_28}. Because the algorithm expects curves, it sometimes connects two motion components with very different flow directions. To avoid this situation, a threshold  $\textit th_{\theta}$ is placed on the absolute value of $\theta$. In general, this will be two or three times the value of $\textit th_{\theta\psi}$.		
	
			The parameters so far discussed determine the core of the reachability relation. Among the reachability parameters, search distance deals with the unblocking process and the wedge parameters allow semi-lateral motions. To test this core reachability, the path reachable motion components and signature for some motion components can be computed with discounting the semi-lateral motion and unblocking process by setting the wedge parameters to 0 and search distance to 1. Then the signature for a motion component can be plotted and evaluated. However, for some datasets, these signatures will be proper for most motion components, but for other motion components, the signatures will unexpectedly end because of blocked motion components. To fix this, the search distance can be increased. However, if search distance is too large, the algorithm will make unreasonably far away motion components reachable. Once again, search distance value can be evaluated by visualizing the signatures of a sample of the motion components.

				Additionally, the wedge parameters can be set to nonzero values to include semi-lateral reachability and improve the signatures. In some datasets, the signatures will not cover the entire width of motion patterns, as in Figure \ref{noWedgeSignatureZoomedIn}, where a signature for a motion component is plotted. In these cases, the wedge should be used to improve the results. $th^{w}_{\theta}$ should be just high enough to include the width of motion patterns with somewhat noisy flow direction. $th^{w}_{\rho}$ should be just high enough to include gaps within the width of the motion pattern. $th^{w}_{\psi}$ should be high enough to expand a signature to the width of the motion pattern within a short distance, as demonstrated in Figure \ref{WedgeSignatureZoomedIn}. An example of wedge parameters at work is given in figure \ref{wedge_radius_20_angle_180_2}, where $th^{w}_{\psi}$ = 90, $th^{w}_{\rho}$ = 20, $th^{w}_{\theta}$ = 15, and the algorithm recovers all three parallel southbound lanes as one motion pattern. This concludes the reachability parameters that can be fine-tuned and tested independently by looking at the signatures of some of the motion components. The impact of wedge parameters can be examined by visualizing signatures. Finally, there is one parameter for the clustering of motion components. The cutoff value, determines the size of the motion patterns. The larger the cutoff, the larger the motion patterns will be.

										\section{Discussion}
							In many applications, the task of automatically detecting changes in trends is just as interesting as uncovering existing ones. For example, in traffic control, sudden disappearances of motion patterns may indicate lane or street closures or blockages due to accidents, fallen trees, flooding, and other causes. Changes in motion patterns of animals may be indicative of changes in their environment, such as those due to urban expansion or pollution. These changes are of particular interest to conservation biologists. Furthermore, zoologists are frequently interested in analyzing seasonal changes in movement of animals. In the domain of commerce, merchandisers and marketing professionals can use changes in shopper traffic patterns to improve advertisement and product visibility. 
	 We can mine changes in trends of trajectory data by either discovering newly emerged motion patterns or by finding motion patterns that no longer occur. Given trajectories observed from time $t_1$ to $t_i$ and from time $t_{i+1}$ to $t_j$, we generate two sets of motion patterns, $G$ and $G'$ respectively, using our proposed algorithm. $G$ contains motion patterns $g_i$ where $i=1,2,\ldots,n$ and $G'$ contains motion patterns $g'_i$ where $i=1,2,\ldots,n'$. Then, for any $g'_i$, using Kullback-Leibler (KL) divergence, we try to find a similar motion pattern in $G$. If no match is found, it means that $g'_i$ is a newly emerged motion pattern, a trend that was not previously observed in $t_1$ to $t_i$ interval. Repeating the process for all $g_i$, we can detect the disappearance of a motion pattern if no match is found in $G'$. In order to be able to use KL divergence for comparing two motion patterns, we need to have their probability distribution functions. The pdf of $g_i$ and $g'_i$ denoted as $p_{g_i}$ and $p_{g'_i}$ respectively, can be obtained by learning mixtures of Gaussians from flow vectors that are contained by their constituent motion components. Then we draw a sufficient number of samples from $p_{g_i}$ and evaluate probability of their occurrence in $p_{g'_i}$. A high probability means that two motion patterns are similar as their KL divergence is low.\cite{saleemi2010} has employed this approach to merge large numbers of similar motion patterns that occur at different times together while \cite{khokhar2011} performs event classification by matching the distribution of motion patterns that minimizes KL divergence.
\vspace{-1mm}
										\section{Conclusion}

			In this paper, we considered the general task of trajectory clustering using a novel approach inspired by the motion pattern idea. Our method consists of four main steps. First, we break down trajectories into flow vectors and then, using Kmeans clustering, we extract motion components. In the third step, we use the double-ellipse and the wedge conditions in addition to unblocking procedure to find reachable pairs of motion components. Finally, using the path reachability and signature concepts, we form motion patterns via agglomerative clustering with the weighted Jaccard distance between motion components' signatures. We evaluated our proposed method on five different datasets. Experimental results indicate that our motion pattern approach gives an effective solution to the general task of trajectory clustering regardless of the dataset properties. Extracted motion patterns closely fit the annotations, prevailing paths, or descriptions of trajectory datasets where available. We comprehensively discussed the effects of model parameters and provided a selection process for them. Also, we noted that the actual optimum set of parameters will rely on domain knowledge as well as specific analytical goals. In addition, we discussed how our proposed model is well suited for automatically detecting changes in frequent behaviors of trajectories over time.	
	
			Overall, we believe that we have provided a new approach for understanding trajectory behavior. Its output is comparable to the output of those trajectory clustering methods that look for regional similarities among trajectories. It is capable of handling a variety of challenges provided by different datasets. Our proposed method can provide data analysts a good starting point for understanding the hidden behavioral patterns in enormous and complex trajectory datasets.


						\ifCLASSOPTIONcaptionsoff
						  \newpage
						\fi

						\bibliographystyle{IEEEtran}

									\bibliography{Trajectorty_Behavior}

						\end{document}